\documentclass[letterpaper]{article} % DO NOT CHANGE THIS
\usepackage{aaai2026}  % DO NOT CHANGE THIS
\usepackage{times}  % DO NOT CHANGE THIS
\usepackage{helvet}  % DO NOT CHANGE THIS
\usepackage{courier}  % DO NOT CHANGE THIS
\usepackage[hyphens]{url}  % DO NOT CHANGE THIS
\usepackage{graphicx} % DO NOT CHANGE THIS
\usepackage{natbib}  % DO NOT CHANGE THIS AND DO NOT ADD ANY OPTIONS TO IT
\usepackage{caption} % DO NOT CHANGE THIS AND DO NOT ADD ANY OPTIONS TO IT
\usepackage{multirow}
\usepackage{multicol}
\usepackage{threeparttable}
\usepackage{booktabs}
\usepackage{xcolor}

\usepackage{algorithm}
\usepackage{algorithmic}
\usepackage{subcaption}
\usepackage{adjustbox}
\usepackage{amsmath}
\newcommand{\best}[1]{\textbf{#1}}
\newcommand{\second}[1]{\underline{#1}}
\usepackage{pifont}

\usepackage{amsmath}
\usepackage{amssymb}
\usepackage{amsfonts}

\usepackage{newfloat}
\usepackage{listings}
\usepackage{float}
\DeclareCaptionStyle{ruled}{labelfont=normalfont,labelsep=colon,strut=off} % DO NOT CHANGE THIS
\floatstyle{ruled}
\newfloat{listing}{tb}{lst}{}
\floatname{listing}{Listing}
\makeatletter
\renewcommand*{\@fnsymbol}[1]{%
  \ensuremath{%
    \ifcase#1\or
      \dagger\or        % 1 -> †
      \ddagger\or       % 2 -> ‡
      \mathsection\or   % 3 -> §
      \mathparagraph\or % 4 -> ¶
      \|\or             % 5 -> ‖
      \dagger\dagger\or % 6 -> ††
      \ddagger\ddagger\or % 7 -> ‡‡
      *% 8 -> *
    \else\@ctrerr\fi
  }%
}
\makeatother

\title{TimeMosaic: Temporal Heterogeneity Guided Time Series Forecasting via Adaptive Granularity Patch and Segment-wise Decoding}
\author {
    Kuiye Ding\textsuperscript{\rm 1},
    Fanda Fan\textsuperscript{\rm 1}\thanks{Corresponding author: fanfanda@ict.ac.cn.},
    Chunyi Hou\textsuperscript{\rm 2},
    Zheya Wang\textsuperscript{\rm 3}, \\
    Lei Wang\textsuperscript{\rm 1},
    Zhengxin Yang\textsuperscript{\rm 1},
    Jianfeng Zhan\textsuperscript{\rm 1,4}
}

\affiliations {
    \textsuperscript{\rm 1}Institute of Computing Technology, Chinese Academy of Sciences\\
    \textsuperscript{\rm 2}School of Information Science and Technology, Beijing University of Technology\\
    \textsuperscript{\rm 3}Department of Mathematical Sciences, Durham University\\
    \textsuperscript{\rm 4}University of Chinese Academy of Sciences\\
    \{dingkuiye, fanfanda, wanglei\_2011, yangzhengxin, zhanjianfeng\}@ict.ac.cn,\\
    houchunyi@emails.bjut.edu.cn,
    zheya.wang@durham.ac.uk
}

\usepackage{bibentry}
\usepackage{fancyhdr}
\begin{document}

\maketitle

\begin{abstract}

Multivariate time series forecasting is essential in domains such as finance, transportation, climate, and energy. However, existing patch-based methods typically adopt fixed-length segmentation, overlooking the heterogeneity of local temporal dynamics and the decoding heterogeneity of forecasting. Such designs lose details in information-dense regions, introduce redundancy in stable segments, and fail to capture the distinct complexities of short-term and long-term horizons. We propose \textbf{TimeMosaic}, a forecasting framework that aims to address temporal heterogeneity. TimeMosaic employs adaptive patch embedding to dynamically adjust granularity according to local information density, balancing motif reuse with structural clarity while preserving temporal continuity. In addition, it introduces segment-wise decoding that treats each prediction horizon as a related subtask and adapts to horizon-specific difficulty and information requirements, rather than applying a single uniform decoder. Extensive evaluations on benchmark datasets demonstrate that TimeMosaic delivers consistent improvements over existing methods, and our model trained on the large-scale corpus with 321 billion observations achieves performance competitive with state-of-the-art TSFMs.

% Extensive evaluations on benchmark datasets demonstrate that TimeMosaic delivers consistent improvements over existing methods.

\end{abstract}

\begin{links}
\link{Code}{https://github.com/BenchCouncil/TimeMosaic}
\end{links}

\section{Instruction}

Multivariate time series forecasting~\cite{yu2024ginar, DSformer, HybridZheng} plays a critical role in numerous real-world domains~\cite{qiu2025DBLoss, qiu2025dag}, such as finance~\cite{lsrigru, mcigru}, transportation~\cite{traffic_flow}, climate~\cite{schneider1974climate}, healthcare~\cite{health}, and energy management~\cite{liu2023sadi}. These applications require accurate and efficient modeling of intricate temporal relationships among multiple correlated variables. Recently, forecasting methods employing patch-based representations have shown remarkable performance and have become increasingly prevalent~\cite{chen2024pathformer,timellm,patchwise,patchtst}. Patch-based approaches excel at capturing localized temporal structures and mitigating noise through segmented encoding. Recently proposed time series models, such as Sundial~\cite{liu2025sundial}, also use patch-based time series segmentation to organize event contexts for benchmarking time series foundation models. Typically, these methods divide input sequences into fixed-length patches, implicitly assuming uniform information density and temporal complexity across the entire sequence. 

\begin{figure}[t]
  \centering
  \includegraphics[width=0.95\linewidth]{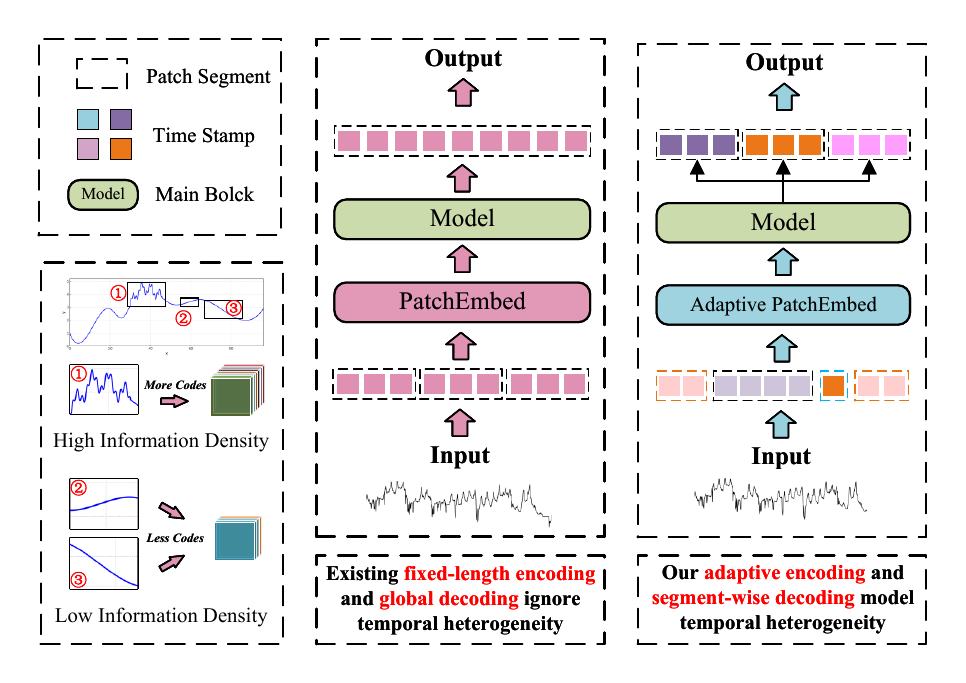}
  \caption{Comparison between existing fixed-patch models and our adaptive patching design, highlighting differences in input representation and prediction structure.}
  \label{fig:diff}
\end{figure}

However, real-world time series often exhibit significant variability in local information density~\cite{Huang_2023_CVPR, WARRENLIAO20051857}. Segments characterized by complex or abrupt changes inherently contain higher information densities, whereas smoother or stationary regions possess relatively lower densities. Current methods employing fixed-size temporal segments disregard this inherent variability, resulting in inadequate modeling of information-rich regions and redundant encoding in smoother ones. Beyond local variability, empirical evidence~\cite{xie2025languagetimelanguagemodel} also reveals two key structural properties: \emph{motif reuse}, characterized by Zipf-like frequency distributions, and \emph{structural clarity}, measured by latent-space separability (see Section~\ref{related_work} for definitions and illustrations), which further highlight a structural limitation of fixed-length patching: it cannot simultaneously preserve reusable long-range patterns and maintain well-delimited local boundaries. A similar limitation has been observed in autoregressive image generation, such as DQ-VAE~\cite{Huang_2023_CVPR}, where fixed-size region encodings likewise fail to adapt to diverse structural patterns.

These observations indicate that the limitations of fixed-length patching are not incidental but stem from a deeper property of time series data: \textbf{heterogeneity}. On the input side, local temporal regions vary greatly in complexity and information density, which we refer to as \textit{encoding heterogeneity}. On the output side, forecasting horizons differ in both difficulty and information requirements, which we refer to as \textit{decoding heterogeneity}. Addressing these two dimensions of heterogeneity is crucial for advancing time series forecasting.

Overall, multivariate time series forecasting faces two fundamental challenges: \textbf{\ding{172} Encoding heterogeneity.} Fixed-length segmentation fails to adapt to the variability of local temporal complexity, leading to a trade-off between reusing long-term motifs and preserving clear structural boundaries. \textbf{\ding{173} Decoding heterogeneity.} Forecasting demands differ substantially across horizons: short-term forecasting mainly rely on recent local information, while long-term forecasting require modeling more abstract and uncertain dynamics over broader contexts. Existing methods typically overlook this asymmetry by applying a single decoder to all horizons.  

To address these two challenges, we propose \textbf{TimeMosaic}, a novel forecasting framework that integrates adaptive patch embedding with segment-wise decoding. On the input side, we design an adaptive granularity patch embedding strategy inspired by dynamic quantization~\cite{Huang_2023_CVPR}. Our approach segments input sequences into variable-length patches based on local temporal information density, balancing motif reuse with structural clarity. Unlike prior multi-granularity methods that produce overlapping or re-ordered patches, TimeMosaic ensures that each time step belongs to exactly one patch, thereby preserving strict temporal continuity that is essential for forecasting. As illustrated in Figure~\ref{fig:diff}, fixed-length patching overlooks temporal heterogeneity: it loses details in information-dense regions and introduces redundancy in smooth intervals.

On the output side, we introduce a segment-wise prediction framework based on multi-task prompt tuning~\cite{liu2021gpt, crawshaw2020multitasklearningdeepneural}. By treating different horizons as distinct yet related subtasks, our method equips the model with segment-aware prompt embeddings that capture horizon-specific difficulty and information requirements without modifying the backbone parameters. This design enables specialization across prediction segments, effectively leveraging the rich and heterogeneous information captured by adaptive patches. As illustrated in Figure~\ref{fig:diff}, TimeMosaic adopting segment-wise forecasting that treats each horizon as a dedicated subtask.

We summarized our main contributions as follows:

\begin{itemize}
    \item We propose \textbf{TimeMosaic}, a novel forecasting framework that explicitly addresses both {encoding heterogeneity} and {decoding heterogeneity} in multivariate time series.
    \item We design an {Adaptive Patch Embedding} module that dynamically allocates region-specific patch sizes according to local information density, effectively balancing motif reuse with structural clarity.
    \item We introduce a {segment-wise prompt tuning} strategy that models horizon-specific difficulty and information requirements by treating each prediction segment as a distinct subtask within a unified multi-task framework.
    \item We conduct extensive experiments across 9 real-world datasets, showing that TimeMosaic achieves state-of-the-art performance in long-term forecasting.
\end{itemize}

\begin{figure}[t]
  \centering
  \includegraphics[width=0.9\linewidth]{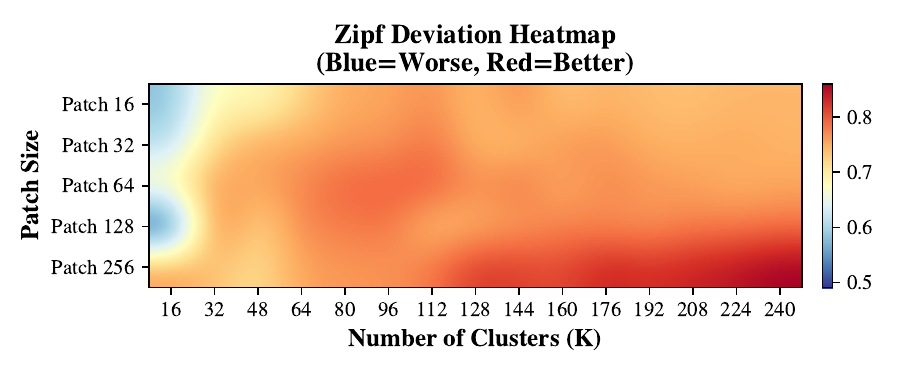}
  \vspace{1mm}
  \includegraphics[width=0.9\linewidth]{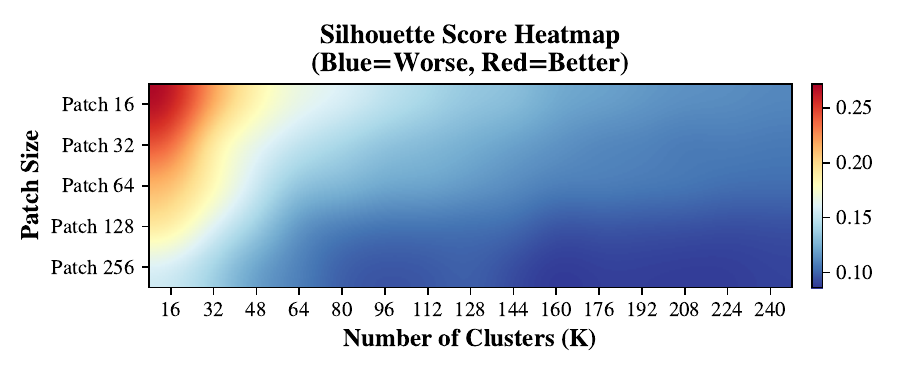}
  \vspace{-5pt}
  \caption{Zipf deviation and Silhouette score. These results are obtained by extracting patches of different lengths from a large-scale collection of time series forecasting datasets (see Appendix~A), followed by K-Means clustering under various cluster settings.}
  \label{fig:zipf_silhouette}
  \vspace{-10pt}
\end{figure}

\section{Related Work}
\label{related_work}

\paragraph{Patch-based Time Series Forecasting.}
Recent advances in time series forecasting have been driven by patch-based representations. PatchTST~\cite{patchtst} first introduced temporal patches into Transformer-based models, followed by Patchwise~\cite{patchwise}, which utilized local statistics, and TimeFilter~\cite{hu2025timefilter}, which applied fine-grained filtration. Time-LLM~\cite{timellm} further demonstrated the utility of patches in LLM-based forecasting. However, all these models adopt fixed-length patching, assuming uniform temporal complexity and lacking flexibility to adapt to local variations.

To overcome this limitation, multi-granularity patching has been explored. PatchMLP~\cite{patchmlp} introduced multiple segment lengths to enrich temporal features, while PathFormer~\cite{chen2024pathformer} and DualSG~\cite{ding2025dualsg} selected patch sizes based on importance scores or maps. Although effective, these methods may break temporal consistency due to overlapping or misordered patches. For instance, coarse patches can appear earlier than fine ones in the sequence, distorting input chronology and undermining forecasting coherence, see Appendix Figure ~\ref{fig:temporal-consistency}.

In contrast, our method introduces adaptive patch embedding with strict temporal alignment, dynamically adjusting patch sizes based on local information density while preserving the original time order.

\paragraph{Zipf Conformity and Clustering Clarity.} Recent research~\cite{xie2025languagetimelanguagemodel} shows that temporal patches in time series follow a Zipf-like frequency distribution~\cite{zipf}, exhibit clear clustering properties (measured by Silhouette scores~\cite{ROUSSEEUW198753}), and display strong sequential dependencies. Zipf conformity reveals the emergence of reusable temporal motifs, enabling compact representation and pattern reuse, while structural clarity reflects how distinctly different temporal patterns can be separated in latent space, facilitating stable representation learning and reliable generalization.

Building on these findings, we identify a key trade-off: fixed-length segmentation cannot optimize Zipf conformity and clustering clarity at the same time. Larger patches capture longer recurring patterns and thus improve Zipf conformity, but they blur behavioral boundaries. In contrast, smaller patches sharpen clustering clarity, yet they fragment long-term patterns and reduce reuse, as shown in Figure \ref{fig:zipf_silhouette}.

\paragraph{Autoregressive Forecasting Paradigm.}
Autoregressive (AR) models~\cite{lin2023segrnn, khaldi2023best, stankeviciute2021conformal} such as RNNs and LSTMs~\cite{kong2025unlockingpowerlstmlong, yadav2024noa} have long dominated time series forecasting. Even recent foundation models~\cite{shi2024timemoe, liu2024timer, ansari2024chronos} largely follow the AR formulation. However, these methods suffer from exposure bias~\cite{ranzato2016sequenceleveltrainingrecurrent} and equal treatment of short- and long-term predictions, despite their differing complexity and uncertainty.

Our approach departs from this paradigm by adopting a segment-wise multi-task formulation. This design avoids recursive error accumulation and allows segment-specific specialization, addressing both exposure bias and forecasting asymmetry.

\paragraph{Multi-task Learning and Prompt-based Adaptation.}
Multi-task learning (MTL)~\cite{ruder2017overviewmultitasklearningdeep, DBLP:journals/corr/abs-2009-09796} enhances generalization by sharing knowledge across tasks. In the context of large language models, it has been applied to summarization~\cite{laban-etal-2023-summedits}, QA~\cite{NEURIPS2023_9cb2a749}, and sentiment analysis~\cite{zhang-etal-2024-sentiment, cao2024enhanced}, often in combination with parameter-efficient methods like LoRA~\cite{NEURIPS2024_747dc7c6} and prompt tuning~\cite{power}.

We follow this trend by introducing prompt-based MTL into time series forecasting. Compared to basis-reconstruction methods such as TimeBase~\cite{huang2024timebase}, which rely on strong periodicity assumptions and static segment structures, our model learns dynamic, segment-aware prompts for flexible and efficient temporal adaptation. This enables better modeling of local heterogeneity without assuming low-rank structure or modifying the shared encoder.

\begin{figure*}[http]
    \centering
    \includegraphics[width=0.91\textwidth]{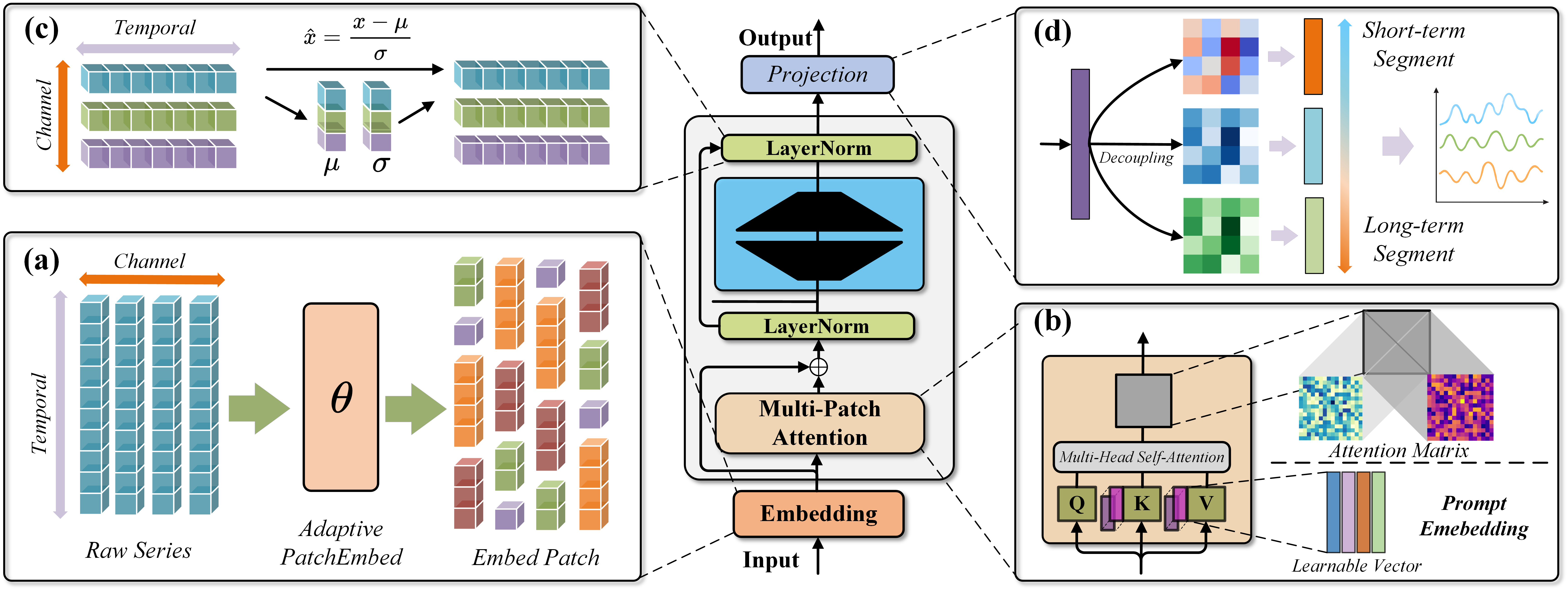}
    \caption{Overall architecture of TimeMosaic, which shares the same modular arrangement with the encoder of Transformer. (a) The input multivariate time series is first processed by the Adaptive Patch Embedding module, which segments the sequence into patches of varying granularity based on learnable region-aware decisions. (b) A set of learnable prompt tokens are injected into the input sequence and interact with patch embeddings via multi-head attention to guide segment-wise forecasting. The attention maps in this panel visualize the interactions among adaptive patches as well as between patches and prompts. (c) Input sequence are normalized at the Channel level, same as iTransformer~\cite{liu2023itransformer}. (d) The model performs segment-wise forecasting, from the short-term to the long-term.
    }
    \label{fig:architecture}
\end{figure*}

\section{Preliminaries and Definition}

% Let $\mathbf{X} \in \mathbb{R}^{C \times L}$ denote an input time series of $C$ channels and $L$ historical timesteps, and $\mathbf{Y} \in \mathbb{R}^{C \times H}$ the target sequence over $H$ future steps. Time series forecasting aims to learn a function $f$ that maps $\mathbf{X}$ to $\mathbf{Y}$.

% Depending on whether inter-channel dependencies are modeled, we distinguish two settings: \emph{channel-independent forecasting}, where predictions are made separately for each channel, and \emph{multivariate forecasting}, where all channels are modeled jointly via $f: \mathbb{R}^{C \times L} \rightarrow \mathbb{R}^{C \times H}$.

% Given a training set $\{(\mathbf{X}_i, \mathbf{Y}_i)\}_{i=1}^N$, we minimize the empirical forecasting loss:

% \begin{equation}
%     \mathcal{L}(f) = \frac{1}{N} \sum_{i=1}^N \| \mathbf{Y}_i - f(\mathbf{X}_i) \|_F,
% \end{equation}

% where $\|\cdot\|_F$ denotes the Frobenius norm.

\paragraph{Preliminaries.} In multivariate time series forecasting, we are given a historical sequence $\mathbf{X} \in \mathbb{R}^{C \times L}$, where $C$ is the number of channels and $L$ is the number of past timesteps. The goal is to predict the future $H$ steps, denoted as $\mathbf{Y} \in \mathbb{R}^{C \times H}$. A forecasting model learns a mapping $f: \mathbb{R}^{C \times L} \rightarrow \mathbb{R}^{C \times H}$ from past observations to future values.

\paragraph{Temporal heterogeneity.} Temporal heterogeneity refers to the non-uniform characteristics of time series across different temporal regions, encompassing both input-side variability and output-side forecasting challenges. On the input side, real-world sequences exhibit varying local complexity—volatile regions contain rapid fluctuations requiring fine-grained modeling, while stable segments allow for coarser abstraction. On the output side, prediction targets across different horizons pose asymmetric difficulties: short-term predictions often rely on recent patterns, whereas long-term predictions involve more abstract, uncertain dynamics. Ignoring such heterogeneity leads to suboptimal representations and a mismatch between model capacity and forecasting needs.

\section{Proposed Method}

\subsection{Rethinking Temporal Heterogeneity in Time Series Encoding and Decoding}
Real-world time series exhibit varying local complexity and asymmetric forecasting difficulty. Existing patch-based models typically adopt fixed-length segmentation, assuming uniform temporal dynamics. However, volatile regions demand fine-grained modeling, while smooth segments can be coarsely encoded; using a single patch size thus leads to either detail loss or redundant computation. This motivates a rethinking of both encoding and decoding strategies.

Forecasting over long horizons further introduces diverse semantic demands: early segments rely on recent trends, whereas distant predictions involve more abstract and uncertain dynamics. Yet most models decode all horizons with a shared head, ignoring this temporal asymmetry.

To address these challenges, we propose (1) an adaptive patch embedding module that learns region-specific granularity from local information density, and (2) a segment-wise prompt tuning framework that treats each forecast segment as a subtask. This design enables input-aware encoding and horizon-specific specialization, improving performance and robustness while maintaining parameter efficiency.

\subsection{Method Overview}

We propose a forecasting framework that combines adaptive patch embedding and segment-wise prompt tuning. The model first partitions the input into fixed-length regions, then selects a patch size per region from a candidate set. This yields variable-length patch sequences aligned to a uniform format via replication, enabling Transformer-compatible processing while preserving temporal coherence.

To forecast multi-step targets, we treat each future segment as a distinct subtask and assign it a learnable prompt. These prompts inject segment-specific biases into a shared encoder, allowing the model to specialize without altering backbone parameters. Additionally, we support both channel-independent and channel-dependent processing to flexibly handle inter-variable correlations. The entire framework is trained end-to-end with a loss that balances forecasting accuracy and patch usage diversity.

\subsection{Adaptive Patch Embedding}

To effectively model temporal heterogeneity in time series, we introduce an Adaptive Patch Embedding (APE) module that adjusts patch granularity based on local dynamics.

\paragraph{Patch Granularity Search Space.} We refer to \textit{regions} as local partitions of the input used for adaptive patching, and \textit{segments} as forecast sub-horizons used in multi-task prediction. This distinction allows us to separately model spatial encoding and temporal decoding. Given an input time series $x \in \mathbb{R}^{B \times C \times L}$, we divide the sequence into $R = L / f_{\max}$ non-overlapping regions, where $f_{\max}$ is the maximum patch length in the candidate set:
\begin{equation}
\mathcal{F} = \{f_1, f_2, \dots, f_K\}, \quad f_1 < f_2 < \dots < f_K.
\end{equation}
Each candidate patch size $f_k$ represents a different temporal granularity. For each region, our goal is to select the most suitable $f_k$ to match its local information density.

\paragraph{Region-wise Granularity Classification.}
To predict the optimal patch size for each region, we employ a lightweight classifier $\mathcal{G}_\theta$:
\begin{equation}
\mathbf{g}_r = \mathcal{G}_\theta(x_r) \in \mathbb{R}^{K}, \quad \theta_r = \arg\max_k g_{r,k},
\end{equation}
where $g_r$ is the predicted logits over candidate patch sizes for region $x_r$, and $\theta_r$ is the index of the selected patch size. The classifier is shared across all regions and channels, and is implemented as a two-layer MLP. To ensure end-to-end differentiability, we apply the Gumbel-Softmax~\cite{jang2017categorical} during training.

\paragraph{Patch Alignment and Embedding.}
After determining the patch size $f_{\theta_r}$ for region $x_r$, we unfold it into patches of size $f_{\theta_r}$:
\begin{equation}
z_r = \text{Linear}_{f_{\theta_r}} \left( \text{Unfold}(x_r; f_{\theta_r}) \right) \in \mathbb{R}^{N_r \times d},
\end{equation}
where $z_r$ is the embedded patch sequence, $N_r = f_{\max}/f_{\theta_r}$ is the number of patches in region $r$, and $d$ is the embedding dimension.
 Each patch is projected to a shared latent dimension $d$ using a dedicated linear layer.

To maintain a uniform input shape across regions, we upsample all patch sequences to a fixed length $N = f_{\max} / f_{\min}$ using replication:
\begin{equation}
\tilde{z}_r = \text{RepeatPad}(z_r; N).
\end{equation}

where $\tilde{z}_r$ is the length-aligned patch sequence for region $r$, padded to $N = f_{\max}/f_{\min}$ using replication. RepeatPad preserves temporal alignment and avoids introducing artificial content. In practice, we find it more stable than interpolation or learned resampling under varying patch configurations.

All region patch embeddings $\{\tilde{z}_1, \dots, \tilde{z}_R\}$ are concatenated and enriched with positional encodings to form the final input:
\begin{equation}
Z = \text{PE}\left( \text{Concat}(\tilde{z}_1, \dots, \tilde{z}_R) \right) \in \mathbb{R}^{B \cdot C \times N \cdot R \times d}.
\end{equation}

where $Z$ is the final encoder input sequence after adding positional encodings, $\mathrm{PE}(\cdot)$ denotes the positional encoding function, and $R$ is the number of input regions. We explore several alternative designs for positional encoding, including fixed, learnable, and hybrid schemes, which are analyzed in Section~\ref{Experiments}.

\subsubsection{Regularization via Budget Loss}

Without constraints, the classifier may degenerate to always selecting the finest patch size. To avoid this, we introduce a \emph{budget loss} to match the empirical distribution of selected granularities to a predefined target:
\begin{equation}
\mathcal{L}_{\text{budget}} = \sum_{k=1}^{K-1} \left( r_k - \hat{r}_k \right)^2,
\end{equation}
where $r_k$ is the desired usage ratio of patch size $f_k$, and $\hat{r}_k$ is the actual usage in the current batch. The last ratio is defined by normalization:
\begin{equation}
r_K = 1 - \sum_{k=1}^{K-1} r_k, \quad \hat{r}_K = 1 - \sum_{k=1}^{K-1} \hat{r}_k.
\end{equation}

\paragraph{Training Objective.}

We jointly optimize the forecasting loss and the budget regularization. The total training loss is defined as:
\begin{equation}
\mathcal{L}_{\text{total}} = \mathcal{L}_{\text{forecast}} + \lambda \mathcal{L}_{\text{budget}},
\end{equation}
where $\mathcal{L}_{\text{forecast}}$ is the standard mean squared error (MSE) between the prediction $\hat{y}$ and ground truth $y$:
\begin{equation}
\mathcal{L}_{\text{forecast}} = \frac{1}{B} \sum_{i=1}^{B} \| \hat{y}^{(i)} - y^{(i)} \|_2^2.
\end{equation}
$\lambda$ is a hyperparameter balancing the two terms.

\begin{table*}[http]
  \scriptsize
  \centering
  \setlength\tabcolsep{2.8pt}
  \begin{tabular}{c|cc|cc|cc|cc|cc|cc|cc|cc|cc|cc|cc}
    \hline
    \multirow{2}{*}{Models}                                                     &
    \multicolumn{2}{|c}{\textbf{TimeMosaic}}    & \multicolumn{2}{|c}{SimpleTM} & \multicolumn{2}{|c}{TimeFilter}         &
    \multicolumn{2}{|c}{xPatch}          & \multicolumn{2}{|c}{PatchMLP} & 
    \multicolumn{2}{|c}{DUET}            & \multicolumn{2}{|c}{PathFormer}     & 
    \multicolumn{2}{|c}{iTransformer}               & \multicolumn{2}{|c}{TimeMixer}      &
    \multicolumn{2}{|c}{PatchTST}           & \multicolumn{2}{|c}{DLinear}    \\
    \multirow{2}{*}{}                                                           &
    \multicolumn{2}{|c}{\textbf{(Ours)}}    & \multicolumn{2}{|c}{(2025)} & \multicolumn{2}{|c}{(2025)}       & 
    \multicolumn{2}{|c}{(2025)}             & \multicolumn{2}{|c}{(2025)}       & 
    \multicolumn{2}{|c}{(2025)}             & \multicolumn{2}{|c}{(2024)}       & 
    \multicolumn{2}{|c}{(2024)}             & \multicolumn{2}{|c}{(2024)}       &
    \multicolumn{2}{|c}{(2023)}             & \multicolumn{2}{|c}{(2023)}       \\
    \hline
    Metric
    & MSE & MAE & MSE & MAE & MSE & MAE & MSE & MAE & MSE & MAE & MSE & MAE & MSE & MAE & MSE & MAE & MSE & MAE & MSE & MAE & MSE & MAE \\
    \hline
    ETTm1
    & \best{0.381} & \best{0.381} &	0.391 &	{0.403} &	{0.395} &	0.407 &	0.391 &	0.402 &	0.401 &	0.407 &	0.416 &	0.407 &	0.392 &	\second{0.385} &	0.409 &	0.411 &	\second{0.381} &	0.396 &	0.387 &	0.398 & 0.407 & 0.409 \\
    \hline
    ETTm2
    & \second{0.273} &	\textbf{0.314} &	{0.283} &	 {0.325} &	0.282 &	0.329 &	0.283 &	0.327 &	0.287 &	0.331 &	0.286 &	0.328 &	\best{0.269} &	\second{0.314} &	0.293 &	0.334 &	0.279 &	0.325 &	0.279 &	0.323 & 0.350 & 0.400 \\
    \hline
    ETTh1
    & \best{0.425} &	\best{0.424} &	0.442 &	0.436 &	0.463 &	0.447 &	0.446 &	0.439 &	{0.455} &	{0.445} &	0.447 &	0.440 &	0.454 &	\second{0.428} &	0.450 &	0.442 &	0.445 &	0.439 &	\second{0.432} &	0.431 &	0.462 &	0.459 \\
    \hline
    ETTh2
    & \best{0.363} &	\best{0.388} &	0.424 &	{0.413} & 0.389 &	0.422 &	0.409 &	0.405 &	{0.402} &	0.416 &	0.379 &	0.400 &	\second{0.372} & \second{0.397} & 0.386 &	0.407 &	0.381 &	0.404 &	0.376 &	0.402 &	0.556 & 0.516\\
    \hline
    Weather
    & \second{0.251} &	\second{0.267} &	{0.258} &	{0.282} &	0.242 &	0.273 &	0.264 &	0.284 &	0.254 &	0.289 &	0.269 &	0.304 &	\best{0.242} &	\best{0.266} &	0.346 &	0.276 &	0.309 &	0.360 &	0.264 &	0.283 & 0.265 & 0.317 \\
    \hline
    Traffic
    & \second{0.458} &	\best{0.283} &	{0.545} &	{0.347} &	0.461 &	0.309 &	{0.520} &	{0.335} &	0.495 &	0.337 &	0.628 &	0.392 &	0.581 &	0.343 &	\best{0.453} &	0.303 &	0.499 &	\second{0.291} & 0.486 &	0.314 & 0.627 & 0.387 \\
    \hline
    Electricity
    & 0.187 & \second{0.279} &	{0.248} &	{0.284} &	\best{0.170} &	\best{0.266} &	{0.209} &	0.292 &	0.224 &	0.289 &	0.216 &	0.311 &	0.214 &	0.293 &	0.225 &	0.310 &	\second{0.185} &	0.294 &	0.205 &	0.292 & 0.215 & 0.305 \\
    \hline
    Exchange
    & \best{0.348} & \second{0.403} &	0.356 &	{0.411} &	0.366 &	0.411 &	0.372 &	0.410 &	0.374 &	0.412 &	0.356 &	{0.400} &	{0.375} &	0.411 &	{0.366} &	0.414 &	0.385 &	0.423 &	0.359 &	\best{0.401} & \second{0.354} & 0.420 \\
    \hline
    Solar
    & \best{0.240} & \second{0.270} &	0.315 &	{0.333} &	\second{0.248} &	\second{0.270} &	\second{0.248} &	0.276 &	0.255 &	0.280 &	0.349 &	0.350 &	0.255 &	\best{0.258} &	0.378 &	0.350 &	0.292 &	0.281 &	0.326 &	0.397 & 0.256 & 0.304\\
    \hline
    % $1^{st}$ Count & \textbf{29} & \textbf{31} & \underline{6} & \underline{7} & 3 & 0 & 0 & 0 & 1 & 0 & 4 & 4 & 0 & 0 & 0 & 0 \\
    % \hline
    \end{tabular}
    \caption{Averaged forecasting results under unified experimental settings. For \textbf{long-term} forecasting tasks, we use a fixed lookback window of $L = 96$. Long-term forecasting results are averaged over four prediction lengths: $T = \{96, 192, 336, 720\}$. The best model is in \textbf{boldface}, and the second best is \underline{underlined}. See Table~\ref{tab::long-term} in Appendix~\ref{detres} for complete results.}
    \label{tab:experiments1}
\end{table*}

\subsection{Prompt Tuning for Segment-wise Forecasting}

% We formulate multi-interval time series forecasting as a segment-wise multitask learning problem, where each forecast segment corresponds to an individual prediction subtask. To achieve parameter-efficient segment adaptation, we draw inspiration from Prompt Tuning \cite{power, liu2021gpt} and extend it to the temporal domain. Specifically, we design learnable segment-aware prompts to inject segment-specific inductive biases into a shared encoder, allowing the model to specialize its attention behavior for each segment without modifying backbone parameters. Compared with adapter-based or decoder-tuning approaches, prompt tuning achieves similar segment-level specialization with fewer trainable parameters and better modularity, as prompts can be flexibly added or removed without altering the shared encoder.

We formulate multi-interval time series forecasting as a segment-wise multitask learning problem, where each forecast segment corresponds to an individual prediction subtask. To achieve parameter-efficient segment adaptation, we draw inspiration from Prompt Tuning \cite{power, liu2021gpt} and extend it to the temporal domain. Specifically, we design learnable segment-aware prompts to inject segment-specific inductive biases into a shared encoder, allowing the model to specialize its attention behavior for each segment without modifying backbone parameters. 

Following standard practices in prompt tuning, these prompts are inserted only into the key and value paths of the attention mechanism, while the query vectors are derived exclusively from the input data. This asymmetric design ensures that the original data tokens retain their semantic role as information seekers, while prompts act as soft guidance to steer the attention focus in a segment-aware manner. Compared with adapter-based or decoder-tuning approaches, prompt tuning achieves similar segment-level specialization with fewer trainable parameters and better modularity, as prompts can be flexibly added or removed without altering the shared encoder.

For a given segment $k \in \{1, \dots, K\}$, we associate a prompt embedding $\boldsymbol{\phi}_k \in \mathbb{R}^{l \times d}$ and prepend it to the input representation $\mathbf{X} \in \mathbb{R}^{n \times d}$:
\begin{equation}
\tilde{\mathbf{X}}_k = \text{Concat}(\boldsymbol{\phi}_k, \mathbf{X}).
\end{equation}

To preserve a clear functional separation between prompts and data tokens, we apply a prompt-masked attention mechanism. In the self-attention computation, query vectors are derived exclusively from data tokens, while key and value vectors include both data and prompts:
\begin{equation}
\mathbf{Q}_k = \mathbf{X}W^Q, \quad
\mathbf{K}_k = \tilde{\mathbf{X}}_k W^K, \quad
\mathbf{V}_k = \tilde{\mathbf{X}}_k W^V.
\end{equation}
This allows the segment-aware prompt to modulate attention flow and inject task-relevant semantics, while being excluded from explicit decoding.

After encoding, each segment is decoded by a segment-specific head $f_k(\cdot; \theta_k)$, yielding the prediction $\hat{\mathbf{Y}}^{(k)} \in \mathbb{R}^{m_k \times C}$:
\begin{equation}
\hat{\mathbf{Y}}^{(k)} = f_k(\mathbf{H}_k; \theta_k),
\end{equation}
where $\mathbf{H}_k$ is the shared encoder output.

This segment-wise prompt tuning framework enables flexible modeling of segment-specific dynamics, improves parameter efficiency, and avoids interference across subtasks. All encoder parameters remain frozen during training, and only prompts $\{\boldsymbol{\phi}_k\}$ and decoding heads $\{\theta_k\}$ are updated.

\section{Experiments}
\label{Experiments}

We thoroughly evaluate the proposed TimeMosaic on various time series forecasting applications, validate the generality of the proposed framework and further delve into the effectiveness of adaptive patch embedding and Segment-wise Forecasting in other models.

\paragraph{Datasets.} We conduct extensive experiments on 17 real-world multivariate time series datasets, which are divided into long-term and short-term forecasting tasks. For long-term forecasting, we use ETTh1, ETTh2, ETTm1, ETTm2~\cite{haoyietal-informer-2021}, Weather, Traffic, Electricity, ExchangeRate~\cite{timesnet}, Solar-Energy~\cite{lai2018modelinglongshorttermtemporal}, and Wind (Location1–4)~\cite{xie2025languagetimelanguagemodel}. For short-term forecasting, we adopt four benchmark traffic datasets: PEMS03, PEMS04, PEMS07, and PEMS08~\cite{wang2023timemixer, wang2024timemixer++}.

\paragraph{Baselines.} We choose the last state-of-the-art LTSF models, including TimeFilter~\cite{hu2025timefilter}, SimpleTM~\cite{chen2025simpletm}, PatchMLP~\cite{patchmlp}, xPatch~\cite{stitsyuk2025xpatch}, DUET~\cite{qiu2025duet}, PathFormer~\cite{chen2024pathformer}, iTransformer~\cite{liu2023itransformer}, TimeMixer~\cite{wang2023timemixer}, PatchTST~\cite{patchtst},  FreTS~\cite{yi2023frequencydomain}, DLinear~\cite{Zeng2022AreTE}, LightTS~\cite{zhang2022morefastmultivariatetime} as baselines for our experiments. We compared the ability of zero-shot with the GPT4TS~\cite{zhou2023one} and LLMTime~\cite{gruver2023llmtime}.And we compare against pre-trained foundation time series models, including TimeMoE~\cite{shi2024timemoe}, MOIRAI~\cite{woo2024moirai}, Chronos~\cite{ansari2024chronos}, TimesFM~\cite{timesfm}, and Moment~\cite{goswami2024moment}, using the parameter settings as specified in their respective publications.

\paragraph{Evaluation Metrics.} Following previous works, we use Mean Squared Error (MSE) and Mean Absolute Error (MAE) metrics to assess the performance.

\paragraph{Channel Modeling Strategy.} Although our core design centers on segment-wise forecasting with adaptive patch granularity, handling multivariate inputs remains essential. To ensure compatibility, we incorporate a modular \textit{channel modeling component}, which supports both {Channel-Independent} (CI)~\cite{patchtst} and {Channel-Dependent} (CD)~\cite{hu2025timefilter} schemes. Specifically, we provide multiple instantiations ranging from simple per-variable modeling to prompt-augmented or calendar-aware variants. These options are fully pluggable and do not interfere with the main forecasting pipeline. By default, we adopt the Channel dependent strategy. A comprehensive description and comparison of all variants are provided in Appendix~\ref{channel}.

\paragraph{Implementation Details.} All the experiments are implemented in PyTorch~\cite{paszke2019pytorchimperativestylehighperformance}, and conducted on eight A800 GPU. Even in recent studies, it is common to set $drop\_last=True$ during data preprocessing~\cite{chen2025simpletm, wang2025fredf}, and this issue has been explicitly discussed in the TFB~\cite{qiu2024tfb}, highlighting its potential impact on evaluation fairness. Following the time series forecasting benchmark TFB's settings~\cite{qiu2024tfb}, we do not use the "drop last" trick during the testing phase to ensure a fair comparison. Our benchmark comprehensively includes 20+ state-of-the-art forecasting models across 17 real-world datasets, ensuring a fair comparison by unifying implementation settings and avoiding over-tuning parameter. And some approaches~\cite{hu2025timefilter} discard the use of \textit{EarlyStopping}, which in our view disregards the essential role of the validation set. Therefore, when integrating with other methods, we consistently retain this setting to ensure fair and reliable evaluation.

\begin{figure}[t]
  \centering
  \hfill
  \begin{subfigure}[b]{0.45\linewidth}
    \centering
    \includegraphics[width=\linewidth]{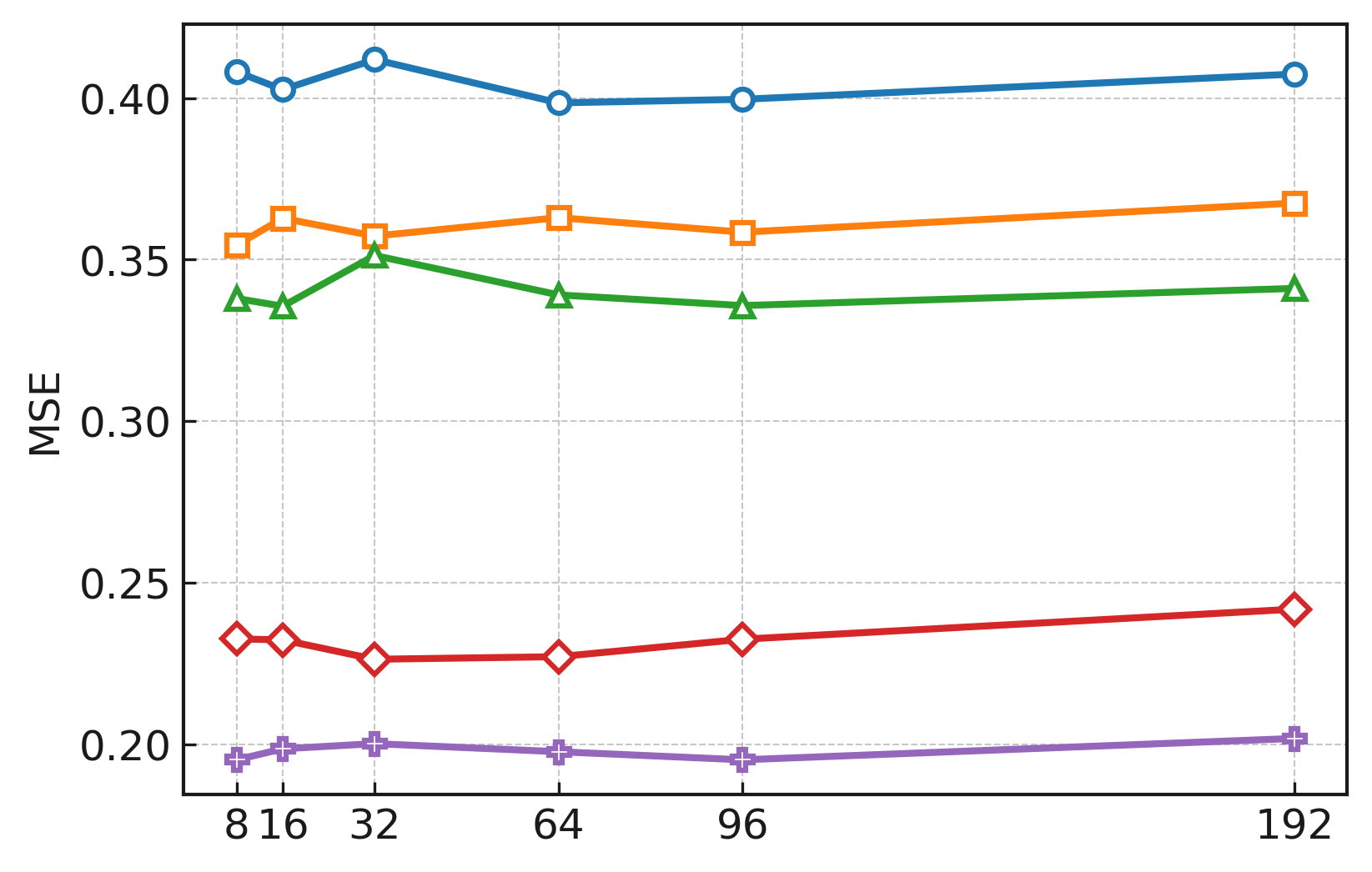}
  \end{subfigure}
  \hfill
  \begin{subfigure}[b]{0.45\linewidth}
    \centering
    \includegraphics[width=\linewidth]{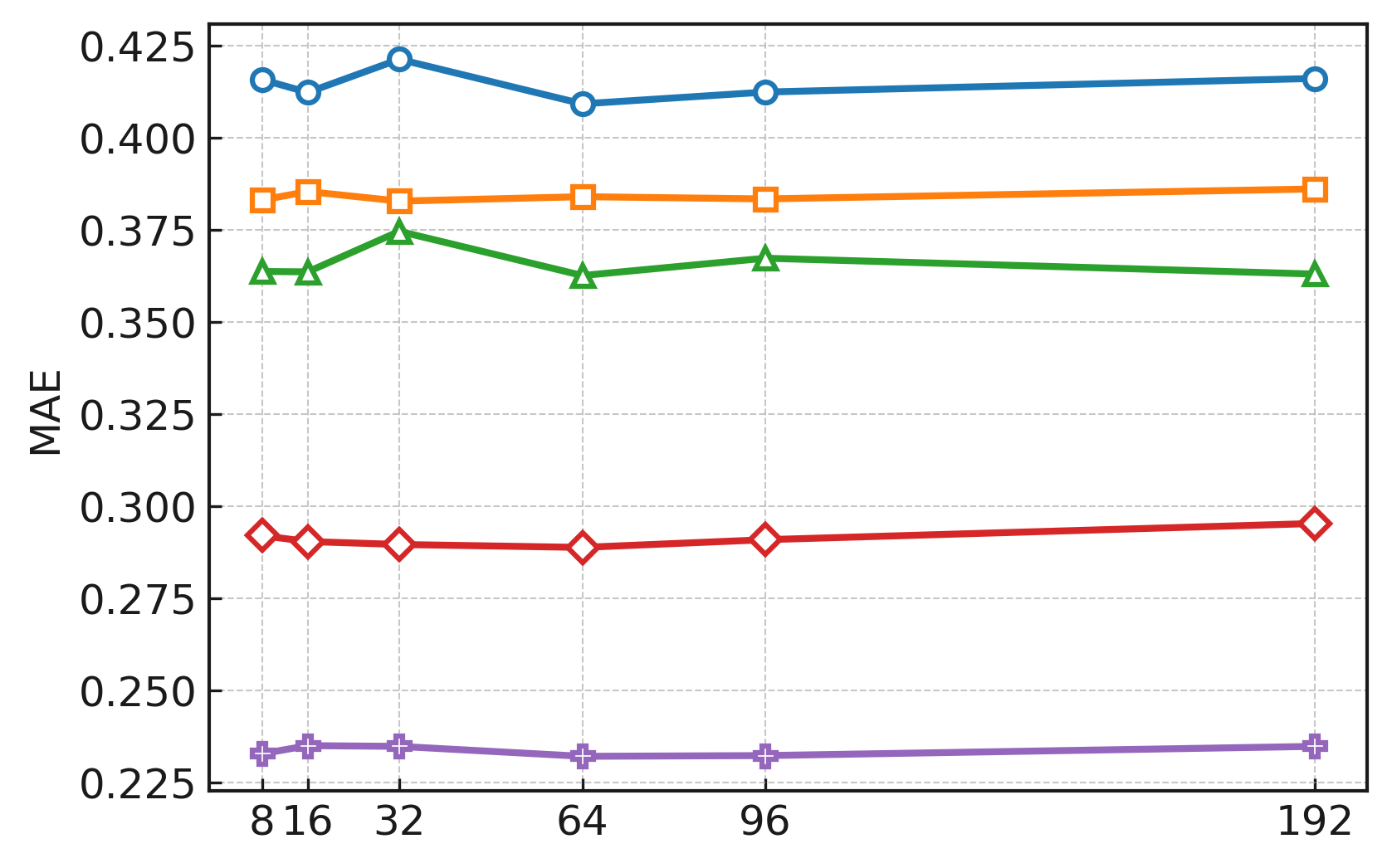}
  \end{subfigure}
  \hfill
  
  \begin{subfigure}[b]{0.8\linewidth}
    \centering
    \includegraphics[width=\linewidth]{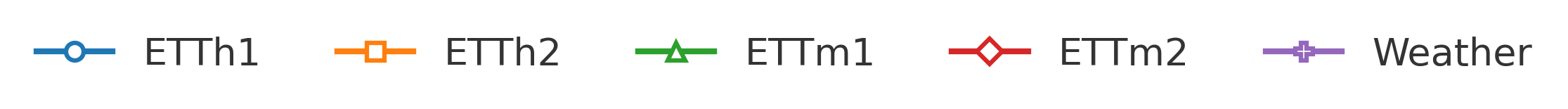}
  \end{subfigure}
  \caption{Segments of different sizes on five datasets with a prediction length of 192. Left: MSE. Right: MAE.}
  \label{fig:segment_mse_mae_paper}
\end{figure}

\begin{table}[h!]
\centering
\setlength{\tabcolsep}{5pt}
\renewcommand{\arraystretch}{1.1}
\begin{tabular}{@{}lcc@{}}
\toprule
\textbf{Model Variant} & \textbf{MSE} & \textbf{MAE} \\
\midrule
Baseline        & 0.277 & 0.326 \\
+ Adaptive Patch Embedding (APE)    & 0.258 & 0.302 \\
+ APE and Segment-wise Prompt Tuning        & \textbf{0.254} & \textbf{0.301} \\
\bottomrule
\end{tabular}
\caption{Ablation study of each component on five datasets. See datils in Appendix Table~\ref{tab:ablation}.}
\label{tab:ablation_study}
\end{table}

\begin{table}[h]
  \centering
  \begin{tabular}{ccc}
    \toprule
    \textbf{Granularity} & \textbf{MSE} & \textbf{MAE} \\
    \midrule
    \texttt{[8, 16]}         & 0.231 & 0.288 \\
    \texttt{[8, 32]}         & \textbf{0.228} & \textbf{0.286} \\
    \texttt{[8, 16, 32]}     & 0.232 & 0.290 \\
    \texttt{[8, 16, 64]}     & 0.230 & 0.289 \\
    \texttt{[8, 32, 64]}     & 0.235 & 0.289 \\
    \texttt{[4, 8, 16, 32]}  & 0.232 & 0.290 \\
    \texttt{[8, 16, 32, 64]} & \textbf{0.228} & \textbf{0.286} \\
    \bottomrule
  \end{tabular}
  \caption{Impact of patch granularity on six datasets, see details in Appendix Table~\ref{tab:patch-mode-ablation}.}
  \label{tab:granularity_ablation}
\end{table}

\paragraph{Unified Experimental Settings.} To ensure fair comparison, we conduct two types of experiments. For long-term forecasting, we follow the unified evaluation protocol proposed by TimesNet~\cite{timesnet}, using a lookback length $L=96$ and prediction lengths $T = {96, 192, 336, 720}$ across all long-term datasets.
For short-term forecasting, we adopt a fixed prediction length $T=12$, which is commonly used in traffic prediction tasks. The lookback length is also set to $L=96$.
The averaged results under this unified setting are reported in Table~\ref{tab:experiments1}. Table~\ref{tab:experiments1} presents results across all models using best configurations obtained via hyperparameter tuning, including the number of layers, attention heads, hidden dimension ($d_{\text{model}}$), and feedforward dimension ($d_{\text{ff}}$), with a lookback window of 96. In contrast, Table~\ref{tab:fair-experiments} reports results with an identical parameter setting shared by all models, using a longer lookback window of 320.

\begin{table*}[t!]
\begin{center}{
% \footnotesize
\scriptsize
\setlength\tabcolsep{3pt}
\begin{tabular}{c|c|cc|cc|cc|cc|cc|cc|cc|cc|cc}
\toprule
\multicolumn{2}{c|}{Methods}&\multicolumn{2}{c|}{TimeMosaic}&\multicolumn{2}{c|}{$\mathrm{TimeMoE_{l}}$}&\multicolumn{2}{c|}{$\mathrm{TimeMoE_{b}}$}&\multicolumn{2}{c|}{$\mathrm{MOIRAI_{l}}$}&\multicolumn{2}{c|}{$\mathrm{MOIRAI_{b}}$}&\multicolumn{2}{c|}{$\mathrm{Chronos_{b}}$}&\multicolumn{2}{c|}{$\mathrm{Chronos_{s}}$}&\multicolumn{2}{c|}{TimesFM}&\multicolumn{2}{c}{Moment}\\

\midrule
\multicolumn{2}{c|}{Metric} & MSE & MAE & {MSE} & {MAE} & MSE & MAE & MSE & MAE & MSE & MAE & MSE & MAE & MSE & MAE & MSE & MAE & MSE & MAE \\
\midrule
\multirow{5}{*}{\rotatebox[origin=c]{90}{ETTh1}}  
& 96  & 0.367 & 0.395 
& \best{0.350} & 0.382 
& \second{0.357} & \second{0.381} 
& 0.381 & 0.388  
& 0.376 & 0.392 
& 0.384 & \best{0.379} 
& 0.394 & \second{0.381} 
& 0.414 & 0.404 
& 0.688 & 0.557 \\

& 192 
& 0.395 & \second{0.412} 
& \second{0.388} & \second{0.412} 
& \best{0.384} & \best{0.404} 
& 0.434 & 0.415 
& 0.412 & 0.413 
& 0.441 & \second{0.412} 
& 0.455 & 0.414 
& 0.465 & 0.434 
& 0.688 & 0.560 \\

& 336 
& \best{0.410} & \best{0.423} 
& \second{0.411} & 0.430 
& \second{0.411} & 0.434 
& 0.495 & 0.445 
& 0.433 & \second{0.428} 
& 0.475 & 0.430 
& 0.499 & 0.444 
& 0.503 & 0.456 
& 0.675 & 0.563 \\

& 720 
& \best{0.422} & \best{0.443} 
& \second{0.427} & 0.455 
& 0.449 & 0.477 
& 0.611 & 0.510 
& 0.447 & \second{0.444} 
& 0.472 & 0.446 
& 0.520 & 0.476 
& 0.511 & 0.481 
& 0.683 & 0.585 \\

\cmidrule(lr){2-20}

& Avg 
& \second{0.399} & \second{0.418} 
& \best{0.394} & 0.419 
& 0.400 & 0.424 
& 0.480 & 0.439 
& 0.417 & 0.419 
& 0.443 & \best{0.416} 
& 0.467 & 0.428 
& 0.473 & 0.443 
& 0.683 & 0.566 \\

\midrule
\multirow{5}{*}{\rotatebox[origin=c]{90}{ETTm2}}  
& 96  
& 0.190 & 0.276 
& 0.197 & 0.286 
& 0.201 & 0.291 
& 0.211 & \second{0.274} 
& 0.205 & \second{0.273} 
& \best{0.177} & \best{0.244} 
& \second{0.180} & 0.251 
& 0.202 & 0.270 
& 0.260 & 0.335 \\

& 192 
& \best{0.249} & 0.313 
& 0.250 & 0.322 
& 0.258 & 0.334 
& 0.281 & 0.318 
& 0.275 & 0.316 
& \second{0.251} & \best{0.293} 
& \second{0.251} & \second{0.298} 
& 0.289 & 0.321 
& 0.289 & 0.350 \\

& 336 
& \best{0.305} & 0.347 
& 0.337 & 0.375 
& 0.324 & 0.373 
& 0.341 & 0.355 
& 0.329 & 0.350 
& \best{0.305} & \best{0.327} 
& \second{0.315} & \second{0.338} 
& 0.360 & 0.366 
& 0.324 & 0.369 \\

& 720 
& \second{0.396} & 0.407 
& 0.480 & 0.461 
& 0.488 & 0.464 
& 0.428 & \second{0.428} 
& 0.437 & \second{0.411} 
& 0.419 & \best{0.394} 
& 0.421 & \best{0.403} 
& 0.462 & 0.430 
& \best{0.394} & 0.409 \\

\cmidrule(lr){2-20}

& Avg 
& \best{0.285} & 0.336 
& 0.316 & 0.361 
& 0.317 & 0.365 
& \second{0.315} & 0.343 
& 0.311 & \second{0.337} 
& \second{0.288} & \best{0.314} 
& 0.291 & 0.330 
& 0.328 & 0.346 
& 0.316 & 0.365 \\

\bottomrule
\end{tabular}
}
\caption{
Zero-shot forecasting results on two datasets. TimeMosaic is trained with an input length of 512 and an output horizon of 720. The symbols $s$, $b$, and $l$ represent the small, base, and large versions, respectively. See details in Appendix Table~\ref{tab:blast}
}

\label{tab:short_blast}
\end{center}
\end{table*}

\begin{table*}[h]
  \scriptsize
  \centering
  \setlength\tabcolsep{2.8pt}
  \begin{tabular}{c|cc|cc|cc|cc|cc|cc|cc|cc|cc|cc|cc}
    \hline
    \multirow{2}{*}{Models}                                                     &
    \multicolumn{2}{|c}{\textbf{TimeMosaic}}    & \multicolumn{2}{|c}{SimpleTM} & \multicolumn{2}{|c}{TimeFilter}         &
    \multicolumn{2}{|c}{xPatch}          & \multicolumn{2}{|c}{PatchMLP} & 
    \multicolumn{2}{|c}{DUET}     & 
    \multicolumn{2}{|c}{iTransformer}               & \multicolumn{2}{|c}{TimeMixer}      &
    \multicolumn{2}{|c}{PatchTST}           & \multicolumn{2}{|c}{DLinear}    & \multicolumn{2}{|c}{FreTS}\\
    \multirow{2}{*}{}                                                           &
    \multicolumn{2}{|c}{\textbf{(Ours)}}    & \multicolumn{2}{|c}{(2025)} & \multicolumn{2}{|c}{(2025)}       & 
    \multicolumn{2}{|c}{(2025)}             & \multicolumn{2}{|c}{(2025)}       & 
    \multicolumn{2}{|c}{(2025)}       & 
    \multicolumn{2}{|c}{(2024)}             & \multicolumn{2}{|c}{(2024)}       &
    \multicolumn{2}{|c}{(2023)}             & \multicolumn{2}{|c}{(2023)}     & \multicolumn{2}{|c}{(2023)}       \\
    \hline
    Metric
    & MSE & MAE & MSE & MAE & MSE & MAE & MSE & MAE & MSE & MAE & MSE & MAE & MSE & MAE & MSE & MAE & MSE & MAE & MSE & MAE & MSE & MAE \\
    \hline
    ETTm1
    & 0.360 & \best{0.377} & \best{0.350} & \second{0.379} & 0.399 & 0.408 & \second{0.354} & 0.386 & 0.374 & 0.396 & 0.360 & 0.380 & 0.386 & 0.406 & 0.356 & 0.385 & 0.372 & 0.399 & 0.357 & 0380 & 0.369 & 0.391 \\
    \hline
    ETTm2
    & \best{0.256} & \best{0.310} & 0.263 & 0.318 & 0.278 & 0.330 & 0.262 & 0.322 & 0.269 & 0.324 & \second{0.257} & \second{0.314} & 0.293 & 0.338 & 0.265 & 0.323 & 0.278 & 0.332 & 0.282 & 0.348 & 0.284 & 0.339 \\
    \hline
    ETTh1
    & \best{0.412} & \best{0.423} & 0.428 & 0.438 & 0.511 & 0.482 & 0.446 & 0.445 & 0.462 & 0.460 & \second{0.414} & \second{0.429} & 0.487 & 0.475 & 0.430 & 0.438 & 0.503 & 0.485 & 0.429 & 0.444 & 0.461 & 0.465 \\
    \hline
    ETTh2
    & 0.363 & \second{0.393} & 0.377 & 0.404 & 0.402 & 0.421 & \second{0.354} & 0.397 & 0.380 & 0.416 & \best{0.344} & \best{0.389} & 0.411 & 0.425 & 0.358 & 0.396 & 0.398 & 0.427 & 0.492 & 0.478 & 0.495 & 0.483 \\
    \hline
    Weather
    & 0.231 & \best{0.256} & 0.230 & 0.267 & 0.231 & 0.270 & \best{0.224} & \second{0.262} & 0.230 & 0.267 & 0.248 & 0.280 & 0.237 & 0.272 & \second{0.227} & 0.266 & 0.236 & 0.271 & 0.246 & 0.299 & 0.230 & 0.277 \\  
    \hline
    Traffic
    & 0.433 & \second{0.287} & 0.468 & 0.336 & \best{0.420} & \best{0.284} & 0.430 & 0.300 & 0.525 & 0.382 & 0.462 & 0.330 & 0.463 & 0.335 & 0.444 & 0.316 & \second{0.428} & 0.297 & 0.445 & 0.308 & 0.461 & 0.313 \\
    \hline
    Electricity
    & \second{0.170} & \second{0.263} & 0.180 & 0.277 & \best{0.167} & \best{0.260} & \second{0.170} & 0.264 & 0.200 & 0.302 & 0.185 & 0.287 & 0.180 & 0.277 & 0.173 & 0.267 & 0.173 & 0.272 & \second{0.170} & 0.269 & 0.171 & 0.270 \\
    \hline
    Wind
    & 0.774 & \best{0.680} & 0.779 & 0.690 & 0.815 & 0.703 & 0.762 & 0.682 & 0.772 & 0.689 & 0.762 & 0.683 & 0.807 & 0.701 & 0.762 & 0.684 & 0.811 & 0.710 & \second{0.744} & \second{0.681} & \best{0.742} & \best{0.680} \\
    \hline
    Solar
    & 0.226 & \best{0.241} & 0.272 & 0.314 & \best{0.200} & 0.255 & \best{0.200} & \second{0.253} & 0.250 & 0.288 & 0.265 & 0.289 & 0.232 & 0.280 & 0.224 & 0.276 & 0.221 & 0.262 & 0.255 & 0.315 & \second{0.212} & 0.267 \\
    \hline
    % $1^{st}$ Count & \textbf{29} & \textbf{31} & \underline{6} & \underline{7} & 3 & 0 & 0 & 0 & 1 & 0 & 4 & 4 & 0 & 0 & 0 & 0 \\
    % \hline
    \end{tabular}
    \caption{Averaged forecasting results under a unified and more fair evaluation setting with a farther lookback window. Compared to Table~\ref{tab:experiments1}, Table~\ref{tab:fair-experiments} offers a fairer comparison by adopting an extended lookback window, which better supports models like DUET and TimeFilter to realize their potential. Moreover, we fix key hyperparameters such as \texttt{d\_model}, \texttt{d\_ff}, learning rate, and training epochs, avoiding test-set-based tuning and ensuring a fair evaluation. See details in Appendix~\ref{detres} and Table~\ref{tab::long-term-fair}.}
    \label{tab:fair-experiments}
\end{table*}

\paragraph{Ablation Experiments.} We evaluate the contributions of each component in Table~\ref{tab:ablation_study}. Adaptive Patch Embedding (APE) improves performance by enabling input-aware tokenization, while segment-wise prompt tuning further enhances accuracy through localized decoding. Table~\ref{tab:granularity_ablation} shows the results of different patch size combination. Figure~\ref{fig:segment_mse_mae_paper} demonstrates that segmented prediction consistently outperforms non-segmented baselines, while showing robustness across different segment sizes, see details in Appendix Figure~\ref{fig:segment_mse_mae}. We adopt a fixed segment length of $L/3$ without cherry-picking.

\paragraph{Main Results.} As shown in Table~\ref{tab:experiments1}, TimeMosaic consistently achieves the best or second-best performance across all benchmarks. In the more fair comparison shown in Table~\ref{tab:fair-experiments}, our method also achieves competitive performance. Furthermore, TimeMosaic demonstrates robust performance under longer lookback windows, as illustrated in Figure~\ref{fig:lookback_curve}. See short-term forecasting results in Appendix Table~\ref{tab:experiments_short}. Table~\ref{tab:search} compares the best hyperparameter configurations of seven SOTA models, where all models are tuned under the same search protocol. Due to the extremely high computational cost, each model requires exploring \textbf{10,800} different parameter settings, we only report results on seven models across five datasets. 

\paragraph{Zero$-$shot.} To assess cross-dataset generalization, we conduct zero-shot forecasting where models are trained on one dataset and tested on another. As shown in Appendix Table~\ref{tab:zero-shot-forecasting}, TimeMosaic consistently achieves the best performance across all transfer settings, outperforming GPT4TS~\cite{zhou2023one}, DLinear, and PatchTST. In addition, we further scale up TimeMosaic on the BLAST dataset~\cite{blast} with an input length of 512 and an output horizon of 720. This large-scale setting (27M parameters, trained on two V100 GPUs for about 40 hours) allows a direct comparison with recent time series foundation models (TSFMs). The results (Appendix~\ref{appendix:tsfm}, Tables~\ref{tab:short_blast} and~\ref{tab:params-efficiency}) show that TimeMosaic achieves zero-shot performance on par with representative TSFMs such as TimeMoe~\cite{shi2024timemoe}, Moirai~\cite{woo2024moirai}, Chronos~\cite{ansari2024chronos}, TimesFM~\cite{timesfm}, and Moment~\cite{goswami2024moment} while maintaining a moderate model size and inference cost. This confirms that the architectural choices of adaptive patch embedding and segment-wise decoding remain effective even at the foundation-model scale.

\begin{table*}[h] 
    \setlength{\tabcolsep}{3pt}
    % \scriptsize
    \footnotesize
    \centering
    \begin{threeparttable}
        \begin{tabular}{l*{8}{cc}@{}} 
             \toprule
                 \multicolumn{1}{c}{\multirow{2}{*}{Models}} & \multicolumn{2}{c}{\textbf{TimeMosaic}} & \multicolumn{2}{c}{SimpleTM} & \multicolumn{2}{c}{TimeFilter} & \multicolumn{2}{c}{DUET} & \multicolumn{2}{c}{iTransformer} & \multicolumn{2}{c}{TimeMixer} & \multicolumn{2}{c}{PatchTST} & \multicolumn{2}{c}{DLinear}\\
            
            & \multicolumn{2}{c}{\scalebox{0.8}{\textbf{Ours}}} &
            \multicolumn{2}{c}{\scalebox{0.8}{\citeyearpar{chen2025simpletm}}} & 
            \multicolumn{2}{c}{\scalebox{0.8}{\citeyearpar{hu2025timefilter}}} & 
            \multicolumn{2}{c}{\scalebox{0.8}{\citeyearpar{qiu2025duet}}}  & 
            \multicolumn{2}{c}{\scalebox{0.8}{\citeyearpar{liu2023itransformer}}} &
            \multicolumn{2}{c}{\scalebox{0.8}{\citeyearpar{wang2023timemixer}}} & 
            \multicolumn{2}{c}{\scalebox{0.8}{\citeyearpar{patchtst}}} & 
            \multicolumn{2}{c}{\scalebox{0.8}{\citeyearpar{Zeng2022AreTE}}}\\

            \cmidrule(lr){2-3} \cmidrule(lr){4-5} \cmidrule(lr){6-7} \cmidrule(lr){8-9} \cmidrule(lr){10-11} \cmidrule(lr){12-13} \cmidrule(lr){14-15} \cmidrule(lr){16-17}
            \multicolumn{1}{c}{Metric}& MSE & MAE & MSE & MAE & MSE & MAE & MSE & MAE & MSE & MAE & MSE & MAE & MSE & MAE & MSE & MAE\\
            \toprule
              \multicolumn{1}{c}{ETTm1} & \best{0.342} & \best{0.367} & \second{0.346} & 0.378 & 0.353 & \second{0.377} & 0.354 & \second{0.377} & 0.356 & 0.385 & 0.349 & 0.392 & 0.348 & 0.380 & 0.356 & 0.378 \\
              \multicolumn{1}{c}{ETTm2} & \best{0.250} & \best{0.305} & 0.257 & 0.316 & \second{0.253} & 0.317 & \second{0.253} & \second{0.313} & 0.267 & 0.325 & 0.256 & 0.315 & 0.256 & 0.318 & 0.262 & 0.326 \\
              \multicolumn{1}{c}{ETTh1} & \best{0.397} & \best{0.417} & 0.416 & 0.434 & 0.414 & 0.429 & \second{0.406} & \second{0.425} & 0.425 & 0.438 & 0.416 & 0.428 & 0.410 & 0.430 & 0.419 & 0.437 \\
              \multicolumn{1}{c}{ETTh2} & 0.348 & \best{0.383} & 0.358 & 0.396 & \second{0.347} & 0.394 & \best{0.334} & \second{0.384} & 0.362 & 0.398 & 0.349 & 0.392 & 0.359 & 0.399 & 0.412 & 0.433 \\
              \multicolumn{1}{c}{Weather} & \second{0.223} & \best{0.251} & \best{0.220} & \second{0.259} & \best{0.220} & 0.261 & 0.240 & 0.277 & 0.230 & 0.269 & 0.229 & 0.265 & \second{0.223} & 0.260 & 0.240 & 0.291 \\ 
            \bottomrule
        \end{tabular}
    \end{threeparttable}
    \caption{Results under the hyperparameter search setting described in Appendix Section~\ref{sec:hparam_search}. The lookback window is selected from $\{96, 192, 320, 512\}$, and the best configuration is reported for each model. This setup ensures that the comparison reflects each model’s optimal performance rather than a fixed setting constraint. As far as we know, we are the first open source parameter search script. See Details in Appendix Table \ref{tab::long-term-search}.}
    \label{tab:search}
\end{table*}

\begin{figure}[http]
    \centering
    \includegraphics[width=0.48\textwidth]{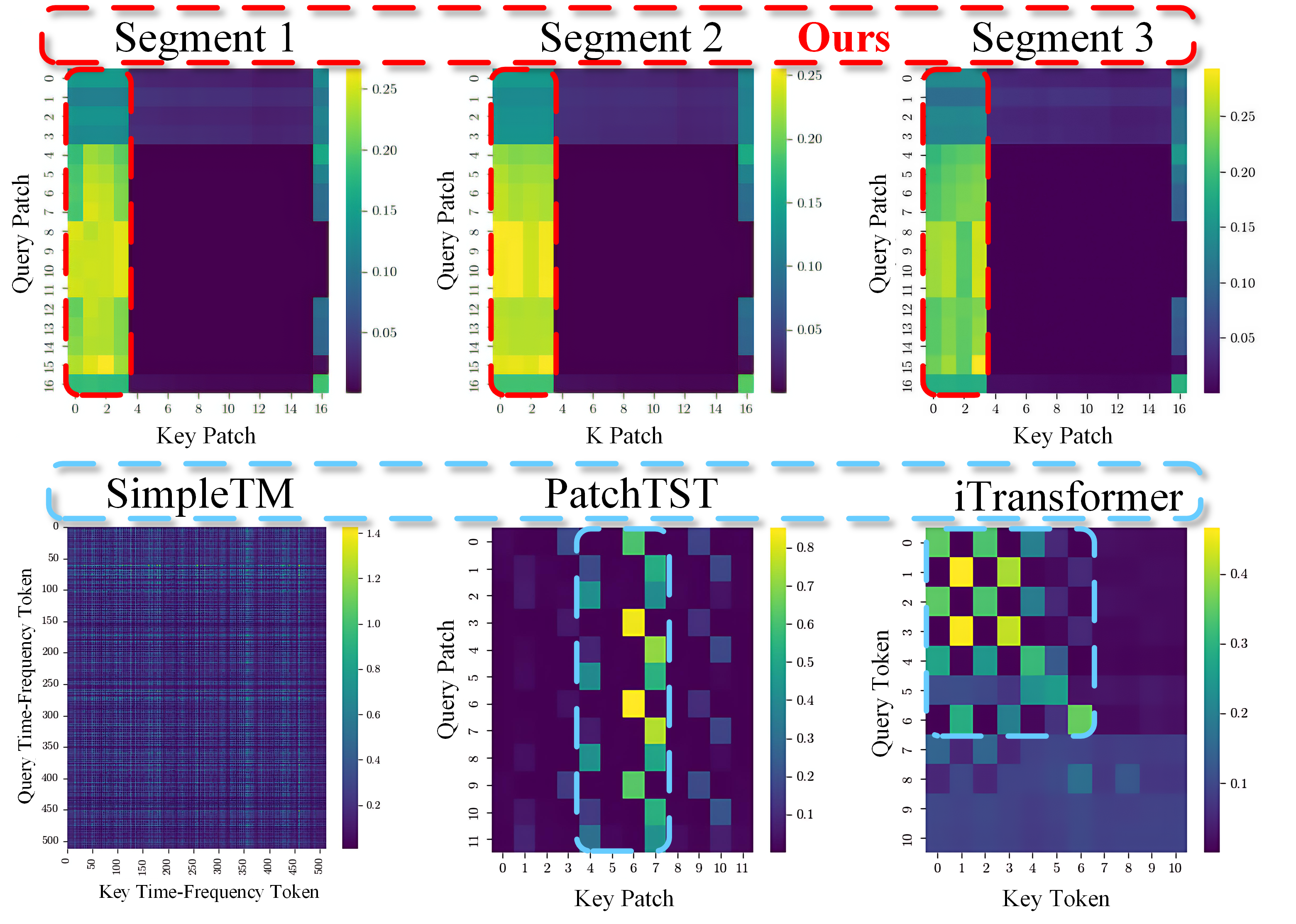}
    \caption{Attention Pattern Visualization Across Models.}
    \label{fig:attn}
\end{figure}

\begin{figure}[t!]
  % \centering
  \begin{subfigure}[b]{0.232\textwidth}
    \includegraphics[width=\linewidth]{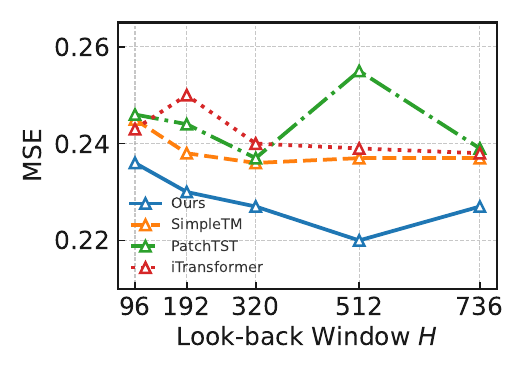}
  \end{subfigure}
  \begin{subfigure}[b]{0.232\textwidth}
    \includegraphics[width=\linewidth]{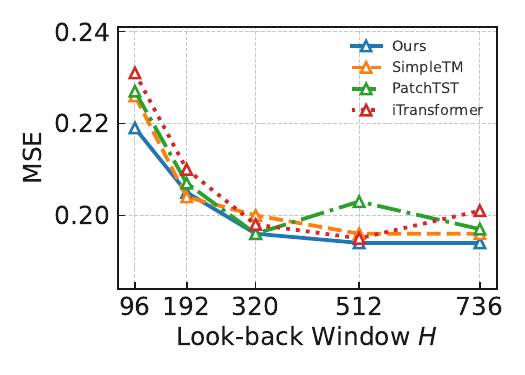}
  \end{subfigure}
  \caption{Sensitivity of multivariate forecasting performance to look-back window size. We evaluate four models across five input lengths ($T=192$) on ETTm2 and Weather. See more details in Appendix~\ref{lookback}.}
  \label{fig:lookback_curve}
\end{figure}

\paragraph{Efficiency Analysis.} Appendix Table~\ref{tab:model_efficiency} reports forecasting error and efficiency under the unified search protocol described in Appendix Section~\ref{sec:hparam_search}. For each dataset and model, we first conduct hyperparameter search and select the configuration that achieves the lowest validation error averaged over MSE and MAE. We then measure parameter count, inference latency, and GPU memory usage at this setting. This protocol anchors every method at its best attainable error level within the same search space and avoids biased comparisons that could arise from deliberately changing hyperparameters to influence efficiency. The results show that TimeMosaic achieves competitive or leading error on all five datasets, for example attaining the lowest MSE on ETTm2 and the lowest MAE on Weather, while keeping parameter scale in the range of a few hundred thousand, which is substantially smaller than SimpleTM or iTransformer. Although its inference time is higher than that of extremely lightweight models such as iTransformer, it remains within a reasonable margin and consistently uses moderate memory. The slightly slower inference mainly arises from the segment-wise step-by-step decoding design, and this behavior is consistent with our experimental observations. Overall, TimeMosaic presents a balanced error–efficiency performance, showing that its gains stem from architectural design rather than hyperparameter tuning.

\paragraph{Visualization.} Existing models differ in their attention interaction schemes. PatchTST slices each variable into temporal patches and applies self-attention across patches within the same variable. iTransformer treats each variable as a token and performs self-attention across channels. In contrast, SimpleTM constructs time-frequency tokens via wavelet transforms and enables joint attention across both time and channels. Our TimeMosaic instead interacts primarily through learnable \textit{prompt embeddings} assigned to different forecast segments, guiding the model to specialize decoding over distinct temporal intervals in Figure ~\ref{fig:attn}.

\section{Conclusion}

We proposed {TimeMosaic}, a forecasting framework that jointly addresses the dual challenges of {encoding heterogeneity} and {decoding heterogeneity} in multivariate time series. Through adaptive patch embedding, TimeMosaic dynamically adjusts granularity according to local information density, ensuring both motif reuse and structural clarity while maintaining temporal coherence. Complementarily, segment-wise prompt tuning enables horizon-aware forecasting by treating different prediction segments as related subtasks within a unified multi-task paradigm. Extensive experiments verify that TimeMosaic achieves consistent improvements in forecasting accuracy and efficiency. In future work, we will extend TimeMosaic to streaming scenarios and investigate its potential for large-scale pretraining towards time series foundation modeling.

\clearpage
\appendix
\setcounter{secnumdepth}{0}
\section{Acknowledgments}
This research was partly supported by the National Natural Science Foundation of China under Grant Nos. U24A20232 and 62402475.
% \setcounter{secnumdepth}{2}
% State Key Lab of Processors, Institute of Computing Technology, CAS 
\bibliography{aaai2026}

% \clearpage
% \input{checklist/ReproducibilityChecklist}

\clearpage
\appendix
\setcounter{secnumdepth}{2}

\section*{Appendix}

\section{Experiment Details}
\paragraph{Datasets.}
We evaluate our model on a total of \textbf{seventeen} real-world multivariate time series datasets, covering both \textbf{long-term} and \textbf{short-term} forecasting tasks. The datasets span diverse domains including electricity, weather, traffic, and renewable energy, ensuring the robustness and generality of our approach. Details are as follows:

\begin{itemize}
    \item \textbf{Electricity Transformer Temperature (ETT)}\footnote{\url{https://github.com/zhouhaoyi/ETDataset}}:
    Consists of four subsets. ETTh1 and ETTh2 are recorded hourly, while ETTm1 and ETTm2 are recorded every 15 minutes. The data is collected from two different electric transformers.
    
    \item \textbf{Weather}\footnote{\url{https://www.bgc-jena.mpg.de/wetter}}:
    Contains 21 meteorological indicators collected every 10 minutes in Germany in 2020, from the Weather Station of the Max Planck Biogeochemistry Institute.

    \item \textbf{Traffic}\footnote{\url{https://pems.dot.ca.gov}}:
    Hourly road occupancy rates recorded by 862 sensors on freeways in the San Francisco Bay Area from January 2015 to December 2016.

    \item \textbf{Electricity}\footnote{\url{https://archive.ics.uci.edu/dataset/321/electricityloaddiagrams20112014}}:
    Hourly electricity consumption data from 321 clients spanning the period from 2012 to 2014.

    \item \textbf{Exchange-rate}\footnote{\url{https://github.com/laiguokun/multivariate-time-series-data}}:
    Daily exchange rates of 8 countries from 1990 to 2016.

    \item \textbf{Solar-energy}\footnote{\url{https://www.nrel.gov/grid/solar-power-data.html}}:
    10-minute solar power production records from 137 photovoltaic plants throughout 2006.

    \item \textbf{Wind}\footnote{\url{https://www.kaggle.com/datasets/mubashirrahim/wind-power-generation-data-forecasting}}:
    Wind power and meteorological measurements collected from four real-world wind farm locations. The dataset includes hourly wind speed and power generation data, designed for wind power forecasting tasks.

    \item \textbf{PEMS}\footnote{\url{https://pems.dot.ca.gov/}}:
    The PEMS datasets are designed for short-term forecasting and contain high-frequency traffic measurements at multiple locations. It includes four subsets: PEMS03, PEMS04, PEMS07, and PEMS08, with 5-minute sampling intervals and varying spatial dimensions (number of sensors).
\end{itemize}

\paragraph{Forecasting Settings.}
In accordance with prior works such as \cite{timesnet}, we adopt prediction lengths of $\{96, 192, 336, 720\}$ for long-term forecasting tasks, and a fixed length of $12$ for short-term forecasting. However, we identify a common experimental flaw: the test data loader in these works often sets \texttt{drop\_last=True}, which may discard up to \texttt{batch\_size - 1} samples during evaluation. This issue has been discussed in \cite{qiu2024tfb}, yet persists in recent literature \cite{chen2025simpletm, wang2025fredf}. To ensure fair comparison and full test coverage, we uniformly set \texttt{drop\_last=False} in our experiments.

Table~\ref{tab:datasets} summarizes the statistics of all datasets used.

\begin{table}[th]
  \centering
    \setlength\tabcolsep{3pt}
  \begin{tabular}{c|c|c|c}
    \hline
    Dataset & Dim & Dataset Size & Frequency \\
    \hline
    ETTh1,ETTh2 & 7 & (8545,2881,2881) & Hourly \\
    \hline
    ETTm1,ETTm2 & 7 & (34465,11521,11521) & 15 min \\
    \hline
    Weather & 21 & (36792,5271,10540) & 10 min \\
    \hline
    Traffic & 862 & (12185,1757,3509) & Hourly \\
    \hline
    Electricity & 321 & (18317,2633,5261) & Hourly \\
    \hline
    Exchange-rate & 8 & (5120,665,1422) & Daily \\
    \hline
    Solar-energy & 137 & (36792,5271,10540) & 10 min \\
    \hline
    Wind & 9 & (30468,4286,8665) & Hourly \\
    \hline
    PEMS03 & 358 & (15617,5135,5135) & 5 min \\
    \hline
    PEMS04 & 307 &  (10172,3375,3375) & 5 min \\
    \hline
    PEMS07 & 883 & (16911,5622,5622) & 5 min \\
    \hline
    PEMS08 & 170 & (10690,3548,265) & 5 min \\
    \hline
    \end{tabular}
  \caption{Detailed dataset descriptions.
  Dim denotes dimension, which is the variate number of each dataset.
  Dataset size denotes the total number of time points in (Train, Validation, Test) split.
  Frequency denotes the sampling interval of time points.
    }
    \label{tab:datasets}
\end{table}

\begin{figure}[h]
  \centering
  \includegraphics[width=0.9\linewidth]{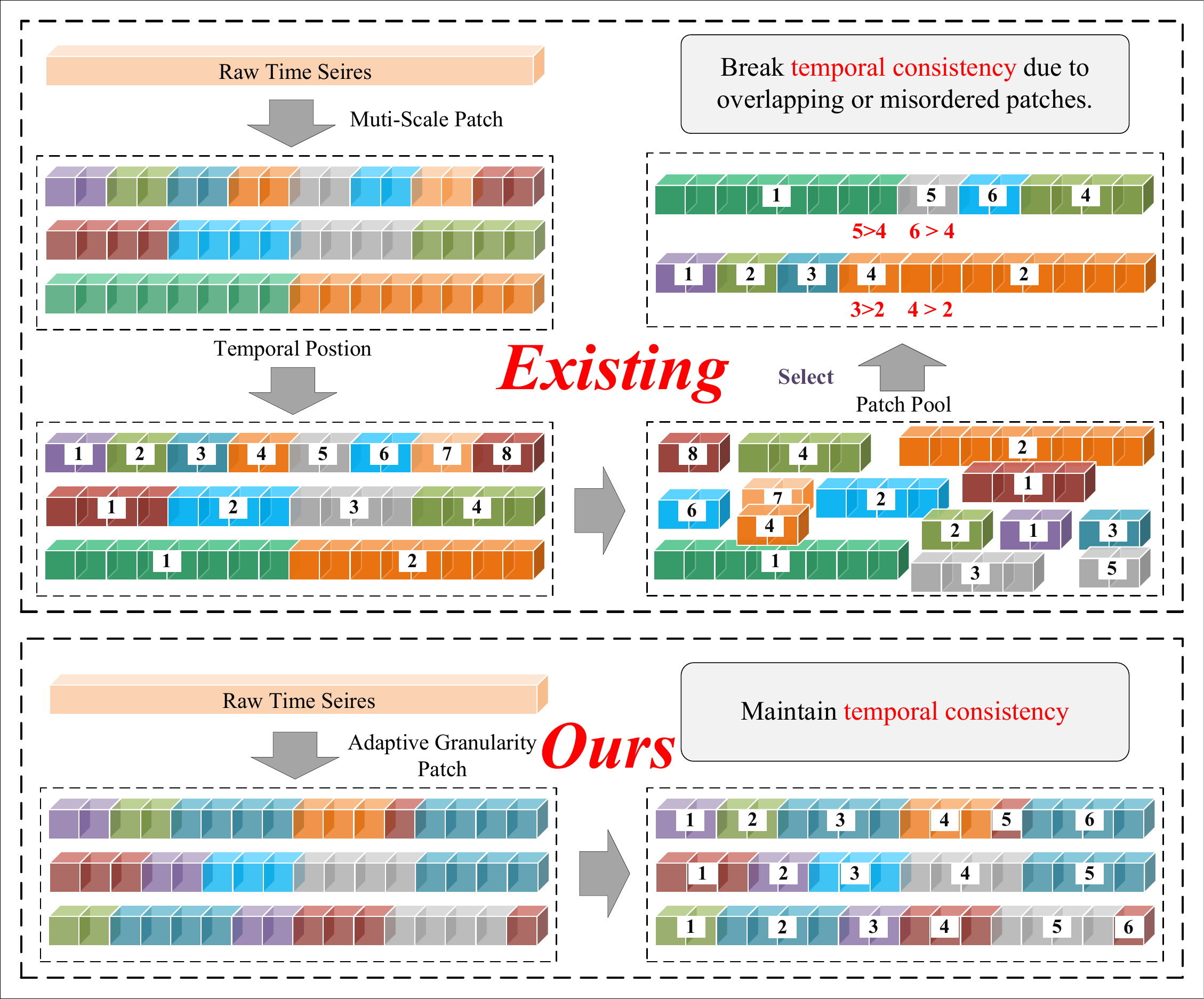}
  \caption{
  Comparison between existing multi-granularity patching methods and our approach.
  Previous methods may disrupt temporal order by inserting coarse and fine patches out of sequence, whereas our method ensures strict chronological alignment with adaptive patch sizes.
  }
  \label{fig:temporal-consistency}
\end{figure}

\section{Construction of Patch Transition Matrix}

To visualize sequential dependencies among temporal patches, we construct a patch-level transition probability matrix based on clustered patch assignments. The procedure is as follows:

\paragraph{Patch extraction.} We collect a large-scale corpus of multivariate time series forecasting datasets (see Appendix~A for full dataset list). For each univariate channel, we normalize the values using min-max scaling and segment the sequence into non-overlapping fixed-length patches of size $P$ (set to 16 by default). This yields a collection of patch vectors $\{\mathbf{x}_i \in \mathbb{R}^P\}_{i=1}^N$ extracted from diverse domains.

\paragraph{Patch quantization.} We apply K-Means clustering to the patch collection to obtain a discrete tokenization of the continuous patch space. Each patch $\mathbf{x}_i$ is assigned to one of $K$ clusters, resulting in a token sequence $\{t_i \in \{1, \dots, K\}\}_{i=1}^N$.

\paragraph{Transition matrix computation.} We define a transition matrix $\mathbf{C} \in \mathbb{N}^{K \times K}$ where $\mathbf{C}_{i,j}$ counts the number of times a patch of cluster $i$ is immediately followed by a patch of cluster $j$ in the tokenized sequence. The row-normalized transition probability matrix $\mathbf{P} \in [0,1]^{K \times K}$ is then computed as:
\[
\mathbf{P}_{i,j} = \frac{\mathbf{C}_{i,j}}{\sum_{j'} \mathbf{C}_{i,j'} + \epsilon}
\]
where $\epsilon$ is a small constant to avoid division by zero.

\begin{figure*}[http]
  % \centerin
  \begin{subfigure}[b]{0.232\textwidth}
    \includegraphics[width=\linewidth]{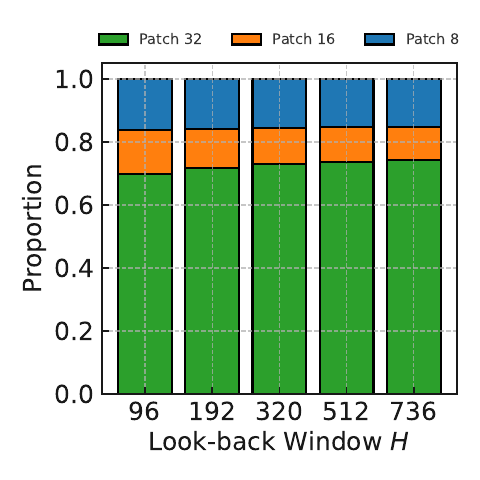}
  \end{subfigure}
  \begin{subfigure}[b]{0.232\textwidth}
    \includegraphics[width=\linewidth]{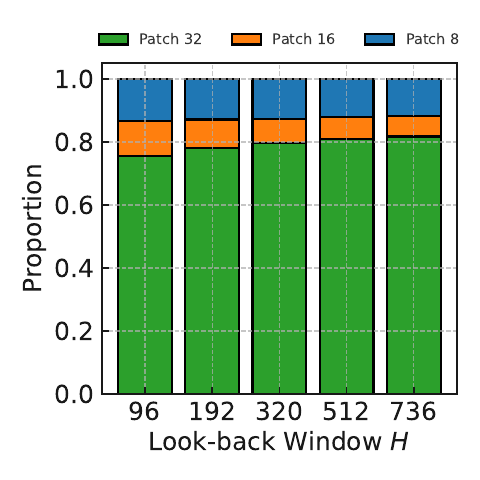}
  \end{subfigure}
  \begin{subfigure}[b]{0.232\textwidth}
    \includegraphics[width=\linewidth]{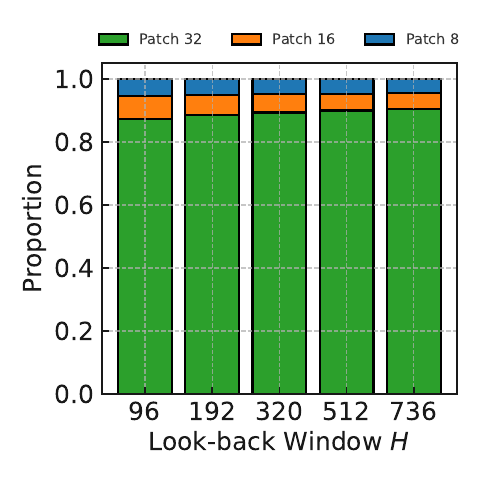}
  \end{subfigure}
  \begin{subfigure}[b]{0.232\textwidth}
    \includegraphics[width=\linewidth]{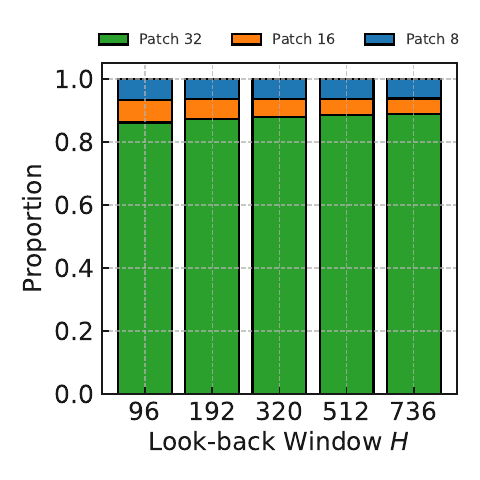}
  \end{subfigure}
  \caption{No budget strategy. Segments of different sizes on different dataset with a predicted length of 192. From left to right the datasets are ETTh1, ETTh2, ETTm1 and ETTm2.}
  \label{fig:segment_size_nobudget}
\end{figure*}

\begin{figure*}[http]
  % \centerin
  \begin{subfigure}[b]{0.232\textwidth}
    \includegraphics[width=\linewidth]{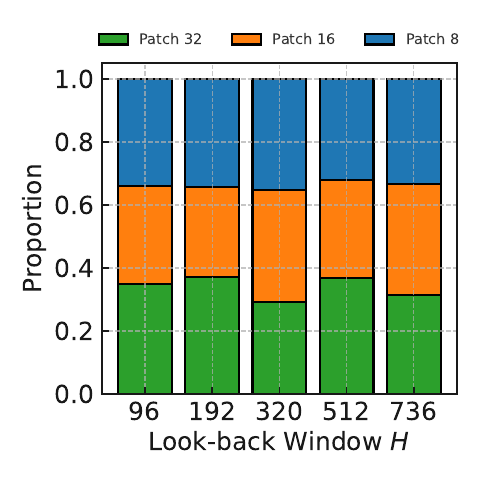}
  \end{subfigure}
  \begin{subfigure}[b]{0.232\textwidth}
    \includegraphics[width=\linewidth]{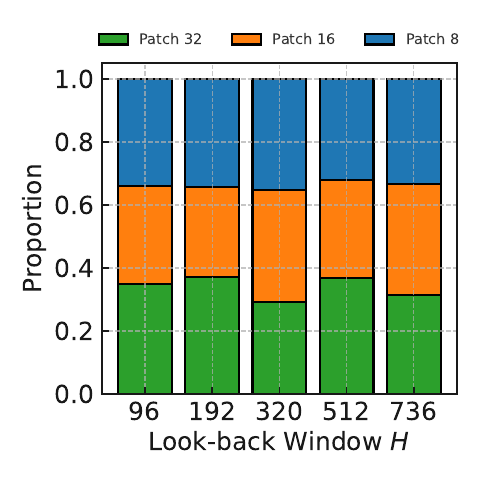}
  \end{subfigure}
  \begin{subfigure}[b]{0.232\textwidth}
    \includegraphics[width=\linewidth]{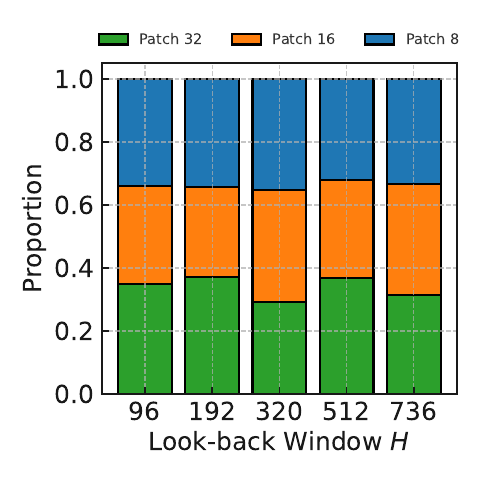}
  \end{subfigure}
  \begin{subfigure}[b]{0.232\textwidth}
    \includegraphics[width=\linewidth]{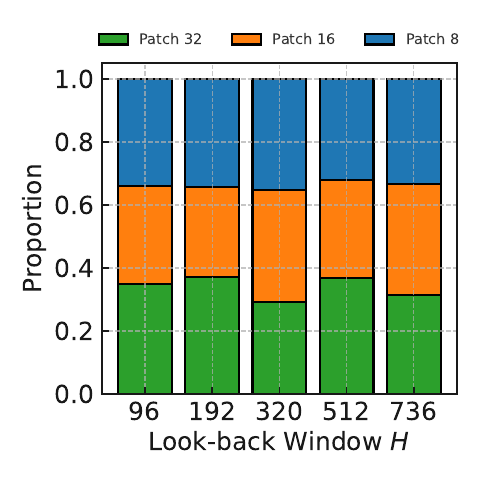}
  \end{subfigure}
  \caption{Segments of different sizes on different dataset with a predicted length of 192. From left to right the datasets are ETTh1, ETTh2, ETTm1 and ETTm2.}
  \label{fig:segment_size}
\end{figure*}

\paragraph{Visualization.} We identify the top-$k$ most frequent patch clusters based on row-wise sums of $\mathbf{C}$ and extract the corresponding $k \times k$ submatrix from $\mathbf{P}$. This submatrix is visualized using a heatmap (Figure~\ref{fig:zipf_silhouette}, where high-probability transitions indicate strong sequential dependencies between dominant patch types.

\paragraph{Implementation.} The full construction and visualization pipeline is implemented in Python using \texttt{scikit-learn}, \texttt{seaborn}, and \texttt{matplotlib}. The source code is included in our supplementary materials. The key parameters are set to $P=16$, $K=100$, and $k=20$ in our default configuration.

\section{Patch Granularity Selection.}

Figure~\ref{fig:segment_size} and Figure~\ref{fig:segment_size_nobudget} compare the distribution of selected patch sizes under different look-back windows across four datasets (ETTh1, ETTh2, ETTm1, and ETTm2), with and without the proposed \textit{budget} strategy, respectively.
Without the budget constraint (Figure~\ref{fig:segment_size_nobudget}), the model exhibits a clear tendency to select coarse-grained patches, with \texttt{Patch 32} dominating across all settings.
In contrast, the budget-aware variant (Figure~\ref{fig:segment_size}) promotes a more balanced selection across all three patch sizes, indicating that the budget loss effectively encourages granularity diversity and prevents overuse of coarse patterns.

\begin{figure}[t!]
  \centering
  \begin{subfigure}[b]{0.48\textwidth}
    \centering
    \includegraphics[width=\linewidth]{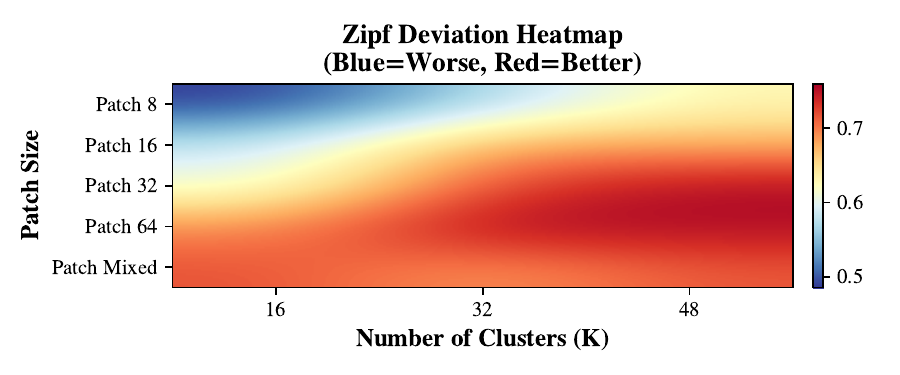}
    \vspace{1mm}
    \includegraphics[width=\linewidth]{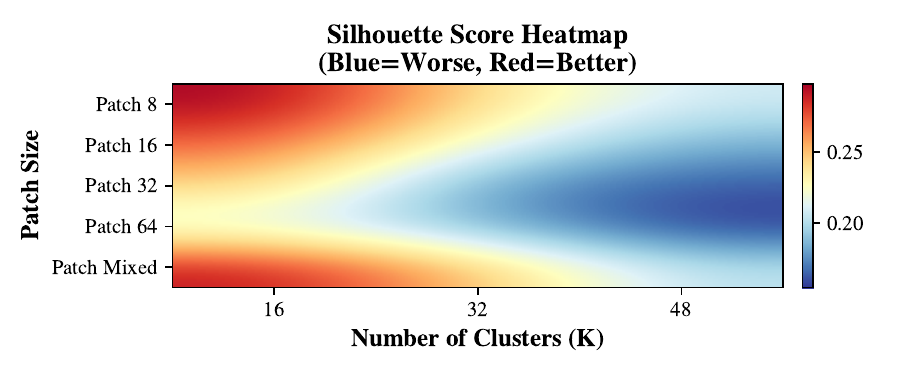}
  \end{subfigure}
  \caption{Zipf deviation and Silhouette score with Patch Mixed.}
  \label{fig:mixer}
  \vspace{-13pt}
\end{figure}

Figure~\ref{fig:mixer} shows the Zipf deviation (top) and Silhouette score (bottom) across different patch granularities (\texttt{8}, \texttt{16}, \texttt{32}, and \texttt{Mixed}) under varying numbers of clusters. These metrics are computed after applying PCA projection and $k$-means clustering on patch representations. We observe a clear trade-off: finer patches (e.g., \texttt{Patch 8}) exhibit better structural clarity (higher Silhouette score) but lower pattern reuse (higher Zipf deviation), while coarser patches (e.g., \texttt{Patch 32}) show better reuse but less cluster cohesion. The \texttt{Mixed} configuration balances both objectives by projecting patches of different granularities into a shared space before clustering, achieving a favorable compromise between diversity and structural consistency.

\paragraph{Quantitative Analysis.} Table~\ref{tab:zipf_silhouette} reports the Zipf deviation and Silhouette scores under various patch sizes and cluster configurations. Finer-grained patches (e.g., size 8) exhibit clearer clustering structure (higher Silhouette scores) but reduced pattern reuse (higher Zipf deviation), while coarser patches (e.g., size 64) show the opposite trend. Notably, the \texttt{Mixed} configuration, which merges patches of size 8, 16, and 32, achieves a favorable balance between diversity and structural clarity, further supporting the effectiveness of patch granularity mixing.

% \paragraph{State Transition Matrix.} Moreover, state transition matrix analysis (Figure \ref{fig:transition}) reveals significant sequential dependencies among patches: certain patches consistently precede specific others with high probability, whereas transitions to other patches occur with negligible probability. This observation highlights strong structured relationships at the patch level. 

\begin{table}[htbp]
\centering
\setlength\tabcolsep{4.5pt}
\scriptsize
\begin{tabular}{c|c|c|c|c}
\toprule
\textbf{Patch Size} & \textbf{Global Patches} & \textbf{Clusters (C)} & \textbf{Silhouette} & \textbf{Zipf Deviation} \\
\midrule
\multirow{3}{*}{8}    & \multirow{3}{*}{152,418} & 16 & 0.2978 & 0.4853 \\
                      &                          & 32 & 0.2392 & 0.5678 \\
                      &                          & 48 & 0.2073 & 0.6301 \\
\midrule
\multirow{3}{*}{16}   & \multirow{3}{*}{76,202}  & 16 & 0.2717 & 0.5804 \\
                      &                          & 32 & 0.2188 & 0.6856 \\
                      &                          & 48 & 0.1845 & 0.6749 \\
\midrule
\multirow{3}{*}{32}   & \multirow{3}{*}{38,094}  & 16 & 0.2454 & 0.6066 \\
                      &                          & 32 & 0.1783 & 0.7204 \\
                      &                          & 48 & 0.1623 & 0.7529 \\
\midrule
\multirow{3}{*}{64}   & \multirow{3}{*}{19,040}  & 16 & 0.2115 & 0.6896 \\
                      &                          & 32 & 0.1862 & 0.7597 \\
                      &                          & 48 & 0.1547 & 0.7530 \\
\midrule
\multirow{3}{*}{Mixed} & \multirow{3}{*}{266,714} & 16 & 0.2913 & 0.7171 \\
                       &                          & 32 & 0.2491 & 0.6844 \\
                       &                          & 48 & 0.2021 & 0.7147 \\
\bottomrule
\end{tabular}
\caption{Zipf deviation and Silhouette score across different patch granularities and cluster sizes. (ETT Datasets)}
\label{tab:zipf_silhouette}
\end{table}

\section{Channel Modeling Variants}
\label{channel}

To support multivariate time series forecasting, our framework incorporates a modular \textit{channel modeling interface} that controls how variables interact before temporal encoding. While the main paper adopts the CI+ strategy by default, we provide several alternative implementations, which can be flexibly selected based on the correlation structure of the dataset. All variants generate an auxiliary token sequence \texttt{extra\_token} to be concatenated with the patch sequence for each variable before encoding.

\paragraph{CI (Channel-Independent):} Each variable is modeled independently by reshaping the input $x \in \mathbb{R}^{B \times T \times C}$ to $[B \cdot C, T]$, and computing patch embeddings separately. No information is shared across variables.

\paragraph{CD-G (Channel-Dependent with Global Token):} The mean signal across all variables is pooled to generate a global context token, which is broadcast to all variable streams and prepended to their patch embeddings.

\paragraph{CD-P (Channel-Dependent Plus).}
An enhanced channel-dependent modeling strategy that incorporates timestamp (time feature) information into inter-channel interactions. Unlike CI or basic CD designs, CD-P explicitly injects temporal encoding (e.g., positional or calendar time) into each variable’s representation, facilitating temporally aware cross-channel modeling.

\paragraph{CD-A (Channel-Dependent with Calendar-Aware Prompts):} Time encoding features (e.g., timestamps, day-of-week) are embedded to generate calendar-aware prompt tokens shared across all variables. This variant is especially effective in periodic domains.

\paragraph{CI+ (Channel-Independent with Prompt Conditioning):} Combines the advantages of CI with prompt-based guidance: each variable retains its independence but is augmented with its own channel token and $K$ global prompt tokens. This hybrid design preserves robustness while introducing controllable task-level bias.

\paragraph{Usage Notes.} All variants are plug-and-play, require minimal code change, and share the same forecasting backbone. In our experiments, CI+ consistently offers a good balance between performance and stability across diverse datasets, especially when variable correlation is moderate or unknown.

\paragraph{Ablation Results.} Table~\ref{tab:channel_ablation} compares five channel modeling strategies on ETTh2 and Weather datasets under a unified backbone and training protocol. CI+ achieves the best or near-best performance on both datasets, validating the benefit of shared channel interaction. Notably, CD-P and CD-A perform more competitively on the Weather dataset, which contains 21 variables (vs. 7 in ETTh2), indicating that richer inter-variable relationships allow channel-dependent strategies to better exploit temporal or global correlations. 

\paragraph{Notes Experiment Setting.} In Table~\ref{tab::long-term}, we search over several channel modeling strategies and report the best-performing configuration, while in Table~\ref{tab::long-term-fair}, we fix the channel-independent setting for all models.

\begin{table}[htbp]
\centering
\setlength\tabcolsep{4.8pt}
\scriptsize
\begin{tabular}{l|cccc|cccc}
\toprule
\multirow{2}{*}{Method} & \multicolumn{4}{c|}{ETTh2} & \multicolumn{4}{c}{Weather} \\
& 96 & 192 & 336 & 720 & 96 & 192 & 336 & 720 \\
\midrule
CI     & 0.295 & 0.380 & \best{0.421} & 0.425 & 0.178 & 0.223 & 0.281 & 0.358 \\
CI+    & \best{0.291} & 0.\best{375} & 0.425 & \best{0.415} & 0.170 & 0.221 & \best{0.276} & \best{0.354} \\
CDG    & 0.299 & 0.395 & 0.427 & 0.444 & 0.174 & 0.222 & 0.280 & 0.357 \\
CDP    & 0.292 & 0.377 & 0.425 & 0.425 & \best{0.169} & 0.222 & 0.279 & 0.357 \\
CDA    & 0.297 & \best{0.375} & 0.429 & 0.424 & 0.173 & \best{0.218} & \best{0.276} & 0.355 \\
\bottomrule
\end{tabular}
\caption{Ablation of channel modeling strategies (MSE) on two representative datasets.}
\label{tab:channel_ablation}
\end{table}

% \begin{table}[htbp]
%   \centering
%   \caption{Patch size selection frequency under adaptive granularity on different datasets and lookback windows.}
%   \label{tab:adaptive_patch_freq}
%   \begin{tabular}{llrrr}
%     \toprule
%     \textbf{Dataset} & \textbf{H} & \textbf{Patch 8} & \textbf{Patch 16} & \textbf{Patch 32} \\
%     \midrule
%     \multirow{5}{*}{ETTh1}
%     & 96  &  9523  &  8111   &  40851 \\
%     & 192 & 18544  & 14671   &  83755 \\
%     & 320 & 30165  & 22564   & 142221 \\
%     & 512 & 47379  & 34493   & 230048 \\
%     & 736 & 67732  & 47305   & 333348 \\
%     \midrule
%     \multirow{5}{*}{ETTh2}
%     & 96  &  7805  &  6559   &  44121 \\
%     & 192 & 15034  & 10565   &  91371 \\
%     & 320 & 24523  & 15145   & 155282 \\
%     & 512 & 38010  & 21071   & 252839 \\
%     & 736 & 53089  & 28747   & 366549 \\
%     \midrule
%     \multirow{5}{*}{ETTm1}
%     & 96  & 12700  & 17571   & 209654 \\
%     & 192 & 24072  & 30315   & 425463 \\
%     & 320 & 38665  & 46402   & 714683 \\
%     & 512 & 59834  & 68301   & 1151465 \\
%     & 736 & 83723  & 92973   & 1662729 \\
%     \midrule
%     \multirow{5}{*}{ETTm2}
%     & 96  & 15851  & 17276   & 206798 \\
%     & 192 & 30753  & 30442   & 418655 \\
%     & 320 & 50510  & 46647   & 702593 \\
%     & 512 & 80340  & 67367   & 1131893 \\
%     & 736 & 113987 & 89256   & 1636182 \\
%     \bottomrule
%   \end{tabular}
% \end{table}

\section{Benchmark Implementation}
\label{app:benchmark}

To ensure a comprehensive and fair evaluation, we have implemented a broad suite of representative time series forecasting models within a unified benchmarking framework. Our benchmark includes 30+ models spanning diverse architectural paradigms, such as Transformer-based, MLP-based, and hybrid designs.

Specifically, our benchmark covers mainstream Transformer-based models including PatchTST, iTransformer, SimpleTM, TimesNet, TimeMixer, TimeMixer++, Informer, Autoformer, FEDformer, Pyraformer, Reformer, ETSformer, Crossformer, TiDE, SCINet, MICN, and SegRNN. These models capture various design philosophies such as decomposition, frequency-domain modeling, inverted attention, and non-stationary transformation.

We also include lightweight and MLP-based approaches such as DLinear, LightTS, TSMixer, PatchMLP, xPatch, and TimeFilter, which emphasize efficiency and simplicity over architectural complexity.

In addition, we integrate several recently proposed hybrid and multi-resolution models, including DUET, PathFormer, FreTS, and WPMixer, to assess the effectiveness of multi-scale temporal modeling strategies.

Finally, our own method, TimeMosaic, is evaluated under the same pipeline. All models share consistent training protocols, data loading procedures, and evaluation metrics. This unified implementation allows for a rigorous and reproducible comparison across a diverse range of baselines.

\paragraph{Discussion on the Unified Benchmark Setting.}
All experimental results reported in this paper are based on our own constructed unified benchmark, where we deliberately avoided over-tuning hyperparameters and fixed them consistently across models. It is therefore understandable that our reported results differ from those in original papers. However, this difference does not imply that the results in other papers are incorrect or unreliable; rather, it simply reflects performance under our more strictly controlled and unified evaluation setting. Currently, there remains debate within the community regarding the fairness of this approach: one perspective argues that unified hyperparameter settings provide a fairer basis for comparison, while another contends that certain models require specific hyperparameter choices to achieve convergence or optimal performance. Investigating this trade-off comprehensively is an important direction for our future research.

\begin{figure*}[t]
  \centering
  \begin{subfigure}[b]{0.195\textwidth}
    \includegraphics[width=\linewidth]{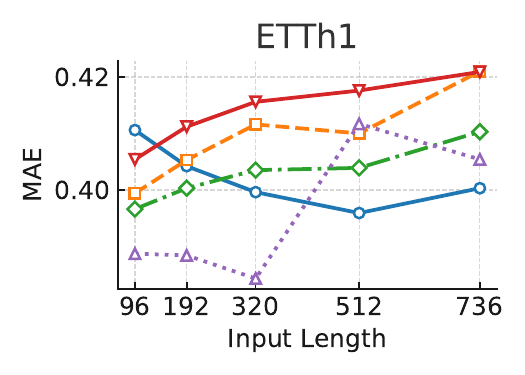}
    % \caption{ETTh1}
  \end{subfigure}
  \hfill
  \begin{subfigure}[b]{0.195\textwidth}
    \includegraphics[width=\linewidth]{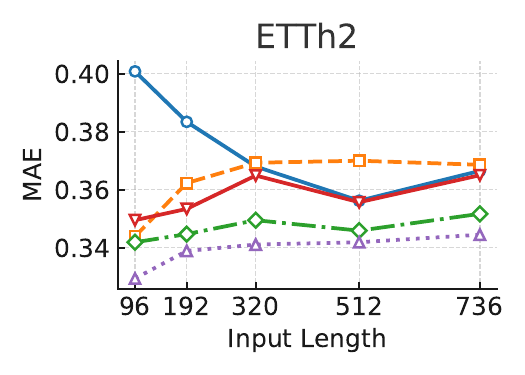}
    % \caption{ETTh2}
  \end{subfigure}
  \hfill
  \begin{subfigure}[b]{0.195\textwidth}
    \includegraphics[width=\linewidth]{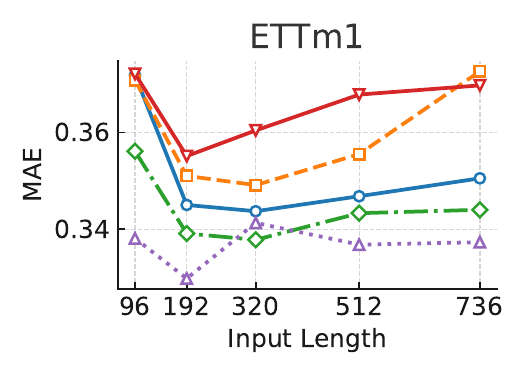}
    % \caption{ETTm1}
  \end{subfigure}
  \hfill
  \begin{subfigure}[b]{0.195\textwidth}
    \includegraphics[width=\linewidth]{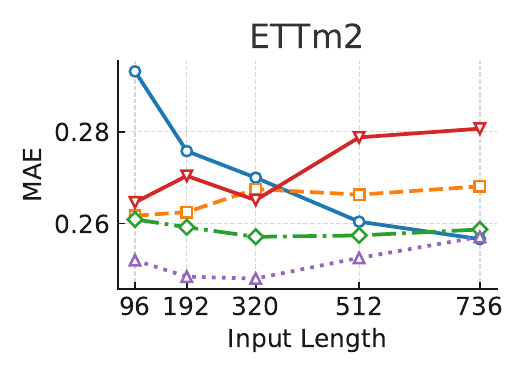}
    % \caption{ETTm2}
  \end{subfigure}
  \hfill
  \begin{subfigure}[b]{0.195\textwidth}
    \includegraphics[width=\linewidth]{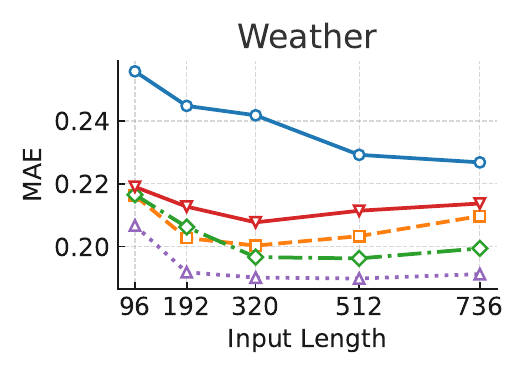}
    % \caption{Weather}
  \end{subfigure}
  \includegraphics[width=0.6\textwidth]{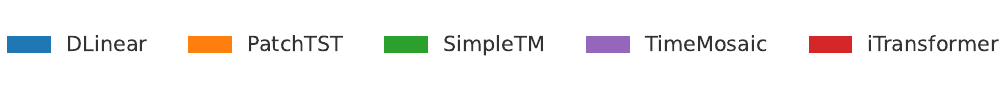}
  \caption{MAE performance across different input lengths for five benchmark datasets. Each subplot compares five models under varying look-back windows.}
  \label{fig:mae_across_datasets}
\end{figure*}

\begin{figure*}[t]
  \centering
  \begin{subfigure}[b]{0.19\textwidth}
    \includegraphics[width=\linewidth]{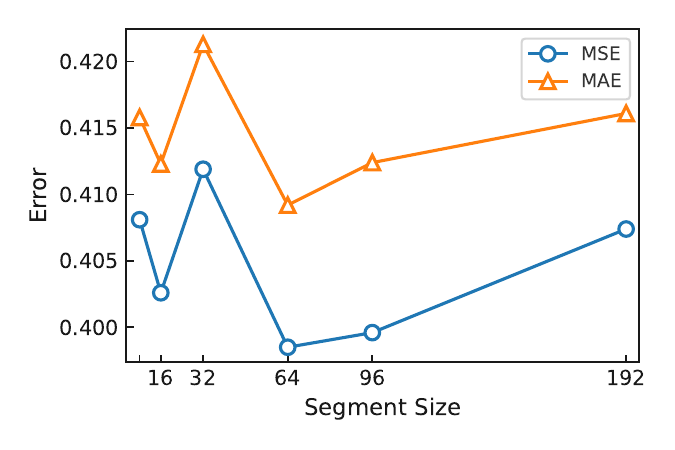}
  \end{subfigure}
  \hfill
  \begin{subfigure}[b]{0.19\textwidth}
    \includegraphics[width=\linewidth]{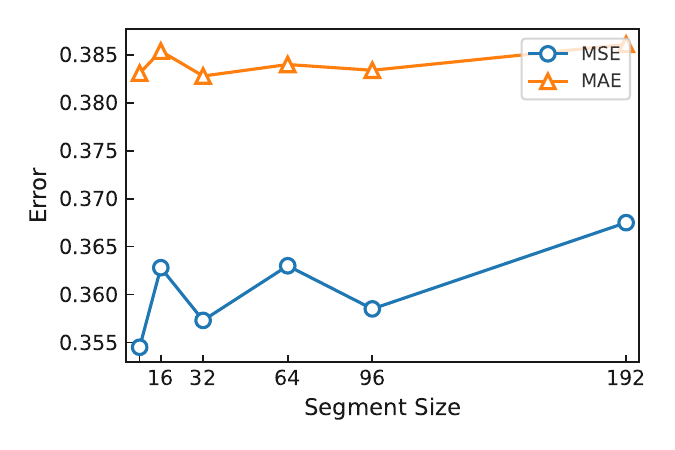}
  \end{subfigure}
  \hfill
  \begin{subfigure}[b]{0.19\textwidth}
    \includegraphics[width=\linewidth]{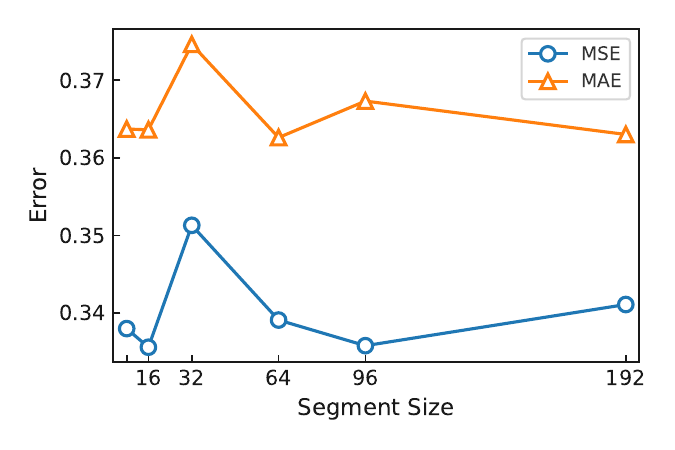}
  \end{subfigure}
  \hfill
  \begin{subfigure}[b]{0.19\textwidth}
    \includegraphics[width=\linewidth]{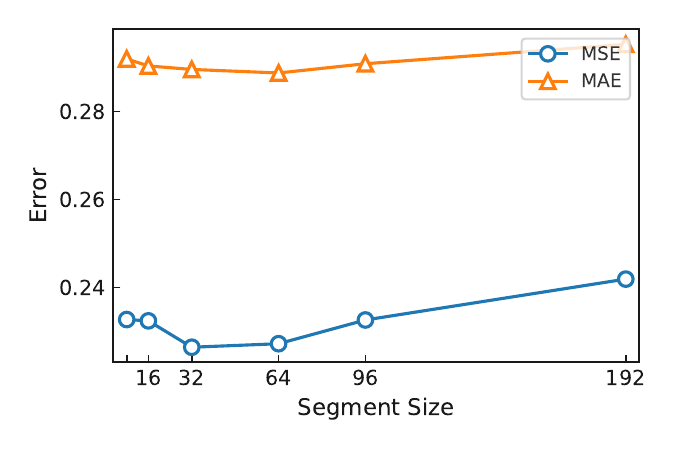}
  \end{subfigure}
  \hfill
  \begin{subfigure}[b]{0.19\textwidth}
    \includegraphics[width=\linewidth]{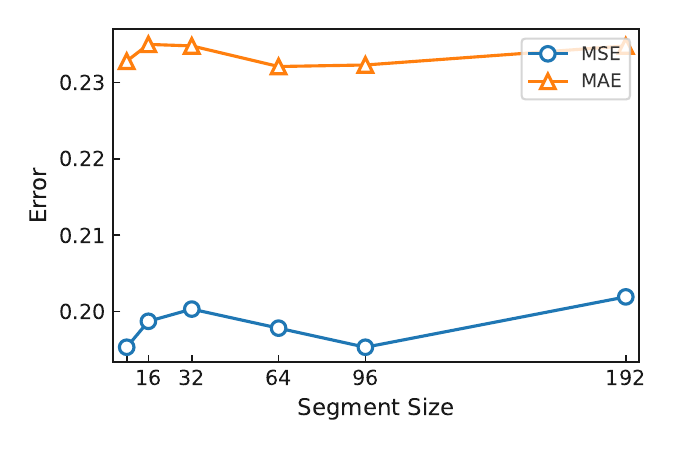}
  \end{subfigure}
  \caption{Segments of different sizes on five datasets with a predicted length of 192. Each figure presents both MSE and MAE performance curves across different segment sizes.}
  \label{fig:segment_mse_mae}
\end{figure*}

\begin{table}[h]
\centering
\scriptsize
\setlength{\tabcolsep}{2pt}
\begin{tabular}{l|ccc|cc}
\hline
\textbf{Dataset} & \textbf{TimeMosaic} & \textbf{SimpleTM} & \textbf{PatchTST} & \textbf{Improve vs ST} & \textbf{Improve vs PT} \\
\hline
ETTm1     & 0.381 & 0.403 & 0.398 & +5.45\% & +4.27\% \\
ETTm2     & 0.314 & 0.325 & 0.323 & +3.38\% & +2.79\% \\
ETTh1     & 0.424 & 0.436 & 0.431 & +2.75\% & +1.63\% \\
ETTh2     & 0.388 & 0.413 & 0.402 & +6.06\% & +3.48\% \\
Weather   & 0.267 & 0.282 & 0.283 & +5.32\% & +5.65\% \\
Traffic   & 0.283 & 0.347 & 0.314 & +18.45\% & +9.87\% \\
Electricity & 0.279 & 0.284 & 0.292 & +1.76\% & +4.45\% \\
Exchange  & 0.403 & 0.411 & 0.401 & +1.95\% & -0.50\% \\
Solar     & 0.270 & 0.333 & 0.397 & +18.92\% & +32.00\% \\
\hline
\textbf{Average} & — & — & — & \textbf{+7.12\%} & \textbf{+7.07\%} \\
\hline
\end{tabular}
\caption{Per-dataset MAE and relative improvement of TimeMosaic over SimpleTM (ST) and PatchTST (PT).}
\label{tab:stat-mae}
\end{table}

\section{Statistical Significance Analysis}

To evaluate the statistical significance of our model's improvements, we compare {TimeMosaic} against two baselines, SimpleTM and PatchTST, on the 9 benchmark datasets using the MAE metric. For each dataset, we report the MAE values and compute the relative improvement of TimeMosaic over each baseline. Furthermore, we conduct Wilcoxon signed-rank tests to assess whether the improvements are statistically significant.

We further perform Wilcoxon signed-rank tests to assess the significance of these improvements. The results confirm that the improvements are statistically significant:

\begin{itemize}
  \item {TimeMosaic vs SimpleTM}: $p = 0.00195$
  \item {TimeMosaic vs PatchTST}: $p = 0.00391$
\end{itemize}

These results demonstrate that TimeMosaic consistently and significantly outperforms strong baselines across datasets under the MAE metric.

\section{Detailed Results}
\label{detres}

In accordance with prior works such as \cite{timesnet}, we adopt the standard prediction lengths of $\{96, 192, 336, 720\}$ for long-term forecasting tasks, and a fixed length of 12 for short-term forecasting. 

\paragraph{Drop Last.} We further identify a common experimental flaw in many recent works: the test dataloader is often configured with \texttt{drop\_last=True}, which can silently discard up to \texttt{batch\_size - 1} samples during evaluation. 
Although this issue has been highlighted in \cite{qiu2024tfb}, it remains present in recent literature \cite{chen2025simpletm, wang2025fredf}. 
To address this, we explicitly set \texttt{drop\_last=False} in all our evaluations, ensuring complete coverage of the test set.

\begin{table*}[htbp]
  \centering
  
  \begin{tabular}{l|cccc|cccc}
    \toprule
    \textbf{Dataset} & $d_{model}$ & $d_{ff}$ & Layers & Heads & Batch Size & Epochs & LR$^*$ & Patience \\
    \midrule
    ECL      & 128 & 256  & 1 & 2 & 16 & 10 & $10^{-4}$ & 3 \\
    Traffic  & 128 & 256  & 1 & 2 & 16 & 10 & $10^{-4}$ & 3 \\
    Solar    & 128 & 256  & 1 & 2 & 16 & 10 & $10^{-4}$ & 3 \\
    ETTh1    & 512 & 2048 & 2 & 8 & 32 & 10 & $10^{-4}$ & 3 \\
    ETTh2    & 512 & 2048 & 2 & 8 & 32 & 10 & $10^{-4}$ & 3 \\
    ETTm1    & 512 & 2048 & 2 & 8 & 32 & 10 & $10^{-4}$ & 3 \\
    ETTm2    & 512 & 2048 & 2 & 8 & 32 & 10 & $10^{-4}$ & 3 \\
    Weather  & 512 & 2048 & 2 & 8 & 32 & 10 & $10^{-4}$ & 3 \\
    Exchange & 512 & 2048 & 2 & 8 & 32 & 10 & $10^{-4}$ & 3 \\
    Wind     & 512 & 2048 & 2 & 8 & 32 & 10 & $10^{-4}$ & 3 \\
    \bottomrule
  \end{tabular}
  \caption{Experiment configuration of TimeMosaic in Table~\ref{tab::long-term-fair}. All experiments use the Adam optimizer with a learning rate of $10^{-3}$ and early stopping (patience = 3).}
  \label{tab:config}
\end{table*}

\paragraph{Over-tuning hyperparameters.} Moreover, we observe that several recent methods rely on overly aggressive hyperparameter tuning strategies, such as assigning distinct learning rates to each forecasting length 
(e.g., $lr = \{2.42\!\times\!10^{-4}, 2.01\!\times\!10^{-4}, 1.32\!\times\!10^{-4}, 2.39\!\times\!10^{-4}\}$ for $T=\{96,192,336,720\}$) or customizing hyperparameters for each dataset. 
Such practices inflate performance but undermine comparability. 
In contrast, we use consistent settings across horizons and datasets unless explicitly stated, thereby ensuring a level playing field for all methods evaluated.

% \paragraph{Details Results.} Table~\ref{tab:apt_vs_simpletm} shows the predictive performance of TimeMosaic and SimpleTM on the Wind dataset. Table~\ref{tab::long-term} presents the complete results of long-term forecasting. Table~\ref{tab:zero-shot-forecasting} shows the results of zero-shot prediction.

\paragraph{Details Results.} Table~\ref{tab::long-term} and Table~\ref{tab::long-term-fair} presents the complete results of long-term forecasting. Table~\ref{tab:zero-shot-forecasting} shows the results of zero-shot prediction.As shown in Table~\ref{tab:patch-mode-ablation}, to ensure fairness and robust performance across datasets, we choose the patch granularity \texttt{[8, 16, 32]} as a representative configuration for all subsequent experiments.

\paragraph{Lookback Windows.} Our model is designed to leverage longer lookback windows more effectively, which can potentially lead to even better forecasting performance. However, due to time and resource constraints, we have not conducted a comprehensive evaluation across extended lookback lengths. We plan to include this analysis in future work to further validate the robustness of our approach. We observed that the Exchange dataset is highly sensitive to the length of the lookback window, and its forecasting performance drops substantially as the window increases across all model. To ensure fairness and representativeness in the long-window comparison (Table~\ref{tab::long-term-fair}), we report results on the Wind1 dataset.

\paragraph{Short-term Forecasting.} To validate the effectiveness of our approach in short-term forecasting, we selected a representative subset of baseline models for comparison, as shown in Table~\ref{tab:experiments_short}. Under unified experimental settings, our method achieves consistently competitive performance across multiple traffic datasets. While we do not claim absolute superiority, the results indicate that our design is capable of delivering robust short-term predictions without significant degradation.

\paragraph{ERROR BARS.} In this paper, we repeat  the experiments three times. Here we report the standard deviation of our model, as well as the statistical significance test in Table \ref{tab:confidence}.

\paragraph{Experiment Details.} All experiments were conducted on four NVIDIA A800 GPUs and implemented using PyTorch. We ran each experiment three times and report the average results. The initial learning rate was set to $10^{-3}$ and optimized using the Adam optimizer with MAE loss. We applied an early stopping strategy with a patience of 3 epochs. The batch size was set to 16 for smaller datasets such as ECL, Traffic, and Solar, and 32 for the others. To ensure consistency across datasets, we fixed the training epochs to 10. Model configurations, including the number of layers, hidden dimensions, and attention heads, were selected based on dataset scale and complexity to balance performance and efficiency. Detailed model and training settings are summarized in Table~\ref{tab:config} about Table~\ref{tab::long-term-fair}.

\paragraph{Effect of Lookback Window.} \label{lookback}
To investigate how models perform under varying input lengths, we evaluate five representative methods on six datasets with a fixed prediction length of 96 and increasing lookback windows ranging from 96 to 736. As illustrated in Figure~\ref{fig:mae_across_datasets}, the performance of each model exhibits distinct trends as the input window lengthens. Traditional models like DLinear generally benefit from longer inputs, showing steady MAE improvements on datasets such as ETTh2 and Weather. In contrast, token-heavy transformer models like PatchTST and iTransformer often degrade in performance with longer inputs, likely due to the increased difficulty in extracting relevant temporal patterns from excessive tokens. SimpleTM shows moderate stability but still suffers mild degradation on some datasets.

Notably, TimeMosaic consistently achieves the lowest MAE across all datasets and most input lengths, demonstrating superior robustness to input size variation. For example, on the ETTh2 dataset, TimeMosaic maintains low and stable MAE across all input lengths (from 0.3294 to 0.3445), while PatchTST and iTransformer show increasing errors. This suggests that our model's adaptive granularity mechanism and segment-wise decoding enable it to effectively utilize longer contexts without overfitting or introducing noise, supporting both scalability and stability.

\paragraph{Experimental Settings and Fairness Considerations.} 
To ensure a fair and reproducible comparison across models, we adopt a unified evaluation setting in Table~\ref{tab::long-term-fair}, which is more rigorous than the setup used in Table~\ref{tab::long-term}. Specifically, Table~\ref{tab::long-term-fair} employs an {extended lookback window}, enabling models such as {DUET} and {TimeFilter} to better realize their potential. Furthermore, to avoid test-set-driven tuning, we strictly fix all major hyperparameters, including \texttt{d\_model}, \texttt{d\_ff}, the number of layers, attention heads, learning rate, training epochs, patience, and batch size. We summarize the detailed hyperparameter settings in Table~\ref{tab:hyperparameters}.

\begin{table*}[h]
\centering
\small
\renewcommand{\arraystretch}{1.2}
\setlength{\tabcolsep}{8pt}
\begin{tabular}{lccccccc}
\toprule
\textbf{Datasets} & \texttt{d\_model} & \texttt{d\_ff} & Layers & Heads & LR & Epochs (Patience) & Batch Size \\ 
\midrule
ETT, Weather, Exchange & 512 & 2048 & 2 & 8 & 0.0001 & 10 (3) & 32 \\[2pt]
Traffic, Electricity, Solar & 128 & 256 & 1 & 2 & 0.0001 & 10 (3) & 16 \\[2pt]
Traffic (TimeFilter only) & 128 & 256 & 1 & 2 & 0.0001 & 10 (3) & 2* \\ 
\bottomrule
\end{tabular}
\caption{Unified hyperparameter settings used for experiments in Table~\ref{tab::long-term-fair}. *Batch size is set to 2 specifically for TimeFilter on Traffic due to GPU memory constraints (59,551\,MiB / 81,920\,MiB).}
\label{tab:hyperparameters}
\end{table*}

Note that we excluded PathFormer from Table~\ref{tab::long-term-fair} due to its excessive GPU memory requirements when extending the lookback window, which made its evaluation computationally infeasible.

\begin{figure}[t]
  \centering
  \includegraphics[width=0.6\linewidth]{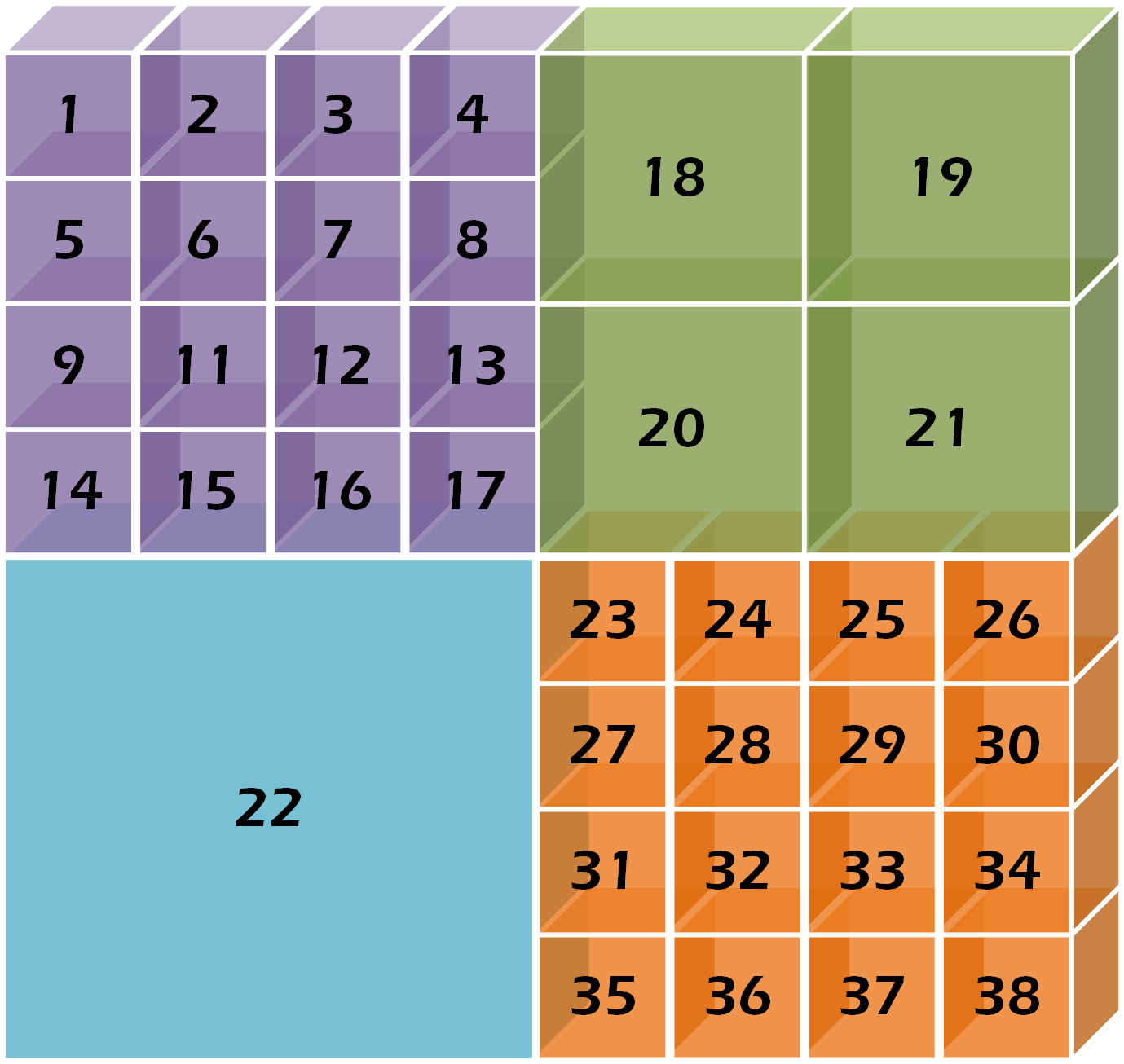}
  \caption{Overview of position embedding adopted in ours.}
  \label{fig:pos}
\end{figure}

\paragraph{Ablation on Normalization, Position Encoding, and Budget Loss Settings.} Table~\ref{tab:windpos_updated} presents a comprehensive comparison of ablation studies on the Wind1–Wind4 datasets, focusing on three key model components: normalization method, position encoding strategy, and budget loss configuration. Specifically, we compare the following variants:

\begin{itemize}
    \item \textbf{TimeMosaic}: The standard model, using z-score normalization, sinusoidal position encoding, and a default budget loss weight of 0.01.
    \item \textbf{Revin}: Replaces z-score normalization with RevIN~\cite{kim2021reversible}. We note that z-score normalization is also used for all experiments in Table~\ref{tab::long-term}.
    \item \textbf{LearnPos}: Substitutes sinusoidal encoding with learnable position embeddings. For adaptive patching, positional indices are assigned sequentially starting from each region (see visualization in Fig.~\ref{fig:pos}).
    \item \textbf{NoPos}: Removes position encoding entirely.
    \item \textbf{Budget$_{0.001}$}: Sets budget loss weight to 0.001.
    \item \textbf{NoBudget}: Removes budget loss term.
\end{itemize}

All variants are evaluated on the Wind1–Wind4 datasets, using the same backbone architecture.

As shown in Table~\ref{tab:normalization-comparison}, we compare z-score and RevIN normalization across eight benchmark datasets and four forecasting horizons. We observe that RevIN achieves lower MSE or MAE in the majority of cases, especially on datasets such as ETTh1, Weather, and Solar-Energy. However, z-score normalization still performs competitively on datasets like ETTm1 and Traffic. These results suggest that the effectiveness of normalization methods is data-dependent, and RevIN tends to offer better robustness on datasets with complex or shifting distributions.

\paragraph{Analysis.} Across all settings, TimeMosaic and its ablations demonstrate robust performance. Notably, switching from z-score to RevIN normalization provides a slight and consistent improvement in MSE and MAE across most datasets and horizons, indicating RevIN’s suitability for wind series with non-stationary statistics.

Replacing sinusoidal with learnable position encodings (LearnPos) yields comparable or slightly improved results, confirming the effectiveness of data-driven position embeddings when patch boundaries are not fixed. In contrast, removing positional encoding (NoPos) generally leads to marginally worse performance, highlighting the importance of explicit temporal cues, especially under adaptive patching.

Regarding budget loss, decreasing its weight from 0.01 to 0.001 (Budget$_{0.001}$) or removing it altogether (NoBudget) results in nearly identical outcomes, suggesting that the model is relatively insensitive to this hyperparameter within the tested range.

Overall, these results validate the robustness of TimeMosaic to different normalization, positional encoding, and budget loss configurations.

\section{Detailed Analysis of Forecasting Results.}
Table~\ref{tab::long-term-fair} presents an extensive evaluation of our proposed method {TimeMosaic} compared to eleven state-of-the-art baseline models across eight widely-used multivariate time series datasets (ETTm1, ETTm2, ETTh1, ETTh2, Weather, Electricity, Traffic, Solar-Energy, Wind) with four prediction horizons (96, 192, 336, and 720). Our analysis highlights TimeMosaic's consistent superiority in forecasting performance, and we summarize key insights below.

\textbf{1) Comprehensive Performance Advantage.}
TimeMosaic distinctly outperforms all competing methods in terms of the number of times it achieves top performance, leading in {32 cases}—significantly ahead of other models such as TimeFilter (22 cases) and DUET (16 cases). This demonstrates TimeMosaic’s broad effectiveness and robustness in diverse forecasting contexts.

\textbf{2) Dataset-specific Observations.}
On {ETTm1}, TimeMosaic shows a clear advantage at the longest horizon (720), achieving the best MSE/MAE and the lowest average MAE across all horizons.
On \textit{ETTm2}, it consistently ranks in the top two, with particularly strong long-term forecasting.
On \textit{ETTh1} and \textit{ETTh2}, it outperforms others in average metrics and maintains strong short-horizon accuracy.
On \textit{Weather} and \textit{Solar-Energy}, TimeMosaic performs best or second-best in most cases, validating its generalization to nonlinear and periodic signals.
On \textit{Electricity} and \textit{Traffic}, it remains competitive with TimeFilter, ranking second-best in MAE consistently.
On \textit{Wind}, despite LightTS's lead, TimeMosaic still ranks second-best on average, demonstrating robust behavior on highly volatile data.

\textbf{3) Horizon-wise Analysis.}
TimeMosaic maintains strong performance across all horizons. Its capability is especially notable at longer horizons (e.g., ETTm1 at 720 with MAE = 0.413), highlighting its effectiveness in reducing long-range error accumulation.

\textbf{4) Methodological Insights.}
The superior performance of TimeMosaic stems from its \textit{adaptive patch embedding} and \textit{segment-wise prompt tuning}, which jointly address temporal heterogeneity in both input and output. Unlike fixed-patch methods (e.g., PatchTST) or simple baselines (e.g., DLinear), our design enables finer modeling where needed and coarser abstraction elsewhere, achieving high accuracy with efficiency.

Overall, the results in Table~\ref{tab::long-term-fair} show that \textbf{TimeMosaic} sets a new state-of-the-art in long-term multivariate forecasting, demonstrating consistent, robust, and interpretable performance across diverse datasets and conditions.

\begin{table*}[h]
\centering
\scriptsize
\begin{tabular}{c|cc|cc|cc|cc|cc|cc|cc}
\toprule
\multirow{2}{*}{Patch granularity} 
& \multicolumn{2}{c|}{ETTh1} 
& \multicolumn{2}{c|}{ETTh2} 
& \multicolumn{2}{c|}{ETTm1} 
& \multicolumn{2}{c|}{ETTm2} 
& \multicolumn{2}{c|}{Exchange} 
& \multicolumn{2}{c|}{Weather}
& \multicolumn{2}{c}{Average} \\
& MSE & MAE & MSE & MAE & MSE & MAE & MSE & MAE & MSE & MAE & MSE & MAE & MSE & MAE \\
\midrule
\texttt{[8, 16]}         & 0.363 & 0.389 & 0.296 & 0.341 & 0.295 & 0.332 & 0.168 & 0.248 & 0.106 & 0.230 & 0.155 & 0.190 & 0.231 & 0.288 \\
\texttt{[4, 8, 16, 32]}     & 0.370 & 0.391 & 0.297 & 0.342 & 0.292 & 0.332 & 0.171 & 0.249 & 0.099 & 0.222 & 0.156 & 0.191 & 0.232 & 0.290 \\
\texttt{[8, 16, 32]}     & 0.359 & 0.387 & 0.307 & 0.346 & 0.300 & 0.339 & 0.169 & 0.248 & 0.105 & 0.230 & 0.154 & 0.190 & 0.231 & 0.289 \\
\texttt{[8, 16, 32, 64]} & 0.359 & 0.384 & 0.298 & 0.341 & 0.295 & 0.336 & 0.167 & 0.246 & \textbf{0.093} & \textbf{0.217} & 0.153 & 0.191 & \textbf{0.228} & \textbf{0.286} \\
\texttt{[8, 16, 64]}     & 0.363 & 0.386 & \textbf{0.295} & \textbf{0.340} & 0.294 & 0.337 & 0.171 & 0.250 & 0.105 & 0.230 & 0.154 & 0.192 & 0.230 & 0.289 \\
\texttt{[8, 32]}         & \textbf{0.355} & \textbf{0.382} & 0.298 & 0.341 & \textbf{0.289} & \textbf{0.332} & 0.173 & 0.251 & 0.097 & 0.219 & \textbf{0.153} & \textbf{0.189} & \textbf{0.228} & \textbf{0.286} \\
\texttt{[8, 32, 64]}     & 0.369 & 0.390 & 0.306 & 0.345 & 0.303 & 0.336 & \textbf{0.166} & \textbf{0.245} & 0.108 & 0.231 & 0.155 & 0.191 & 0.235 & 0.289 \\
\bottomrule
\end{tabular}
\caption{Performance comparison of different patch granularities across six datasets ($L=320$, $H = 96$).}
\label{tab:patch-mode-ablation}
\end{table*}

\begin{table}[ht]
\centering
\scriptsize
\setlength{\tabcolsep}{2pt}
\begin{tabular}{c|cc|cc|cc}
\toprule
\multirow{2}{*}{Dataset} 
& \multicolumn{2}{c|}{Baseline} 
& \multicolumn{2}{c|}{+APE} 
& \multicolumn{2}{c}{+APE + Prompt} \\
& MSE & MAE & MSE & MAE & MSE & MAE \\
\midrule
ETTh1     & 0.403 & 0.423 & 0.361 & 0.387 & 0.362 & 0.387 \\
ETTh2     & 0.324 & 0.374 & 0.297 & 0.342 & 0.300 & 0.342 \\
ETTm1     & 0.321 & 0.361 & 0.298 & 0.335 & 0.304 & 0.344 \\
ETTm2     & 0.181 & 0.270 & 0.172 & 0.251 & 0.164 & 0.247 \\
Weather   & 0.155 & 0.204 & 0.160 & 0.196 & 0.144 & 0.189 \\
\midrule
Average   & 0.277 & 0.326 & 0.258 & 0.302 & 0.254 & 0.301 \\
\bottomrule
\end{tabular}
\caption{Ablation results on five datasets with different module combinations. The length of input is 96.}
\label{tab:ablation}
\end{table}

\begin{table*}[h!]
% \scriptsize
\footnotesize
    \centering
    \setlength\tabcolsep{3pt}
\begin{tabular}{llcccccc}
\toprule
\textbf{Dataset} & \textbf{Model} & \textbf{MSE} & \textbf{MAE} & \textbf{Total Parameter} & \textbf{Inference Time (s)} & \textbf{GPU Memory Footprint (MB)} &  \textbf{Peak Mem (MB)} \\
\midrule
\multirow{5}{*}{ETTh1}
& TimeMosaic & \textbf{0.413} & \textbf{0.428}  & \textbf{365,825} & 0.059504 & \textbf{21.78} & {161.29} \\
& TimeFilter   & 0.425 & 0.438  & 1,830,640 & 0.014385 & 81.95 & 240.01 \\
& SimpleTM & 0.429 & 0.443  & 9,893,368 & 0.017082 & 94.06 & 872.34 \\
& iTransformer & 0.437 & 0.450  & 3,588,432 & \textbf{0.002159} & 45.95 & \textbf{53.66} \\
& PatchTST & 0.423 & 0.437  & 2,953,424 & 0.002301 & 43.54 & 155.70 \\
\midrule
\multirow{5}{*}{ETTh2}
& TimeMosaic & 0.370 & \textbf{0.397}  & 2,367,329 & 0.010097 & 38.73 & 121.91 \\
& TimeFilter & \textbf{0.368} & 0.411  & 176,460 & 0.006502 & 26.61 & 175.04 \\
& SimpleTM & 0.381 & 0.413  & 14,216 & 0.003503 & 18.30 & 146.40 \\
& iTransformer & 0.385 & 0.415  & \textbf{14,160} & \textbf{0.002070} & \textbf{18.29} & \textbf{19.75} \\
& PatchTST & 0.384 & 0.413  & 232,064 & 0.005510 & 20.29 & 99.58 \\
\midrule
\multirow{5}{*}{ETTm1}
& TimeMosaic & \textbf{0.356} & \textbf{0.374}  & 227,169 & 0.033050 & 20.28 & 155.07 \\
& TimeFilter & 0.362 & 0.384 & 111,444 & 0.005413 & 25.37 & 83.53 \\
& SimpleTM & 0.362 & 0.387  & 3,490,184 & 0.005095 & 44.81 & 822.48 \\
& iTransformer & 0.370 & 0.394  & \textbf{12,112} & \textbf{0.003450} & \textbf{18.01} & \textbf{19.37} \\
& PatchTST & 0.362 & 0.389  & 139,488 & 0.003512 & 19.31 & 40.52 \\
\midrule
\multirow{5}{*}{ETTm2}
& TimeMosaic & \textbf{0.269} & \textbf{0.318}  & 365,825 & 0.031740 & 21.78 & 161.29 \\
& TimeFilter & 0.272 & 0.329 & 118,076 & 0.010274 & 25.44 & 83.88 \\
& SimpleTM & 0.273 & 0.328  & 1,076,328 & 0.079104 & 26.40 & 76.93 \\
    & iTransformer & 0.287 & 0.340  & \textbf{17,232} & 0.003550 & \textbf{18.71} & \textbf{20.57} \\
& PatchTST & 0.272 & 0.333  & 347,968 & \textbf{0.002239} & 21.54 & 109.19 \\
\midrule
\multirow{5}{*}{Weather}
& TimeMosaic & 0.240 & \textbf{0.267}  & 372,413 & 0.055469 & 26.30 & 342.41 \\
& TimeFilter & \textbf{0.237} & 0.278 & 179,148 & 0.032229 & 35.48 & 1363.94 \\
& SimpleTM & 0.238 & 0.276  & \textbf{17,400} & 0.008303 & \textbf{22.97} & \textbf{26.05} \\
& iTransformer & 0.247 & 0.284  & 1,100,624 & 0.006658 & 31.22 & 36.37 \\
& PatchTST & 0.243 & 0.279  & 347,968 & \textbf{0.002607} & 25.80 & 288.75 \\
\bottomrule
\end{tabular}
\caption{Comparison of model performance and resource utilization across different datasets. Metrics include Mean Squared Error (MSE), Mean Absolute Error(MAE), total parameter count, inference time (seconds), GPU memory footprint (MB), and peak memory usage (MB), the predicted length is fiexed 336.}
\label{tab:model_efficiency}
\end{table*}

\section{Hyperparameter Search Protocol}
\label{sec:hparam_search}

To ensure a fair comparison across models, we adopt a unified grid-search strategy on standard long-term forecasting benchmarks (ETTh1, ETTh2, ETTm1, ETTm2, Weather). The search space covers input sequence lengths $\{96,192,320,512\}$, prediction horizons $\{96,192,336,720\}$, model dimensions $\{16,128,512\}$ with feed-forward size $4\times d_{\text{model}}$, encoder depths $\{1,3,5\}$, training epochs $\{10,50,100\}$, and learning rates $\{10^{-5},10^{-4},10^{-3},10^{-2},0.05\}$. Other hyperparameters (batch size, number of heads, decoder layers, etc.) follow common defaults to keep the search tractable. This design yields $2160$ configurations per dataset, leading to \textbf{10,800} candidate runs per model over the five datasets considered. Results are reported based on the best validation performance under this unified search space.  

For fairness, all models are treated identically. When models involve auxiliary loss terms in addition to the main forecasting objective, we fix the auxiliary loss weight to $0.001$, including for TimeMosaic. Model-specific hyperparameters such as the number and size of granularities in TimeMosaic, or the \texttt{topk} parameter in other architectures, are kept at their default values rather than tuned. Likewise, dropout is uniformly disabled across all methods to avoid providing implicit advantages to any single model. Finally, certain models such as DLinear do not make use of \texttt{d\_model} or feed-forward dimensions; in these cases, the corresponding search dimensions are skipped.

\section{Comparison with TSFMs}
\label{appendix:tsfm}

To further evaluate the scalability and generalization ability of TimeMosaic, we conduct large-scale experiments under the same settings as Table~21, comparing against representative Time Series Foundation Models (TSFMs). Specifically, TimeMosaic is trained with an input length of 512 and an output horizon of 720 on two NVIDIA V100 GPUs, requiring approximately 40 hours of training. The forecasting segment length is set to 16, with \texttt{train\_epochs}=1, batch size of 256, a single Transformer layer, and 8 attention heads. The prompt embedding dimension is 8, and the adaptive patch search space is defined as \texttt{[8, 16, 32, 64]}. We adopt the BLAST dataset~\cite{blast} for training. Under this configuration, the total number of trainable parameters is 27,259,924.

BLAST (BaLAnced Sampling Time series corpus) is a large-scale dataset specifically constructed for pre-training universal time series models. It consists of 321 billion observations, mainly from CMIP6~\cite{CMIP6} (32.5\%), ERA5~\cite{ERA5} (30.0\%), WeatherBench~\cite{weatherbenchmark} (25.7\%), Buildings\_900K~\cite{emami2023buildingsbench} (6.9\%), and over 300 additional public datasets~\cite{blast} (4.9\%), covering diverse domains such as climate, meteorology, and energy. To ensure the independence of downstream evaluation, commonly used benchmark datasets (e.g., ETTh/ETTm, Weather, GlobalTemp) are deliberately excluded. In our experiments, we adopt random sampling on BLAST to generate training data, ensuring broad coverage of diverse patterns.

Table~\ref{tab:blast} reports the zero-shot forecasting results of TimeMosaic across six datasets. The results demonstrate that, when trained on large-scale data, TimeMosaic achieves performance comparable to state-of-the-art TSFMs, highlighting the structural advantages of our design. In particular, the adaptive patch embedding and segment-wise decoding modules enable effective utilization of long historical contexts and robust adaptation to diverse prediction horizons, which is crucial for foundation-model-level generalization.

\begin{table}[h]
\centering
\footnotesize
\caption{Comparison of model size and inference efficiency. TimeMosaic maintains a moderate parameter scale and inference latency compared to existing TSFMs.}
\label{tab:params-efficiency}
\begin{tabular}{lcc}
\toprule
\textbf{Model} & \textbf{Params} & \textbf{Inference Time (s)} \\
\midrule
TimeMoe-50M    & 113M  & 5.725 \\
TimeMoe-200M   & 453M  & 10.813 \\
Moirai-Small   & 14M   & 0.350 \\
Moirai-Base    & 91M   & 0.483 \\
Moirai-Large   & 311M  & 0.894 \\
Chronos-Small  & 48M   & 0.256 \\
Chronos-Base   & 205M  & 0.409 \\
\midrule
\textbf{TimeMosaic (Ours)} & 27M & 0.068 \\
\bottomrule
\end{tabular}
\end{table}

As shown in Table~\ref{tab:params-efficiency}, TimeMosaic maintains a balanced trade-off between model scale and inference efficiency. Its parameter count (27M) is larger than extremely compact baselines such as Moirai-Small (14M), yet smaller than medium- and large-scale TSFMs (e.g., Chronos-Small at 48M, Chronos-Base at 205M, and TimeMoe-200M at 453M). More importantly, TimeMosaic achieves the fastest inference speed among all compared models (0.068s), substantially outperforming both small baselines (e.g., Moirai-Small at 0.350s, Chronos-Small at 0.256s) and heavyweight TSFMs. These results confirm that TimeMosaic achieves a highly favorable balance between scalability and efficiency, making it well-suited for both research and practical deployment.

\section{Limitations and Future work}

While TimeMosaic achieves state-of-the-art forecasting performance, several limitations provide opportunities for future research. Firstly, TimeMosaic currently utilizes a moderate parameter scale, which restricts its capability to capture complex long-range dependencies. Future research could scale TimeMosaic into a foundational model via large-scale pretraining, potentially enhancing its temporal representation and cross-domain generalization.Secondly, TimeMosaic employs fixed prediction horizons, limiting its adaptability. Future work should develop dynamic, adaptive prediction horizons responsive to real-time data characteristics, improving practicality in streaming scenarios. Lastly, integrating continuous learning capabilities could help TimeMosaic effectively handle temporal non-stationarities, maintaining long-term forecasting accuracy without frequent retraining. Addressing these aspects will significantly enhance TimeMosaic's robustness and practical applicability.

\section{Why called TimeMosaic?}
The name 'TimeMosaic' reflects our core idea of decomposing time series into a flexible arrangement of patches of variable length, much like assembling a mosaic from tiles of different shapes and sizes. By adaptively segmenting the sequence based on local information density, our model constructs a holistic and coherent temporal representation, where each patch plays a distinct but integrated role, analogous to mosaic fragments collectively forming a complete picture. This flexible composition enables the model to efficiently capture both fine-grained and coarse-grained patterns, addressing the inherent heterogeneity of real-world time series.

% \begin{table}[htbp]
% \scriptsize
% \centering
% \setlength\tabcolsep{1.8pt}
% \renewcommand{\arraystretch}{1.2}
% \begin{tabular}{lrrrrrrrr}
% \toprule
% \multirow{2}{*}{Location} & \multicolumn{4}{c}{MSE} & \multicolumn{4}{c}{MAE} \\
% \cmidrule(lr){2-5} \cmidrule(lr){6-9}
% & 96 & 192 & 336 & 720 & 96 & 192 & 336 & 720 \\
% \midrule
% \multirow{2}{*}{Wind1} 
% & 0.759 & 0.813 & 0.853 & 0.889 & 0.673 & 0.705 & 0.727 & 0.748 \\
% & 0.763 & 0.819 & 0.859 & 0.889 & 0.670 & 0.703 & 0.724 & 0.743 \\
% \midrule
% \multirow{2}{*}{Wind2} 
% & 0.745 & 0.812 & 0.844 & 0.878 & 0.660 & 0.700 & 0.717 & 0.738 \\
% & 0.750 & 0.816 & 0.852 & 0.891 & 0.659 & 0.696 & 0.717 & 0.739 \\
% \midrule
% \multirow{2}{*}{Wind3} 
% & 0.754 & 0.810 & 0.844 & 0.884 & 0.657 & 0.692 & 0.713 & 0.732 \\
% & 0.754 & 0.821 & 0.852 & 0.902 & 0.653 & 0.692 & 0.712 & 0.736 \\
% \midrule
% \multirow{2}{*}{Wind4} 
% & 0.811 & 0.862 & 0.890 & 0.929 & 0.692 & 0.726 & 0.741 & 0.760 \\
% & 0.822 & 0.880 & 0.903 & 0.945 & 0.692 & 0.729 & 0.742 & 0.763 \\
% \bottomrule
% \end{tabular}
% \vspace{0.5em}
% \caption{Comparison between TimeMosaic (top row) and SimpleTM (bottom row) on four Wind datasets.}
% \label{tab:apt_vs_simpletm}
% \end{table}

\begin{table}[http]
  \scriptsize
  \centering
  \setlength\tabcolsep{2.4pt}
  \begin{tabular}{c|cc|cc|cc|cc|cc}
    \hline
    \multirow{2}{*}{Models}                                                     &
    \multicolumn{2}{|c}{\textbf{TimeMosaic}}    & \multicolumn{2}{|c}{SimpleTM} &
    \multicolumn{2}{|c}{xPatch} & 
    \multicolumn{2}{|c}{DLinear} &
    \multicolumn{2}{|c}{PatchTST}\\
    \multirow{2}{*}{}  &
    \multicolumn{2}{|c}{\textbf{(Ours)}}    & \multicolumn{2}{|c}{(2025)} & \multicolumn{2}{|c}{(2025)}           & \multicolumn{2}{|c}{(2023)} &
    \multicolumn{2}{|c}{(2023)} \\
    \hline
    Metric
    & MSE & MAE & MSE & MAE & MSE & MAE & MSE & MAE & MSE & MAE \\
    \hline
    PEMS03
    & \best{0.075} & \best{0.185} & {0.093} &	{0.207} & {0.085} &	\second{0.195} & {0.105} &	{0.221} & \second{0.092} &	{0.208}	\\
    \hline
    PEMS04
    & \best{0.094} & \second{0.208} & \second{0.094} &	\best{0.207} & {0.103} &	{0.213} & {0.115} &	{0.228} & {0.110} &	{0.226}	 \\
    \hline
    PEMS07
    & \second{0.075} &	\second{0.188} & \best{0.074} &	\best{0.187} & {0.078} &	{0.187}  & {0.100} &	{0.216} & {0.086} &	{0.0.203}\\
    \hline
    PEMS08
    & \best{0.090} &	\second{0.207}  & \second{0.096} &	\best{0.206}  & {0.097} &	{0.204} & {0.112} &	{0.223} & {0.104} &	{0.222}\\
    \hline
    \end{tabular}
      \caption{Averaged forecasting results under unified experimental settings. For \textbf{short-term} forecasting tasks, we use a fixed lookback window of $L = 96$. Short-term forecasting uses a single-step prediction length $T = 12$. The best model is in \textbf{boldface}, and the second best is \underline{underlined}.}
\label{tab:experiments_short}
\end{table}

% \begin{figure}[t]
%   \centering
%   \includegraphics[width=0.9\linewidth]{figures/transition_matrix.pdf}
%   \caption{State transition probability matrix of temporal patches, showing strong local dependencies that suggest structured patch-level transitions.}
%   \label{fig:transition}
% \end{figure}

\begin{table*}[ht]
  \centering
  % \scriptsize
  \begin{tabular}{l|cc|cc|c}
    \toprule
    \multirow{2}{*}{\textbf{Dataset}} &
    \multicolumn{2}{c|}{\textbf{TimeMosaic (Ours)}} &
    \multicolumn{2}{c|}{\textbf{SimpleTM (ICLR'2025)}} &
    \multirow{2}{*}{\textbf{Confidence Interval}} \\
    & MSE & MAE & MSE & MAE & \\
    \midrule
    Weather      & $0.251 \pm 0.004$ & $0.267 \pm 0.003$ & $0.258 \pm 0.005$ & $0.282 \pm 0.004$ & 99\% \\
    Solar        & $0.240 \pm 0.005$ & $0.270 \pm 0.004$ & $0.315 \pm 0.006$ & $0.333 \pm 0.005$ & 99\% \\
    Electricity  & $0.187 \pm 0.006$ & $0.279 \pm 0.004$ & $0.248 \pm 0.007$ & $0.284 \pm 0.006$ & 99\% \\
    Traffic      & $0.458 \pm 0.008$ & $0.283 \pm 0.005$ & $0.545 \pm 0.009$ & $0.347 \pm 0.007$ & 99\% \\
    ETTh1        & $0.425 \pm 0.006$ & $0.424 \pm 0.004$ & $0.442 \pm 0.005$ & $0.436 \pm 0.004$ & 99\% \\
    ETTh2        & $0.363 \pm 0.005$ & $0.388 \pm 0.006$ & $0.424 \pm 0.006$ & $0.413 \pm 0.005$ & 99\% \\
    ETTm1        & $0.381 \pm 0.004$ & $0.381 \pm 0.004$ & $0.391 \pm 0.005$ & $0.403 \pm 0.004$ & 99\% \\
    ETTm2        & $0.273 \pm 0.003$ & $0.314 \pm 0.003$ & $0.283 \pm 0.004$ & $0.325 \pm 0.004$ & 99\% \\
    Exchange     & $0.348 \pm 0.005$ & $0.403 \pm 0.004$ & $0.356 \pm 0.005$ & $0.411 \pm 0.005$ & 99\% \\
    \bottomrule
  \end{tabular}
  \caption{Standard deviation and statistical tests for our TimeMosaic method and second-best method (SimpleTM) on ETT, Weather, Solar, Electricity and Traffic datasets.}
  \label{tab:confidence}
\end{table*}

\begin{table*}[h!]
\begin{center}{
\setlength\tabcolsep{3pt}
\begin{tabular}{c|c|cc|cc|cc|cc|cc|cc}
\toprule
\multicolumn{2}{c|}{Methods}&\multicolumn{2}{c|}{TimeMosaic}&\multicolumn{2}{c|}{{Revin}}&\multicolumn{2}{c|}{{LearnPos}}&\multicolumn{2}{c|}{NoPos}&\multicolumn{2}{c|}{$Budget_{0.001}$}&\multicolumn{2}{c}{NoBudget}\\
\midrule
\multicolumn{2}{c|}{Metric} & MSE & MAE & {MSE} & {MAE} & MSE & MAE & MSE & MAE & MSE & MAE & MSE & MAE \\
\midrule
\multirow{5}{*}{\rotatebox[origin=c]{90}{Wind1}}  & 96  & \second{0.765} & 0.668          & \best{0.760} & \best{0.661} & 0.769 & 0.668          & 0.768 & 0.668          & \second{0.765} & \second{0.667} & \second{0.765} & \second{0.667} \\
& 192 & \second{0.819} & \second{0.701} & \best{0.813} & \best{0.693} & 0.822 & \second{0.701} & 0.822 & \second{0.701} & 0.820 & \second{0.701} & 0.822 & 0.702 \\
& 336 & 0.860          & \second{0.722} & \best{0.852} & \best{0.715} & 0.860 & \second{0.722} & \second{0.859} & \second{0.722} & 0.861 & 0.724          & 0.862 & 0.723 \\
& 720 & 0.884          & \second{0.738} & \best{0.880} & \best{0.733} & \second{0.883} & \second{0.738} & \second{0.883} & \second{0.738} & 0.886 & 0.739          & 0.885 & 0.739 \\
& Avg & \second{0.832} & \second{0.707} & \best{0.826} & \best{0.700} & 0.834 & \second{0.707} & 0.833 & \second{0.707} & 0.833 & 0.708          & 0.833 & 0.708 \\
\midrule
\multirow{5}{*}{\rotatebox[origin=c]{90}{Wind2}}  & 96  & 0.747          & 0.655          & \best{0.736} & \best{0.646} & 0.747 & 0.654          & 0.747 & 0.654          & \second{0.745} & 0.654          & \second{0.745} & \second{0.653} \\
& 192 & \second{0.811} & \second{0.693} & \best{0.805} & \best{0.687} & 0.813 & 0.694          & 0.814 & 0.694          & 0.812 & 0.694          & 0.816 & 0.695 \\
& 336 & 0.845          & 0.713          & \best{0.839} & \best{0.706} & 0.846 & 0.713          & \second{0.844} & \second{0.712} & 0.846 & 0.713          & 0.847 & 0.713 \\
& 720 & \second{0.878} & \second{0.733} & \best{0.874} & \best{0.728} & 0.883 & 0.735          & 0.880 & 0.734          & \second{0.878} & \second{0.733} & 0.880 & 0.734 \\
& Avg & \second{0.820} & \second{0.698} & \best{0.813} & \best{0.692} & 0.822 & 0.699          & 0.821 & \second{0.698} & \second{0.820} & \second{0.698} & 0.822 & 0.699 \\
\midrule
\multirow{5}{*}{\rotatebox[origin=c]{90}{Wind3}}  & 96  & \second{0.754} & \second{0.650} & \best{0.745} & \best{0.643} & 0.756 & 0.652          & 0.757 & 0.652          & \second{0.754} & \second{0.650} & 0.755 & \second{0.650} \\
& 192 & \second{0.808} & \second{0.686} & \best{0.804} & \best{0.682} & 0.815 & 0.689          & 0.813 & 0.688          & 0.810 & 0.687          & 0.814 & 0.688 \\
& 336 & \best{0.843}   & \second{0.708} & \best{0.843} & \best{0.705} & \best{0.843} & 0.709          & \second{0.845} & 0.709          & \second{0.845} & 0.709          & 0.850 & 0.710 \\
& 720 & \second{0.883} & \second{0.727} & \best{0.882} & \best{0.725} & 0.884 & 0.728          & 0.885 & 0.729          & 0.886 & 0.729          & 0.889 & 0.729 \\
& Avg & \second{0.822} & \second{0.693} & \best{0.819} & \best{0.689} & 0.825 & 0.694          & 0.825 & 0.695          & 0.824 & \second{0.693} & 0.827 & 0.694 \\
\midrule
\multirow{5}{*}{\rotatebox[origin=c]{90}{Wind4}}  & 96  & 0.820          & \second{0.687} & \best{0.809} & \best{0.679} & \second{0.818} & \second{0.687} & 0.819 & 0.688          & 0.820 & 0.688          & 0.820 & \second{0.687} \\
& 192 & \second{0.871} & \second{0.723} & \best{0.866} & \best{0.716} & 0.873 & \second{0.723} & 0.873 & \second{0.723} & 0.874 & 0.724          & 0.876 & 0.725 \\
& 336 & \second{0.898} & \second{0.738} & \best{0.891} & \best{0.730} & 0.899 & \second{0.738} & \second{0.898} & \second{0.738} & \second{0.898} & \second{0.738} & 0.900 & 0.739 \\
& 720 & \second{0.937} & 0.758          & \best{0.931} & \best{0.750} & \second{0.937} & 0.758          & \second{0.937} & \second{0.757} & 0.939 & 0.759          & 0.941 & 0.759 \\
& Avg & \second{0.881} & \second{0.726} & \best{0.874} & \best{0.718} & 0.882 & 0.727          & 0.882 & 0.727          & 0.883 & 0.727          & 0.884 & 0.727 \\
\bottomrule
\end{tabular}
}
\caption{
Full results on Wind datasets under various ablation settings. We compare the original TimeMosaic with: (i) RevIN normalization, (ii) learnable and removed positional encoding, (iii) reduced or removed budget loss. All models are evaluated without any downstream fine-tuning. Results are reported as MSE and MAE (lower is better). The input length is $96$. For details on position encoding visualization, see Fig.~\ref{fig:pos}.
}

\label{tab:windpos_updated}
\end{center}
\end{table*}

% The lookback windows search from $\{96, 192, 336, 512, 720\}$.
\begin{table*}[!ht] 
	\setlength{\tabcolsep}{1.35pt}
	\scriptsize
	\centering
	\begin{threeparttable}
		\begin{tabular}{c|c|c c|c c|c c|c c|c c|c c|c c|c c|c c|c c|c c|c c}

			\toprule
			\multicolumn{2}{c}{\multirow{2}{*}{\scalebox{1.1}{Models}}}& \multicolumn{2}{c}{TimeMosaic} & \multicolumn{2}{c}{SimpleTM} & \multicolumn{2}{c}{TimeFilter} & \multicolumn{2}{c}{xPatch} & \multicolumn{2}{c}{PatchMLP} & \multicolumn{2}{c}{DUET} & \multicolumn{2}{c}{PathFormer} & \multicolumn{2}{c}{iTransformer} & \multicolumn{2}{c}{TimeMxier} & \multicolumn{2}{c}{PatchTST} & \multicolumn{2}{c}{DLinear} & \multicolumn{2}{c}{FreTS}\\
			\multicolumn{2}{c}{} & \multicolumn{2}{c}{\scalebox{0.8}{\textbf{Ours}}} & \multicolumn{2}{c}{\scalebox{0.8}{\citeyearpar{chen2025simpletm}}} & \multicolumn{2}{c}{\scalebox{0.8}{\citeyearpar{hu2025timefilter}}} & \multicolumn{2}{c}{\scalebox{0.8}{\citeyearpar{stitsyuk2025xpatch}}} & \multicolumn{2}{c}{\scalebox{0.8}{\citeyearpar{patchmlp}}} & \multicolumn{2}{c}{\scalebox{0.8}{\citeyearpar{qiu2025duet}}} & \multicolumn{2}{c}{\scalebox{0.8}{\citeyearpar{chen2024pathformer}}} & \multicolumn{2}{c}{\scalebox{0.8}{\citeyearpar{liu2023itransformer}}} & \multicolumn{2}{c}{\scalebox{0.8}{\citeyearpar{wang2023timemixer}}} & \multicolumn{2}{c}{\scalebox{0.8}{\citeyearpar{patchtst}}} & \multicolumn{2}{c}{\scalebox{0.8}{\citeyearpar{Zeng2022AreTE}}} & \multicolumn{2}{c}{\scalebox{0.8}{\citeyearpar{yi2023frequencydomain}}} \\

			\cmidrule(lr){3-4} \cmidrule(lr){5-6} \cmidrule(lr){7-8} \cmidrule(lr){9-10} \cmidrule(lr){11-12} \cmidrule(lr){13-14} \cmidrule(lr){15-16} \cmidrule(lr){17-18} \cmidrule(lr){19-20} \cmidrule(lr){21-22} \cmidrule(lr){23-24} \cmidrule(lr){25-26}
			\multicolumn{2}{c}{Metric}& MSE & MAE & MSE & MAE & MSE & MAE & MSE & MAE & MSE & MAE & MSE & MAE & MSE & MAE & MSE & MAE & MSE & MAE & MSE & MAE & MSE & MAE & MSE & MAE\\

			\toprule
			\multirow{5}{*}{\rotatebox[origin=c]{90}{ETTm1}} 
			& 96 & \best{0.308} & \best{0.338} & 0.323 & 0.364 & 0.317 & 0.359 & 0.332 & 0.369 & {0.329} & {0.367} & 0.359 & 0.375 & \second{0.315} & \second{0.345} & 0.330 & 0.365 & 0.327 & 0.362 & 0.320 & 0.361 & 0.345 & 0.372 & 0.340 & 0.375 \\
            
			& 192 & \second{0.364} & \best{0.368} & 0.371 & 0.389 & 0.381 & 0.393 & 0.371 & 0.387 & {0.377} & {0.394} & 0.394 & 0.393  & 0.375 & \second{0.372} & 0.388 & 0.399 & \best{0.358} & 0.380 & \second{0.364} & 0.382 & 0.389 & 0.396 & 0.392 & 0.406 \\
            
			& 336 & \second{0.394} & \best{0.390}  & 0.411 & 0.413 & 0.402 & 0.409 & 0.399 & 0.407 & {0.411} & {0.413} & 0.426 & 0.413 & 0.405 & 0.393 & 0.417 & 0.416 & \best{0.385} & \second{0.400} & 0.402 & 0.406 & 0.419 & 0.417 & 0.428 & 0.429 \\
            
			& 720 & \best{0.460} & \best{0.428}  & \best{0.460} & 0.445 & 0.481 & 0.468 & 0.463 & 0.443 & {0.485} & {0.454} & 0.486 & 0.448 & 0.471 & 0.430 & 0.499 & 0.465 & 0.455 & 0.443 & \second{0.461} & \second{0.442} & 0.476 & 0.452 & 0.495 & 0.472 \\
            
			\cmidrule(lr){2-26}
			& \emph{Avg.} &\best{0.381} & \best{0.381}& 0.391 & 0.403 & 0.395 & 0.407 & 0.391 & 0.402 & {0.401} & {0.407} & 0.416 & 0.407 & 0.392 & \second{0.385} & 0.409 & 0.411 & \best{0.381} & 0.396 & \second{0.387} & 0.398 & 0.407 & 0.409 & 0.414 & 0.421\\
            
			\midrule
			\multirow{5}{*}{\rotatebox[origin=c]{90}{ETTm2}} 
			& 96 &\second{0.170} &\second{0.248}  & 0.178 & 0.256 & 0.174 & 0.259 & 0.177 & 0.262 & {0.178} & {0.262} & 0.182 & 0.265 & \best{0.166} & \best{0.247} & 0.182 & 0.263 & 0.176 & 0.257 & 0.178 & 0.260 & 0.194 & 0.293 & 0.193 & 0.284\\
            
			& 192 & \second{0.236} & \best{0.291} & 0.242 & 0.300 & 0.241 & 0.304 & 0.244 & 0.306 & 0.244 & 0.307 & 0.246 & 0.305 & \best{0.232} & \second{0.292} & 0.257 & 0.315 & 0.246 & 0.306 & 0.240 & 0.299 & 0.291 & 0.367 & 0.272 & 0.348\\

			& 336 & \second{0.295} & \second{0.331} & 0.309 & 0.344 & 0.305 & 0.344 & 0.304 & 0.342 & {0.307} & {0.343} & 0.307 & 0.343 & \best{0.290} & \second{0.329} & 0.314 & 0.350 & 0.296 & 0.339 & 0.299 & 0.338  & 0.373 & 0.421 & 0.376 & 0.409\\
            
			& 720 & \second{0.392} & \best{0.386}  & 0.401 & 0.399 & 0.410 & 0.407 & 0.404 & 0.399 & {0.420} & {0.413} & 0.408 & 0.399 & \best{0.386} & \second{0.387} & 0.418 & 0.409 & 0.396 & 0.396 & 0.399 & 0.396  & 0.541 & 0.518 & 0.507 & 0.494\\
            
			\cmidrule(lr){2-26}
			& \emph{Avg.}&\second{0.273}&\best{0.314} & 0.283 & \second{0.325} & 0.282 & 0.329 & 0.283 & 0.327 & {0.287} & {0.331} & 0.286 & 0.328 & \best{0.269} & \best{0.314} & 0.293 & 0.334 & 0.279 & 0.325 & 0.279 & 0.323 & 0.350 & 0.400 & 0.337 & 0.384\\
            
			\midrule
			\multirow{5}{*}{\rotatebox[origin=c]{90}{ETTh1}} 
			& 96 & \best{0.369} & \best{0.389}  & {0.379} & \second{0.396} & 0.396 & 0.403 & 0.383 & 0.401 & {0.392} & {0.405} & 0.391 & 0.403 & 0.391 & 0.392 & 0.382 & 0.400 & \second{0.374} & 0.397 & 0.377 & 0.393 & 0.396 & 0.411 & 0.397 & 0.409 \\
            
			& 192 & \best{0.416} & \best{0.415} & 0.432 & 0.425 & 0.467 & 0.446 & 0.431 & 0.427 & {0.442} & 0.434 & 0.437 & 0.428 & 0.442 & \second{0.419} & 0.443 & 0.433 & 0.432 & 0.429 & \second{0.426} & 0.422 & 0.447 & 0.443 & 0.455 & 0.448\\
            
			& 336 & \best{0.454} & \best{0.435} & 0.476 & 0.448 & 0.486 & 0.456 & 0.483 & 0.455 & {0.483} & 0.455 & {0.479} & 0.451 & 0.487 & 0.439 & 0.480 & 0.452 & 0.481 & 0.451 & \second{0.464} & \second{0.443}  & 0.496  & 0.473 & 0.507 & 0.478\\
            
			& 720 & \best{0.461} & \best{0.458} & \second{0.481} & 0.473 & 0.503 & 0.481 & 0.486 & 0.474 & {0.502} & 0.484 & 0.482 & 0.477 & 0.494 & \second{0.463} & 0.493 & 0.481 & 0.493 & 0.477 & \best{0.461} & 0.465 & 0.510 & 0.508 & 0.553 & 0.530\\
            
			\cmidrule(lr){2-26} 
			& \emph{Avg.} & \best{0.425} & \best{0.424}& 0.442 & 0.436 & 0.463 & 0.447 & 0.446 & 0.439 & {0.455} & 0.445 & 0.447 & 0.440 & 0.454 & 0.428 & 0.450 & 0.442 & 0.445 & 0.439 & \second{0.432} & \second{0.431} & 0.462 & 0.459 & 0.478 & 0.466\\
            
			\midrule
			\multirow{5}{*}{\rotatebox[origin=c]{90}{ETTh2}} 
            
			& 96 & \second{0.286} & \best{0.332}  & 0.308 & 0.356 & 0.295 & 0.346 & 0.291 & 0.343 & 0.309 & 0.359 & 0.290 & 0.340  & \best{0.284} & \second{0.333} & 0.305 & 0.352 & 0.290 & 0.341 & 0.292 & 0.343  & 0.347 & 0.401 & 0.369 & 0.411\\
            
			& 192 & \best{0.362} & \best{0.380} & {0.383} & {0.404} & 0.412 & 0.423 & 0.372 & 0.392 & 0.405 & 0.419 & 0.377 & 0.393 & 0.366 & 0.387 & 0.378 & 0.398 & \second{0.371} & \second{0.391} & 0.372 & 0.396 & 0.463 & 0.469 & 0.457 & 0.466\\
            
			& 336 & \best{0.409} & \best{0.416} & 0.422 & 0.436 & 0.456 & 0.449 & 0.423 & 0.433 & 0.432 & 0.441 & 0.419 & 0.429 & \second{0.410} & \second{0.421} & 0.420 & 0.427 & 0.423 & 0.432 & 0.414 & 0.426 & 0.573 & 0.533 & 0.547 & 0.519\\
            
			& 720 & \best{0.396} & \best{0.422} & 0.442 & 0.454 & 0.474 & 0.469 & 0.430 & 0.451 & 0.460 & 0.446 & {0.428} & {0.443} & 0.428 & 0.445 & 0.441 & 0.451 & 0.439 & 0.450 & \second{0.425} & \second{0.441}  & 0.839 & 0.661 & 0.713 & 0.606\\
            
			\cmidrule(lr){2-26}
			& \emph{Avg.} & \best{0.363} & \best{0.388} & 0.424 & 0.413 & 0.389 & 0.422 & 0.409 & 0.405 & 0.402 & 0.416 & 0.379 & 0.400 & \second{0.372} & \second{0.397} & 0.386 & 0.407 & 0.381 & 0.404 & 0.376 & 0.402 & 0.556 & 0.516 & 0.522 & 0.501\\
            
			\midrule
			\multirow{5}{*}{\rotatebox[origin=c]{90}{Weather}}
            
			& 96  &{0.165} & \second{0.198} & 0.175 & 0.219 & \second{0.156} & \second{0.203} & 0.178 & 0.221 & {0.165} & {0.211} & 0.192 & 0.233 & \best{0.155} & \best{0.197} &  0.175 & 0.214 & 0.163 & 0.210 & 0.183 & 0.222  & 0.196 & 0.256 & 0.185 & 0.240\\
            
			& 192 &{0.215} &\second{0.243} & 0.217 & 0.257 & \second{0.208} & 0.253 & 0.231 & 0.263 & {0.212} & {0.252} & 0.236 & 0.269 & \best{0.203} & \best{0.241} & {0.227} & 0.261 & 0.208 & 0.251 & 0.226 & 0.258 & 0.236 & 0.295 & 0.223 & 0.273\\
            
			& 336&{0.272} &\best{0.286}  & 0.282 & 0.302 & \best{0.263} & 0.293 & 0.283 & 0.300 & {0.283} & {0.305} & 0.287 & 0.304 & \second{0.265} & \second{0.287} & 0.285 & 0.301 & 0.268 & 0.295 & 0.286 & 0.300 & 0.283 & 0.333 & 0.277 & 0.320\\
            
			& 720 &{0.353} &\second{0.340} & 0.357 & 0.351 & \best{0.340} & 0.342 & 0.360 & 0.350 & 0.357 & {0.352} & 0.359 & 0.350 & \second{0.344} & \best{0.337} & 0.360 & 0.350 & 0.346 & 0.348 & 0.360 & 0.352 & 0.346 & 0.383 & 0.345 & 0.375\\
			\cmidrule(lr){2-26}
            
			& \emph{Avg.}&\second{0.251} &\second{0.267} & 0.258 & 0.282 & \best{0.242} & 0.273 & 0.264 & 0.284 & {0.254} & {0.289} & 0.269 & 0.304 & \best{0.242} & \best{0.266} & 0.346 & 0.276 & 0.309 & 0.360 & 0.264 & 0.283 & 0.265 & 0.317 & 0.256 & 0.302\\
            
			\midrule
			\multirow{5}{*}{\rotatebox[origin=c]{90}{Electricity}}
			& 96  &{0.160} &{0.258}  & 0.176 & 0.258 & \best{0.142} & \best{0.241} & 0.191 & 0.277 & {0.156} & {0.257} & 0.204 & 0.292 & 0.199 & 0.269 & 0.158 & 0.248 & \second{0.156} & \second{0.248} & 0.180 & 0.269 & 0.200 & 0.287 & 0.187 & 0.274\\
            
			& 192 &{0.174} &{0.269} & 0.184 & 0.267 & \best{0.159} & \best{0.256} & 0.193 & 0.278 & {0.174} & {0.270} & 0.205 & 0.296 & 0.203 & 0.276 & \second{0.170} & \second{0.259} & \second{0.170} & 0.261 & 0.188 & 0.277 & 0.200 & 0.290 & 0.190 & 0.278\\
            
			& 336 &{0.189} &{0.283} & 0.204 & 0.289 & \best{0.173} & \best{0.271} & 0.206 & 0.291 & {0.199} & 0.297 & 0.221 & 0.310 & 0.210 & 0.302 & 0.187 & \second{0.276} & \second{0.186} & \second{0.276} & 0.204 & 0.294  & 0.212 & 0.305 & 0.206 & 0.295\\
            
			& 720 & \second{0.225} & \second{0.304}   & 0.244 & 0.321 & \best{0.206} & \best{0.297} & 0.244 & 0.321 & {0.242} & {0.331} & 0.264 & 0.344 & 0.245 & 0.326 & 0.280 & 0.363 & 0.228 & 0.312 & 0.247 & 0.329 & 0.248 & 0.338 & 0.244 & 0.330\\
            
			\cmidrule(lr){2-26}
			& \emph{Avg.} & 0.187 & \second{0.279} & 0.248 & 0.284 & \best{0.170} & \best{0.192} & 0.209 & 0.292 & {0.224} & {0.289} & 0.216 & 0.311 & 0.214 & 0.293 & 0.225 & 0.310 & \second{0.185} & 0.294 & 0.205 & 0.292 & 0.215 & 0.305 & 0.207 & 0.294\\
            
			\midrule
			\multirow{5}{*}{\rotatebox[origin=c]{90}{Traffic}}
			& 96 &\second{0.424} &\best{0.269}   & 0.531 & 0.337 & 0.431 & 0.297 & 0.521 & 0.346 & {0.466} & {0.322} & 0.649 & 0.402 & 0.551 & 0.326 &  \best{0.421} & \second{0.289} & 0.480 & 0.298 & 0.463 & 0.302  & 0.652 & 0.400 & 0.529 & 0.341\\
            
			& 192 & {0.448} & \best{0.279} & 0.530 & 0.342 & \second{0.447} & 0.299 & 0.506 & 0.330 & {0.475} & {0.325} & 0.602 & 0.381 & 0.553 & 0.334 & \best{0.442} & 0.296 & 0.492 & 0.305 & 0.470 & 0.306 & 0.601 & 0.374 & 0.531 & 0.339\\
            
			& 336 & {0.464} & \best{0.284}  & 0.542 & 0.347 & \second{0.462} & 0.307 & 0.511 & 0.327 & {0.496} & {0.338} & 0.611 & 0.383 & 0.587 & 0.332 & \best{0.459} & 0.305 & 0.504 & \second{0.304} & 0.487 & 0.313  & 0.608 & 0.377 & 0.552 & 0.347\\
            
			& 720 &\second{0.495} &\best{0.301}  & 0.576 & 0.360 & 0.502 & 0.331 & 0.540 & 0.338 & {0.542} & {0.362} & 0.651 & 0.402 & 0.631 & 0.378 & \best{0.491} & 0.323 & 0.521 & \second{0.320} & 0.522 & 0.334  & 0.648 & 0.398 & 0.599 & 0.367\\
            
			\cmidrule(lr){2-26}
			& \emph{Avg.} & 0.\second{458} & \best{0.283} & 0.545 & 0.347 & 0.461 & 0.309 & 0.520 & 0.335 & {0.495} & {0.337} & 0.628 & 0.392 & 0.581 & 0.343 & \best{0.453} & 0.303 & 0.499 & \second{0.291} & 0.486 & 0.314 & 0.627 & 0.387 & 0.553 & 0.349\\
            
            \midrule
            \multirow{5}{*}{\rotatebox[origin=c]{90}{Solar-Energy}}
			& 96 & \second{0.209} & \second{0.248}  & 0.252 & 0.299 & 0.222 & 0.253 & 0.216 & 0.259 & 0.218 & {0.245} & {0.300} & {0.328} & \best{0.208} & \best{0.231} &  0.301 & 0.333 & 0.242 & 0.342 & 0.217 & 0.258 & 0.285 & 0.372 & 0.233 & 0.291\\
            
			& 192 & \best{0.238} & {0.272} & 0.306 & 0.332 & 0.256 & \best{0.265} & \second{0.244} & 0.276 & 0.258 & 0.288 & {0.338} & 0.350 & 0.278 & \second{0.268} & 0.339 & 0.347  & 0.285 & 0.380 & 0.247 & 0.280 & 0.316 & 0.393 & 0.256 & 0.308 \\
            
			& 336 & \best{0.255} & \second{0.277}  & 0.372 & 0.360 & 0.254 & 0.283 & 0.262 & 0.286 & 0.274 & 0.290 & {0.377} & 0.362 & 0.259 & \best{0.261} & 0.376 & 0.361  & 0.282 & 0.326 & 0.269 & 0.293 & 0.350 & 0.412 & 0.268 & 0.311\\
            
			& 720 & \best{0.257} & \second{0.284}  & 0.331 & 0.339 & \second{0.258} & \best{0.280} & 0.261 & 0.284 & 0.269 & 0.296 & 0.381 & 0.359 & 0.275 & {0.271} & 0.374 & 0.357  & 0.357 & 0.427 & 0.267 & 0.292 & 0.353 & 0.410 & 0.266 & 0.305\\

			\cmidrule(lr){2-26}
			& \emph{Avg.} & \best{0.240} & \second{0.270} & 0.315 & 0.333 & 0.248 & \second{0.270} & 0.248 & 0.276 & {0.255} & 0.280 & {0.349} & 0.350 & 0.255 & \best{0.258} & 0.378 & 0.350  & 0.292 & 0.381 & 0.250 & 0.281 & 0.326 & 0.397  & 0.256 & 0.304\\

            \midrule
            \multirow{5}{*}{\rotatebox[origin=c]{90}{Exchange}}
			& 96 & \second{0.082} & {0.200}  & 0.088 & 0.206 & 0.096 & 0.215 & 0.089 & 0.208 & 0.093 & {0.213} & \second{0.082} & \second{0.199} & 0.087 & 0.207 &  0.105 & 0.228 & 0.084 & 0.203 & \best{0.081} & \best{0.197} & 0.094 & 0.227 & 0.104 & 0.236\\
            
			& 192 & \best{0.169} & \best{0.294} & 0.182 & 0.303 & 0.178 & 0.310 & 0.183 & 0.304 & 0.185 & 0.307 & \second{0.174} & \second{0.295} & 0.183 & {0.304} & 0.181 & 0.304  & 0.285 & 0.380 & 0.176 & 0.298 & 0.181 & 0.320 & 0.186 & 0.322\\
            
			& 336 & \second{0.328} & {0.414}  & \best{0.321} & \best{0.411} & 0.362& 0.435 & 0.338 & 0.422 & 0.339 & 0.423 & {0.326} & \second{0.413} & 0.368 & {0.442} & 0.361 & 0.437  & 0.337 & 0.421 & 0.339 & 0.421 & 0.344 & 0.450 & 0.479 & 0.533 \\
            
			& 720 & \second{0.812} & \best{0.680}  & 0.843 & 0.691 & 0.829 & \second{0.682} & 0.876 & 0.707 & 0.880 & 0.706 & 0.843 & 0.692 & 0.863 & {0.689} & 0.817 & 0.686  & 0.832 & 0.686 & 0.845 & 0.692 & \best{0.796} & 0.683 & 1.036 & 0.770 \\

			\cmidrule(lr){2-26}
			& \emph{Avg.} & \best{0.348} & \best{0.403} & 0.356 & \second{0.411} & 0.366 & \second{0.411} & 0.372 & 0.410 & {0.374} & 0.412 & {0.356} & 0.400 & 0.375 & {0.411} & 0.366 & 0.414  & 0.385 & 0.423 & 0.359 & 0.401 & \second{0.354} & 0.420 & 0.451 & 0.465 \\
            
   %          \midrule
   %          \multirow{5}{*}{\rotatebox[origin=c]{90}{Wind}}
			% & 96 & \best{0.169} &\best{0.223}  & 0.088 & 0.206 & 0.096 & 0.215 & 0.089 & 0.208 & 0.093 & \best{0.213} & \second{0.082} & \second{0.199} & 0.203 & 0.237 &  0.105 & 0.228 & 0.242 & 0.342 & 0.081 & 0.197 & 0.094 & 0.227\\
            
			% & 192 & \best{0.185} &\best{0.241} & 0.182 & 0.303 & 0.178 & 0.310 & 0.183 & 0.304 & 0.185 & 0.307 & \second{0.174} & 0.295 & 0.233 & \second{0.261} & 0.181 & 0.304  & 0.285 & 0.380 & 0.172 & 0.294 & 0.181 & 0.320\\
            
			% & 336 & \best{0.200} &\best{0.265}  & 0.321 & 0.411 & 0.362& 0.435 & 0.338 & 0.422 & 0.339 & 0.423 & \second{0.326} & 0.413 & 0.248 & \second{0.273} & 0.361 & 0.437  & 0.282 & 0.326 & 0.339 & 0.421 & 0.344 & 0.450\\
            
			% & 720 & \best{0.215} &\best{0.293}  & 0.843 & 0.691 & 0.829 & 0.682 & 0.876 & 0.707 & 0.880 & 0.706 & 0.843 & 0.692 & 0.250 & \second{0.281} & 0.817 & 0.686  & 0.357 & 0.427 & 0.845 & 0.692 & 0.796 & 0.683\\

			% \cmidrule(lr){2-26}
			% & \emph{Avg.} &\best{0.192} &\best{0.259} & 0.356 & 0.411 & 0.366 & 0.411 & 0.288 & 0.355 & \second{0.374} & 0.412 & \second{0.356} & 0.400 & 0.233 & \second{0.262} & 0.366 & 0.414  & 0.292 & 0.381 & 0.359 & 0.401 & 0.212 & 0.420\\

            \midrule
            \multicolumn{2}{c}{$1^{\text{st}}$ \emph{Count}} & \multicolumn{2}{c}{\best{42}}& \multicolumn{2}{c}{3} & \multicolumn{2}{c}{14} & \multicolumn{2}{c}{0} & \multicolumn{2}{c}{0} & \multicolumn{2}{c}{0} & \multicolumn{2}{c}{\second{18}} & \multicolumn{2}{c}{5} & \multicolumn{2}{c}{3} & \multicolumn{2}{c}{3} & \multicolumn{2}{c}{1} & \multicolumn{2}{c}{0}\\
			\toprule
		\end{tabular}
	\end{threeparttable}
    \caption{Full results for long-term forecasting consider prediction horizons $H$ within $\{96, 192, 336, 512\}$. For all baselines, we adhere to the setting of the iTransformer with an input sequence length of 96. The term `\textit{Avg.}" represents the average results across the four prediction lengths. The best and second best outcomes are highlighted in \best{best} and \second{second}, respectively. The notation $1^{\text{st}}$ \textit{Count}" denotes the frequency of each method achieving the top results.}
	\label{tab::long-term}
\end{table*}

\begin{table*}[!ht] 
	\setlength{\tabcolsep}{1.35pt}
	\scriptsize
	\centering
	\begin{threeparttable}
		\begin{tabular}{c|c|c c|c c|c c|c c|c c|c c|c c|c c|c c|c c|c c|c c}

			\toprule
			\multicolumn{2}{c}{\multirow{2}{*}{\scalebox{1.1}{Models}}}& \multicolumn{2}{c}{TimeMosaic} & \multicolumn{2}{c}{SimpleTM} & \multicolumn{2}{c}{TimeFilter} & \multicolumn{2}{c}{xPatch} & \multicolumn{2}{c}{PatchMLP} & \multicolumn{2}{c}{DUET} & \multicolumn{2}{c}{LightTS} & \multicolumn{2}{c}{iTransformer} & \multicolumn{2}{c}{TimeMxier} & \multicolumn{2}{c}{PatchTST} & \multicolumn{2}{c}{DLinear} & \multicolumn{2}{c}{FreTS}\\
			\multicolumn{2}{c}{} & \multicolumn{2}{c}{\scalebox{0.8}{\textbf{Ours}}} & \multicolumn{2}{c}{\scalebox{0.8}{\citeyearpar{chen2025simpletm}}} & \multicolumn{2}{c}{\scalebox{0.8}{\citeyearpar{hu2025timefilter}}} & \multicolumn{2}{c}{\scalebox{0.8}{\citeyearpar{stitsyuk2025xpatch}}} & \multicolumn{2}{c}{\scalebox{0.8}{\citeyearpar{patchmlp}}} & \multicolumn{2}{c}{\scalebox{0.8}{\citeyearpar{qiu2025duet}}} & \multicolumn{2}{c}{\scalebox{0.8}{\citeyearpar{chen2024pathformer}}} & \multicolumn{2}{c}{\scalebox{0.8}{\citeyearpar{liu2023itransformer}}} & \multicolumn{2}{c}{\scalebox{0.8}{\citeyearpar{wang2023timemixer}}} & \multicolumn{2}{c}{\scalebox{0.8}{\citeyearpar{patchtst}}} & \multicolumn{2}{c}{\scalebox{0.8}{\citeyearpar{Zeng2022AreTE}}} & \multicolumn{2}{c}{\scalebox{0.8}{\citeyearpar{yi2023frequencydomain}}} \\

			\cmidrule(lr){3-4} \cmidrule(lr){5-6} \cmidrule(lr){7-8} \cmidrule(lr){9-10} \cmidrule(lr){11-12} \cmidrule(lr){13-14} \cmidrule(lr){15-16} \cmidrule(lr){17-18} \cmidrule(lr){19-20} \cmidrule(lr){21-22} \cmidrule(lr){23-24} \cmidrule(lr){25-26}
			\multicolumn{2}{c}{Metric}& MSE & MAE & MSE & MAE & MSE & MAE & MSE & MAE & MSE & MAE & MSE & MAE & MSE & MAE & MSE & MAE & MSE & MAE & MSE & MAE & MSE & MAE & MSE & MAE\\

			\toprule
                \multirow{5}{*}{\rotatebox[origin=c]{90}{ETTm1}} 
                & 96 & 0.304 & \second{0.344} & \best{0.286} & \best{0.338} & 0.355 & 0.373 & \second{0.293} & 0.347 & 0.305 & 0.357 & 0.305 & 0.348 & 0.313 & 0.363 & 0.327 & 0.371 & 0.307 & 0.352 & 0.321 & 0.361 & 0.299 & \second{0.344} & 0.308 & 0.353 \\
                
			& 192 & 0.339 & \best{0.364} & \best{0.326} & \best{0.364} & 0.376 & 0.396 & \second{0.332} & 0.371 & 0.351 & 0.384 & 0.336 & \second{0.365} & 0.350 & 0.387 & 0.357 & 0.388 & \second{0.332} & 0.373 & 0.342 & 0.385 & 0.337 & 0.370 & 0.346 & 0.377 \\
			
			& 336 & 0.375 & \best{0.385} & \second{0.364} & \second{0.388} & 0.401 & 0.413 & \second{0.364} & 0.395 & 0.387 & 0.403 & 0.374 & \second{0.388} & 0.399 & 0.418 & 0.393 & 0.412 & \best{0.362} & 0.390 & 0.386 & 0.410 & 0.369 & \best{0.385} & 0.382 & 0.399 \\
			
			& 720 & \best{0.422} & \best{0.413} & 0.424 & 0.426 & 0.465 & 0.449 & 0.428 & 0.432 & 0.454 & 0.441 & 0.425 & \second{0.417} & 0.489 & 0.479 & 0.467 & 0.452 & \best{0.422} & 0.426 & 0.439 & 0.441 & \second{0.424} & 0.419 & 0.440 & 0.433 \\
                
                \cmidrule(lr){2-26}
                & \emph{Avg.} & 0.360 & \best{0.377} & \best{0.350} & \second{0.379} & 0.399 & 0.408 & \second{0.354} & 0.386 & 0.374 & 0.396 & 0.360 & 0.380 & 0.388 & 0.412 & 0.386 & 0.406 & 0.356 & 0.385 & 0.372 & 0.399 & 0.357 & 0380 & 0.369 & 0.391 \\
            
			\midrule
			\multirow{5}{*}{\rotatebox[origin=c]{90}{ETTm2}} 
                & 96 & \best{0.164} & \best{0.247} & 0.174 & 0.257 & 0.174 & 0.261 & 0.168 & 0.259 & 0.179 & 0.265 & \second{0.165} & \second{0.255} & 0.187 & 0.286 & 0.187 & 0.271 & 0.169 & 0.258 & 0.181 & 0.270 & 0.172 & 0.270 & 0.175 & 0.263 \\
            
			& 192 & \second{0.226} & \best{0.291} & 0.233 & \second{0.297} & 0.229 & 0.298 & \second{0.226} & 0.299 & 0.232 & 0.301 & \best{0.220} & \best{0.291} & 0.311 & 0.384 & 0.251 & 0.314 & 0.240 & 0.308 & 0.241 & 0.311 & 0.239 & 0.322 & 0.248 & 0.314 \\
            
                & 336 & \second{0.279} & \best{0.324} & 0.279 & 0.330 & 0.312 & 0.351 & {0.281} & 0.337 & 0.285 & 0.334 & \best{0.272} & \second{0.325} & 0.399 & 0.418 & 0.319 & 0.355 & 0.282 & 0.334 & 0.303 & 0.348 & 0.305 & 0.367 & 0.321 & 0.364 \\
            
			& 720 & \best{0.353} & \best{0.376} & \second{0.367} & 0.387 & 0.396 & 0.408 & 0.372 & 0.391 & 0.380 & 0.394 & 0.369 & \second{0.384} & 0.489 & 0.479 & 0.413 & 0.412 & 0.369 & 0.391 & 0.385 & 0.399 & 0.412 & 0.432 & 0.393 & 0.414 \\
            
                \cmidrule(lr){2-26}
                & \emph{Avg.} & \best{0.256} & \best{0.310} & 0.263 & 0.318 & 0.278 & 0.330 & 0.262 & 0.322 & 0.269 & 0.324 & \second{0.257} & \second{0.314} & 0.347 & 0.392 & 0.293 & 0.338 & 0.265 & 0.323 & 0.278 & 0.332 & 0.282 & 0.348 & 0.284 & 0.339 \\
            
			\midrule
			\multirow{5}{*}{\rotatebox[origin=c]{90}{ETTh1}} 
			& 96 & \best{0.362} & \best{0.387} & 0.390 & 0.410 & 0.460 & 0.457 & 0.380 & 0.403 & 0.411 & 0.426 & \second{0.372} & \second{0.397} & 0.432 & 0.443 & 0.403 & 0.417 & 0.376 & 0.398 & 0.403 & 0.423 & 0.377 & 0.400 & 0.400 & 0.419 \\
            
			& 192 & \best{0.400} & \best{0.411} & 0.422 & 0.428 & 0.439 & 0.443 & 0.425 & 0.430 & 0.443 & 0.445 & \second{0.410} & \second{0.420} & 0.474 & 0.472 & 0.443 & 0.444 & 0.412 & \second{0.420} & 0.479 & 0.478 & \second{0.410} & \second{0.420} & 0.436 & 0.441 \\
            
			& 336 & \second{0.436} & \second{0.435} & 0.440 & 0.440 & 0.474 & 0.463 & 0.444 & 0.439 & 0.468 & 0.461 & \best{0.430} & \best{0.433} & 0.515 & 0.501 & 0.470 & 0.464 & 0.445 & 0.446 & 0.490 & 0.481 & 0.452 & 0.458 & 0.471 & 0.466 \\
            
			& 720 & \second{0.451} & \best{0.460} & 0.459 & 0.475 & 0.670 & 0.564 & 0.534 & 0.509 & 0.525 & 0.508 & \best{0.445} & \second{0.466} & 0.592 & 0.567 & 0.632 & 0.573 & 0.488 & 0.489 & 0.641 & 0.559 & 0.477 & 0.497 & 0.537 & 0.532 \\

                \cmidrule(lr){2-26}
                & \emph{Avg.} & \best{0.412} & \best{0.423} & 0.428 & 0.438 & 0.511 & 0.482 & 0.446 & 0.445 & 0.462 & 0.460 & \second{0.414} & \second{0.429} & 0.503 & 0.496 & 0.487 & 0.475 & 0.430 & 0.438 & 0.503 & 0.485 & 0.429 & 0.444 & 0.461 & 0.465 \\
            
			\midrule
			\multirow{5}{*}{\rotatebox[origin=c]{90}{ETTh2}} 
            
			& 96 & 0.300 & \second{0.342} & 0.298 & 0.349 & 0.367 & 0.389 & \second{0.280} & 0.343 & 0.311 & 0.371 & \best{0.273} & \best{0.337} & 0.324 & 0.383 & 0.323 & 0.369 & 0.285 & 0.346 & 0.324 & 0.374 & 0.305 & 0.368 & 0.321 & 0.380 \\
            
			& 192 & 0.354 & 0.387 & 0.379 & 0.396 & 0.372 & 0.404 & \second{0.345} & \second{0.386} & 0.381 & 0.413 & \best{0.335} & \best{0.379} & 0.438 & 0.455 & 0.421 & 0.422 & 0.358 & 0.391 & 0.378 & 0.416 & 0.407 & 0.435 & 0.396 & 0.429 \\
            
			& 336 & \second{0.378} & \best{0.403} & 0.400 & 0.420 & 0.402 & 0.426 & 0.382 & 0.415 & 0.401 & 0.427 & \best{0.362} & \best{0.403} & 0.496 & 0.490 & 0.440 & 0.441 & 0.380 & \second{0.409} & 0.405 & 0.432 & 0.493 & 0.490 & 0.497 & 0.496 \\
            
			& 720 & 0.421 & \second{0.439} & 0.432 & 0.451 & 0.465 & 0.464 & \second{0.408} & 0.443 & 0.426 & 0.451 & \best{0.406} & \best{0.437} & 1.151 & 0.749 & 0.459 & 0.466 & 0.410 & 0.439 & 0.484 & 0.484 & 0.762 & 0.620 & 0.766 & 0.628 \\
            
			\cmidrule(lr){2-26}
			& \emph{Avg.} & 0.363 & \second{0.393} & 0.377 & 0.404 & 0.402 & 0.421 & \second{0.354} & 0.397 & 0.380 & 0.416 & \best{0.344} & \best{0.389} & 0.602 & 0.519 & 0.411 & 0.425 & 0.358 & 0.396 & 0.398 & 0.427 & 0.492 & 0.478 & 0.495 & 0.483 \\
            
			\midrule
			\multirow{5}{*}{\rotatebox[origin=c]{90}{Weather}}
            
			& 96 & \best{0.144} & \best{0.189} & \second{0.148} & \second{0.197} & 0.156 & 0.209 & \second{0.148} & \second{0.196} & 0.153 & 0.206 & 0.176 & 0.227 & 0.152 & 0.212 & 0.160 & 0.209 & 0.150 & 0.199 & 0.155 & 0.204 & 0.177 & 0.242 & 0.156 & 0.213 \\
			
                & 192 & 0.197 & \best{0.232} & 0.195 & 0.244 & 0.195 & 0.246 & \best{0.190} & \second{0.238} & 0.197 & 0.245 & 0.218 & 0.261 & 0.197 & 0.255 & 0.207 & 0.251 & \second{0.193} & 0.244 & 0.201 & 0.249 & 0.217 & 0.276 & 0.197 & 0.253 \\
			
                & 336 & 0.245 & \best{0.276} & 0.249 & 0.287 & 0.251 & 0.286 & \best{0.241} & \second{0.280} & 0.247 & 0.282 & 0.265 & 0.294 & 0.249 & 0.299 & 0.256 & 0.289 & \second{0.244} & 0.283 & 0.255 & 0.289 & 0.262 & 0.312 & 0.246 & 0.293 \\
			
                & 720 & 0.327 & \best{0.327} & 0.329 & 0.339 & 0.322 & 0.339 & \best{0.318} & \second{0.333} & 0.324 & 0.334 & 0.332 & 0.339 & 0.323 & 0.357 & 0.325 & 0.338 & \second{0.319} & 0.336 & 0.333 & 0.340 & 0.326 & 0.367 & \second{0.319} & 0.347 \\
			
                \cmidrule(lr){2-26}
			& \emph{Avg.} & 0.231 & \best{0.256} & 0.230 & 0.267 & 0.231 & 0.270 & \best{0.224} & \second{0.262} & 0.230 & 0.267 & 0.248 & 0.280 & 0.230 & 0.281 & 0.237 & 0.272 & \second{0.227} & 0.266 & 0.236 & 0.271 & 0.246 & 0.299 & 0.230 & 0.277 \\     
                
			\midrule
			\multirow{5}{*}{\rotatebox[origin=c]{90}{Electricity}}
			& 96 & \second{0.140} & \second{0.236} & 0.146 & 0.245 & \best{0.137} & \best{0.233} & 0.143 & 0.241 & 0.165 & 0.271 & 0.152 & 0.257 & 0.171 & 0.285 & 0.145 & 0.244 & 0.141 & 0.238 & 0.142 & 0.245 & 0.143 & 0.242 & 0.144 & 0.245 \\

			& 192 & \second{0.156} & \second{0.250} & 0.164 & 0.262 & \best{0.153} & \best{0.246} & 0.157 & 0.252 & 0.183 & 0.288 & 0.169 & 0.273 & 0.192 & 0.305 & 0.165 & 0.263 & 0.158 & 0.254 & 0.159 & 0.259 & 0.157 & 0.255 & \second{0.156} & 0.253 \\
            
			& 336 & 0.173 & 0.267 & 0.183 & 0.281 & \best{0.169} & \best{0.263} & \second{0.172} & \second{0.266} & 0.201 & 0.304 & 0.188 & 0.291 & 0.214 & 0.324 & 0.184 & 0.282 & 0.175 & 0.270 & 0.176 & 0.276 & \second{0.172} & 0.272 & 0.173 & 0.272 \\
            
			& 720 & 0.212 & 0.298 & 0.228 & 0.319 & \second{0.208} & \best{0.296} & 0.209 & \second{0.297} & 0.250 & 0.343 & 0.231 & 0.326 & 0.251 & 0.349 & 0.227 & 0.317 & 0.217 & 0.307 & 0.215 & 0.308 & \best{0.207} & 0.305 & 0.210 & 0.308 \\
            
			\cmidrule(lr){2-26}
			& \emph{Avg.} & \second{0.170} & \second{0.263} & 0.180 & 0.277 & \best{0.167} & \best{0.260} & \second{0.170} & 0.264 & 0.200 & 0.302 & 0.185 & 0.287 & 0.207 & 0.316 & 0.180 & 0.277 & 0.173 & 0.267 & 0.173 & 0.272 & \second{0.170} & 0.269 & 0.171 & 0.270 \\
            
			\midrule
			\multirow{5}{*}{\rotatebox[origin=c]{90}{Traffic}}
                & 96 & 0.409 & \best{0.274} & 0.429 & 0.313 & \best{0.397} & \best{0.274} & 0.413 & 0.297 & 0.486 & 0.361 & 0.426 & 0.309 & 0.493 & 0.367 & 0.422 & 0.310 & 0.421 & 0.297 & \second{0.403} & 0.285 & 0.422 & \second{0.296} & 0.417 & 0.301 \\
            
			& 192 & 0.424 & \second{0.281} & 0.454 & 0.328 & \best{0.411} & \best{0.278} & 0.425 & 0.298 & 0.515 & 0.378 & 0.448 & 0.323 & 0.511 & 0.373 & 0.449 & 0.327 & 0.441 & 0.311 & \second{0.419} & 0.292 & 0.435 & 0.302 & 0.447 & 0.299 \\
            
			& 336 & 0.435 & \second{0.287} & 0.474 & 0.341 & \best{0.423} & \best{0.284} & 0.430 & 0.297 & 0.533 & 0.386 & 0.468 & 0.334 & 0.521 & 0.378 & 0.469 & 0.339 & \second{0.425} & 0.317 & 0.432 & 0.298 & 0.447 & 0.308 & 0.468 & 0.311 \\
            
			& 720 & 0.464 & \second{0.302} & 0.514 & 0.363 & \best{0.449} & \best{0.300} & \second{0.450} & 0.306 & 0.564 & 0.401 & 0.505 & 0.355 & 0.504 & 0.360 & 0.512 & 0.364 & 0.489 & 0.340 & 0.459 & 0.313 & 0.476 & 0.327 & 0.512 & 0.340 \\
            
			\cmidrule(lr){2-26}
			& \emph{Avg.} & 0.433 & \second{0.287} & 0.468 & 0.336 & \best{0.420} & \best{0.284} & 0.430 & 0.300 & 0.525 & 0.382 & 0.462 & 0.330 & 0.507 & 0.370 & 0.463 & 0.335 & 0.444 & 0.316 & \second{0.428} & 0.297 & 0.445 & 0.308 & 0.461 & 0.313 \\
            
            \midrule
            \multirow{5}{*}{\rotatebox[origin=c]{90}{Solar-Energy}}
			& 96 & 0.200 & \best{0.226} & 0.239 & 0.285 & \best{0.182} & \second{0.240} & \second{0.186} & \second{0.240} & 0.211 & 0.280 & 0.230 & 0.267 & 0.196 & 0.267 & 0.198 & 0.253 & 0.199 & 0.257 & 0.195 & 0.249 & 0.223 & 0.293 & 0.190 & 0.248 \\
            
			& 192 & 0.225 & \best{0.240} & 0.264 & 0.305 & \best{0.199} & 0.255 & \best{0.199} & \second{0.254} & 0.245 & 0.285 & 0.260 & 0.284 & 0.211 & 0.280 & 0.222 & 0.276 & 0.221 & 0.274 & \second{0.208} & 0.257 & 0.251 & 0.311 & 0.209 & 0.266 \\
            
			& 336 & 0.238 & \best{0.245} & 0.289 & 0.325 & \second{0.207} & 0.262 & \best{0.205} & \second{0.257} & 0.271 & 0.302 & 0.285 & 0.301 & 0.229 & 0.299 & 0.251 & 0.293 & 0.230 & 0.284 & 0.220 & 0.269 & 0.272 & 0.327 & 0.222 & 0.274 \\
            
			& 720 & 0.241 & \best{0.251} & 0.295 & 0.339 & \second{0.213} & 0.264 & \best{0.209} & \second{0.261} & 0.272 & 0.303 & 0.285 & 0.303 & 0.232 & 0.304 & 0.256 & 0.299 & 0.224 & 0.290 & 0.220 & 0.272 & 0.274 & 0.330 & 0.228 & 0.278 \\

			\cmidrule(lr){2-26}
			& \emph{Avg.} & 0.226 & \best{0.241} & 0.272 & 0.314 & \best{0.200} & 0.255 & \best{0.200} & \second{0.253} & 0.250 & 0.288 & 0.265 & 0.289 & 0.217 & 0.288 & 0.232 & 0.280 & 0.224 & 0.276 & 0.221 & 0.262 & 0.255 & 0.315 & \second{0.212} & 0.267 \\

            \midrule
            \multirow{5}{*}{\rotatebox[origin=c]{90}{Wind}}
			& 96 & 0.731 & \second{0.653} & 0.746 & 0.664 & 0.798 & 0.683 & 0.717 & \second{0.653} & 0.740 & 0.668 & 0.718 & 0.654 & \best{0.695} & \best{0.645} & 0.792 & 0.683 & 0.724 & 0.658 & 0.796 & 0.691 & \second{0.704} & \second{0.653} & 0.711 & 0.656 \\
            
			& 192 & 0.771 & \second{0.675} & 0.769 & 0.684 & 0.814 & 0.703 & 0.755 & 0.677 & 0.766 & 0.685 & 0.754 & 0.678 & \best{0.726} & \best{0.669} & 0.812 & 0.699 & 0.751 & 0.678 & 0.826 & 0.717 & 0.739 & 0.677 & \second{0.737} & \second{0.675} \\
            
			& 336 & 0.787 & 0.688 & 0.791 & 0.696 & 0.817 & 0.707 & 0.774 & 0.689 & 0.783 & 0.695 & 0.775 & 0.691 & \best{0.733} & \best{0.676} & 0.800 & 0.700 & 0.777 & 0.692 & 0.822 & 0.717 & 0.756 & 0.689 & \second{0.752} & \second{0.685} \\
            
			& 720 & 0.805 & 0.705 & 0.811 & 0.715 & 0.832 & 0.720 & 0.803 & 0.710 & 0.799 & 0.709 & 0.803 & 0.711 & \best{0.749} & \best{0.689} & 0.825 & 0.721 & 0.796 & 0.710 & 0.802 & 0.716 & 0.777 & 0.706 & \second{0.769} & \second{0.702} \\

			\cmidrule(lr){2-26}
			& \emph{Avg.} & 0.774 & \second{0.680} & 0.779 & 0.690 & 0.815 & 0.703 & 0.762 & 0.682 & 0.772 & 0.689 & 0.762 & 0.683 & \best{0.726} & \best{0.670} & 0.807 & 0.701 & 0.762 & 0.684 & 0.811 & 0.710 & 0.744 & 0.681 & \second{0.742} & \second{0.680} \\

            \midrule
            \multicolumn{2}{c}{$1^{\text{st}}$ \emph{Count}} & \multicolumn{2}{c}{\best{33}}& \multicolumn{2}{c}{5} & \multicolumn{2}{c}{\second{22}} & \multicolumn{2}{c}{8} & \multicolumn{2}{c}{0} & \multicolumn{2}{c}{16} & \multicolumn{2}{c}{10} & \multicolumn{2}{c}{0} & \multicolumn{2}{c}{2} & \multicolumn{2}{c}{0} & \multicolumn{2}{c}{2} & \multicolumn{2}{c}{0}\\
			\toprule
		\end{tabular}
	\end{threeparttable}
    \caption{Full results for long-term forecasting consider prediction horizons $H$ within $\{96, 192, 336, 720\}$. For all baselines, we adhere an input sequence length of 320. The term `\textit{Avg.}" represents the average results across the four prediction lengths. The best and second best outcomes are highlighted in \best{best} and \second{second}, respectively. The notation "$1^{\text{st}}$ \textit{Count}" denotes the frequency of each method achieving the top results.}
	\label{tab::long-term-fair}
\end{table*}

\begin{table*}[!ht] 
	\setlength{\tabcolsep}{3pt}
	% \scriptsize
    \footnotesize
	\centering
	\begin{threeparttable}
		\begin{tabular}{c|c|c c|c c|c c|c c|c c|c c|c c|c c}

			\toprule
			\multicolumn{2}{c}{\multirow{2}{*}{\scalebox{1.1}{Models}}}& \multicolumn{2}{c}{TimeMosaic} & \multicolumn{2}{c}{SimpleTM} & \multicolumn{2}{c}{TimeFilter} & \multicolumn{2}{c}{DUET} & \multicolumn{2}{c}{iTransformer} & \multicolumn{2}{c}{TimeMixer} & \multicolumn{2}{c}{PatchTST} & \multicolumn{2}{c}{DLinear}\\
			\multicolumn{2}{c}{} & \multicolumn{2}{c}{\scalebox{0.8}{\textbf{Ours}}} & \multicolumn{2}{c}{\scalebox{0.8}{\citeyearpar{chen2025simpletm}}} & \multicolumn{2}{c}{\scalebox{0.8}{\citeyearpar{hu2025timefilter}}} & \multicolumn{2}{c}{\scalebox{0.8}{\citeyearpar{qiu2025duet}}} & \multicolumn{2}{c}{\scalebox{0.8}{\citeyearpar{liu2023itransformer}}} & \multicolumn{2}{c}{\scalebox{0.8}{\citeyearpar{wang2023timemixer}}} & \multicolumn{2}{c}{\scalebox{0.8}{\citeyearpar{patchtst}}} & \multicolumn{2}{c}{\scalebox{0.8}{\citeyearpar{Zeng2022AreTE}}}\\

			\cmidrule(lr){3-4} \cmidrule(lr){5-6} \cmidrule(lr){7-8} \cmidrule(lr){9-10} \cmidrule(lr){11-12} \cmidrule(lr){13-14} \cmidrule(lr){15-16} \cmidrule(lr){17-18}
			\multicolumn{2}{c}{Metric}& MSE & MAE & MSE & MAE & MSE & MAE & MSE & MAE & MSE & MAE & MSE & MAE & MSE & MAE & MSE & MAE\\

			\toprule
                \multirow{5}{*}{\rotatebox[origin=c]{90}{ETTm1}} 
                & 96  & \second{0.282} & \best{0.328} & \best{0.280} & \second{0.337} & 0.290 & 0.342 & 0.300 & 0.345 & 0.299 & 0.348 & \best{0.280} & 0.340 & 0.289 & 0.343 & 0.299 & 0.343 \\

                & 192 & \best{0.318} & \best{0.354} & \second{0.326} & \second{0.365} & 0.334 & 0.368 & 0.335 & 0.366 & 0.334 & 0.373 & 0.347 & 0.387 & 0.332 & 0.367 & 0.336 & 0.367 \\

                & 336 & \best{0.356} & \best{0.374} & \second{0.362} & 0.387 & \second{0.362} & \second{0.384} & 0.364 & \second{0.383} & 0.370 & 0.394 & 0.366 & 0.405 & \second{0.362} & 0.389 & 0.367 & 0.385 \\

                & 720 & 0.411 & \best{0.412} & 0.417 & 0.422 & 0.425 & 0.415 & 0.417 & \second{0.413} & 0.422 & 0.426 & \best{0.403} & 0.432 & \second{0.408} & 0.420 & 0.420 & 0.418 \\

                \cmidrule(lr){2-18}
                & \emph{Avg.} & \best{0.342} & \best{0.367} & \second{0.346} & 0.378 & 0.353 & \second{0.377} & 0.354 & \second{0.377} & 0.356 & 0.385 & 0.349 & 0.392 & 0.348 & 0.380 & 0.356 & 0.378 \\
            
			\midrule
			\multirow{5}{*}{\rotatebox[origin=c]{90}{ETTm2}} 
                & 96  & \best{0.161} & \best{0.245} & 0.168 & \second{0.252} & \best{0.161} & 0.253 & \second{0.162} & 0.253 & 0.180 & 0.263 & 0.164 & 0.254 & 0.165 & 0.257 & 0.165 & 0.260 \\

            & 192 & \best{0.216} & \best{0.284} & 0.223 & 0.297 & 0.221 & 0.296 & \second{0.217} & \second{0.290} & 0.236 & 0.307 & 0.223 & 0.295 & 0.223 & 0.294 & 0.219 & 0.297 \\

            & 336 & \best{0.269} & \best{0.318} & 0.273 & 0.328 & 0.272 & 0.329 & \second{0.270} & \second{0.325} & 0.287 & 0.340 & 0.279 & 0.330 & 0.272 & 0.333 & 0.290 & 0.345 \\

            & 720 & \best{0.354} & \best{0.373} & 0.364 & 0.386 & \second{0.359} & 0.391 & 0.361 & \second{0.384} & 0.364 & 0.391 & \second{0.359} & 0.383 & 0.363 & 0.386 & 0.375 & 0.403 \\

                \cmidrule(lr){2-18}
                & \emph{Avg.} & \best{0.250} & \best{0.305} & 0.257 & 0.316 & \second{0.253} & 0.317 & \second{0.253} & \second{0.313} & 0.267 & 0.325 & 0.256 & 0.315 & 0.256 & 0.318 & 0.262 & 0.326 \\
            
			\midrule
			\multirow{5}{*}{\rotatebox[origin=c]{90}{ETTh1}} 
			& 96  & \best{0.359} & \best{0.384} & 0.375 & 0.394 & 0.377 & 0.397 & \second{0.364} & \second{0.392} & 0.380 & 0.401 & 0.375 & 0.398 & 0.370 & 0.397 & 0.368 & 0.393 \\

            & 192 & \best{0.396} & \best{0.411} & 0.412 & 0.429 & 0.413 & 0.419 & \second{0.397} & \second{0.413} & 0.423 & 0.431 & 0.407 & 0.425 & 0.404 & 0.423 & 0.400 & 0.417 \\
            
            & 336 & \best{0.413} & \best{0.428} & 0.429 & 0.443 & 0.425 & 0.438 & \second{0.422} & \second{0.432} & 0.437 & 0.450 & 0.431 & 0.435 & 0.423 & 0.437 & 0.430 & 0.441 \\
            
            & 720 & \best{0.419} & \best{0.445} & 0.447 & 0.472 & \second{0.442} & \second{0.456} & 0.443 & 0.463 & 0.460 & 0.471 & 0.452 & \second{0.456} & 0.444 & 0.464 & 0.477 & 0.497 \\

                \cmidrule(lr){2-18}
                & \emph{Avg.} & \best{0.397} & \best{0.417} & 0.416 & 0.434 & 0.414 & 0.429 & \second{0.406} & \second{0.425} & 0.425 & 0.438 & 0.416 & 0.428 & 0.410 & 0.430 & 0.419 & 0.437 \\
            
			\midrule
			\multirow{5}{*}{\rotatebox[origin=c]{90}{ETTh2}} 
            
			& 96  & 0.282 & \best{0.329} & 0.291 & 0.339 & 0.281 & 0.343 & \best{0.265} & \second{0.334} & 0.294 & 0.345 & \second{0.280} & 0.340 & 0.290 & 0.342 & 0.286 & 0.351 \\

            & 192 & 0.342 & \second{0.376} & 0.349 & 0.390 & \second{0.333} & 0.383 & \best{0.324} & \best{0.373} & 0.358 & 0.395 & 0.348 & 0.388 & 0.354 & 0.397 & 0.352 & 0.394 \\

            & 336 & 0.370 & \best{0.397} & 0.381 & 0.413 & 0.368 & 0.411 & \best{0.352} & \second{0.399} & 0.385 & 0.415 & \second{0.366} & 0.405 & 0.384 & 0.413 & 0.439 & 0.456 \\

            & 720 & \second{0.400} & \best{0.428} & 0.410 & 0.442 & 0.407 & 0.440 & \best{0.393} & \second{0.432} & 0.412 & 0.436 & 0.404 & 0.433 & 0.408 & 0.442 & 0.570 & 0.530 \\

			\cmidrule(lr){2-18}
			& \emph{Avg.}  & 0.348 & \best{0.383} & 0.358 & 0.396 & \second{0.347} & 0.394 & \best{0.334} & \second{0.384} & 0.362 & 0.398 & 0.349 & 0.392 & 0.359 & 0.399 & 0.412 & 0.433 \\
            
			\midrule
			\multirow{5}{*}{\rotatebox[origin=c]{90}{Weather}}
            
			& 96  & 0.147 & \best{0.184} & \second{0.143} & 0.195 & \best{0.142} & 0.195 & 0.168 & 0.221 & 0.155 & 0.206 & 0.147 & 0.197 & 0.145 & \second{0.194} & 0.170 & 0.229 \\

        & 192 & 0.192 & \best{0.228} & \second{0.188} & \second{0.236} & \best{0.186} & 0.238 & 0.212 & 0.258 & 0.199 & 0.249 & \second{0.190} & 0.240 & 0.191 & 0.237 & 0.212 & 0.268 \\

        & 336 & 0.240 & \best{0.270} & \second{0.238} & \second{0.276} & \best{0.237} & 0.278 & 0.258 & 0.292 & 0.247 & 0.284 & 0.253 & 0.285 & 0.243 & 0.279 & 0.258 & 0.307 \\

        & 720 & 0.315 & \best{0.324} & \best{0.312} & \second{0.327} & \second{0.313} & 0.330 & 0.324 & 0.338 & 0.318 & 0.336 & 0.326 & 0.339 & \second{0.313} & 0.330 & 0.321 & 0.359 \\

                \cmidrule(lr){2-18}
			& \emph{Avg.} & \second{0.223} & \best{0.251} & \best{0.220} & \second{0.259} & \best{0.220} & 0.261 & 0.240 & 0.277 & 0.230 & 0.269 & 0.229 & 0.265 & \second{0.223} & 0.260 & 0.240 & 0.291 \\

            \midrule
            \multicolumn{2}{c}{$1^{\text{st}}$ \emph{Count}} & \multicolumn{2}{c}{\best{37}}& \multicolumn{2}{c}{3} & \multicolumn{2}{c}{5} & \multicolumn{2}{c}{\second{6}} & \multicolumn{2}{c}{0} & \multicolumn{2}{c}{2} & \multicolumn{2}{c}{0} & \multicolumn{2}{c}{0}\\
			\toprule
		\end{tabular}
	\end{threeparttable}
    \caption{Results under the hyperparameter search setting described in Appendix Section~\ref{sec:hparam_search}. The lookback window is selected from $\{96, 192, 320, 512\}$, and the best configuration is reported for each model. This setup ensures that the comparison reflects each model’s optimal performance rather than a fixed setting constraint. As far as we know, we are the first open source parameter search script.}
	\label{tab::long-term-search}
\end{table*}

\begin{table*}[h!]
\begin{center}{
\setlength\tabcolsep{3pt}
\begin{tabular}{c|c|cc|cc|cc|cc|cc}
\toprule
\multicolumn{2}{c|}{Methods}&\multicolumn{2}{c|}{TimeMosaic}&\multicolumn{2}{c|}{{LLMTime}}&\multicolumn{2}{c|}{GPT4TS}&\multicolumn{2}{c|}{DLinear}&\multicolumn{2}{c}{PatchTST}\\
\midrule
\multicolumn{2}{c|}{Metric} & MSE & MAE & {MSE} & {MAE} & MSE & MAE & MSE & MAE & MSE & MAE\\
\midrule
\multirow{5}{*}{\rotatebox{0}{$ETTh1$} $\rightarrow$ \rotatebox{0}{$ETTh2$}} & 
96  &\best{0.301} & \best{0.341} &0.510 &0.576 & 0.335 & 0.374 & 0.347 & 0.400 & \second{0.304} & \second{0.350}  \\
& 192 & \best{0.375} & \best{0.387} &0.523 &0.586 & 0.412 & 0.417 & 0.447 & 0.460 & \second{0.386} & \second{0.400} \\
& 336 & \best{0.411} & \best{0.416} &0.640 &0.637 & 0.441 & 0.444 & 0.515 & 0.505 & \second{0.414} & \second{0.428} \\
& 720 & \best{0.414} & \best{0.419} &2.296 &1.034 & 0.438 & 0.452 & 0.665 & 0.589 & \second{0.419} & \second{0.443} \\
&Avg & \best{0.375} & \best{0.391} &0.992 &0.708 & 0.406 & 0.422 & 0.493 & 0.488 & \second{0.380} & \second{0.405} \\
\midrule
\multirow{5}{*}{\rotatebox{0}{$ETTh1 $} $\rightarrow$ \rotatebox{0}{$ETTm2 $}}
& 96  & \best{0.212} & \best{0.302} &0.646&0.563& 0.236 & 0.315 & 0.255 & 0.357 & \second{0.215} & \second{0.304} \\
& 192 & \best{0.275} & \best{0.328} &0.934&0.654& 0.287 & 0.342 & 0.338 & 0.413 & \second{0.275} & \second{0.339} \\
& 336 & \best{0.331} & \best{0.361} &1.157&0.728& 0.341 & 0.374 & 0.425 & 0.465 & \second{0.334} & \second{0.373}\\
& 720 & \best{0.428} & \best{0.342} &4.730 &1.531 & 0.435 & \second{0.422} & 0.640 & 0.573 & \second{0.431} & 0.424\\
&Avg & \best{0.312} & \best{0.333} &1.867 &0.869 & 0.325 & 0.363 & 0.415 & 0.452 & \second{0.314} & \second{0.360} \\
\midrule
\multirow{5}{*}{\rotatebox{0}{$ETTh2 $} $\rightarrow$ \rotatebox{0}{$ETTh1 $}}
& 96  & \best{0.471} & \best{0.454} &1.130 &0.777 & 0.732 & 0.577 & 0.689 & 0.555 & \second{0.485} & \second{0.465}\\
& 192 & \best{0.524} & \best{0.483} &1.242 &0.820 & 0.758 & 0.559 & 0.707 & 0.568 &\second{0.565} & \second{0.509} \\
& 336 & \best{0.586} & \best{0.509} &1.328 &0.864 & 0.759 & 0.578 & 0.710 & 0.577 & \second{0.581} & \second{0.515} \\
& 720 & \best{0.595} & \best{0.544} &4.145 &1.461 & 0.781 & 0.597 & 0.704 & 0.596 & \second{0.628} & \second{0.561}\\
&Avg & \best{0.544} & \best{0.498} &1.961 &0.981 & 0.757 & 0.578 & 0.703 & 0.574 & \second{0.565} & \second{0.513}\\
\midrule
\multirow{5}{*}{\rotatebox{0}{$ETTh2 $} $\rightarrow$ \rotatebox{0}{$ETTm2 $}}
& 96  & \best{0.226} & \best{0.308} &0.646&0.563& 0.253 & 0.329 & 0.240 & 0.336 & \second{0.226} & \second{0.309}\\
& 192 & \best{0.285} & \best{0.341} &0.934&0.654& 0.293 & 0.346 & 0.295 & 0.369 & \second{0.289} & \second{0.345} \\
& 336 & \best{0.340} & \best{0.370} &1.157&0.728& 0.347 & \second{0.376} & \second{0.345} & 0.397 & 0.348 & 0.379\\
& 720 & \best{0.431} & \best{0.417} &4.730 &1.531 & 0.446 & 0.429 & \second{0.432} & 0.442 &  0.439 & 0.427\\
&Avg & \best{0.321} & \best{0.359} &1.867 &0.869 & 0.335 & 0.370 & 0.328 & 0.386 & \second{0.325} & \second{0.365}\\
\midrule
\multirow{5}{*}{\rotatebox{0}{$ETTm1 $} $\rightarrow$ \rotatebox{0}{$ETTh2 $}}
& 96  & \best{0.331} & \best{0.379} &0.510 &0.576 & \second{0.353} & 0.392 & 0.365 & 0.415 & 0.354 & \second{0.385}\\
& 192 & \best{0.399} & \best{0.420} &0.523 &0.586 & \second{0.443} & 0.437 & 0.454 & 0.462 & 0.447 & \second{0.434}\\
& 336 & \best{0.418} & \best{0.443} &0.640 &0.637 & \second{0.469} & \second{0.461} & 0.496 & 0.494 & 0.481 & 0.463\\
& 720 & \best{0.416} & \best{0.446} &2.296 &1.034 & \second{0.466} & \second{0.468} & 0.541 & 0.529 & 0.474 & 0.471\\
&Avg & \best{0.391} & \best{0.422} &0.992 &0.708 & \second{0.433} & 0.439 & 0.464 & 0.475 & 0.439 & \second{0.438}\\
\midrule
\multirow{5}{*}{\rotatebox{0}{$ETTm1 $} $\rightarrow$ \rotatebox{0}{$ETTm2 $}}
& 96  & \best{0.179} & \best{0.267} &0.646&0.563& 0.217 & 0.294 & 0.221 & 0.314 & \second{0.195} & \second{0.271}\\
& 192 & \best{0.237} & \best{0.328} &0.934&0.654& 0.277 & 0.327 & 0.286 & 0.359 & \second{0.258} & \second{0.311}\\
& 336 & \best{0.300} & \best{0.348} &1.157&0.728& 0.331 & 0.360 & 0.357 & 0.406 & \second{0.317} & \second{0.348}\\
& 720 & \best{0.385} & \best{0.377} &4.730 &1.531 & 0.429 & 0.413 & 0.476 & 0.476 & \second{0.416} & \second{0.404}\\
&Avg & \best{0.278} & \best{0.330} &1.867 &0.869 & 0.313 & 0.348 & 0.335 & 0.389 & \second{0.296} & \second{0.334}\\

\bottomrule

\end{tabular}
}
\caption{{Full zero-shot learning results on ETT datasets. A lower value indicates better performance.}}
\label{tab:zero-shot-forecasting}
\end{center}
\end{table*}

\begin{table*}[h!]
\begin{center}{
\footnotesize
\setlength\tabcolsep{3pt}
\begin{tabular}{c|c|cc|cc|cc|cc|cc|cc|cc|cc|cc}
\toprule
\multicolumn{2}{c|}{Methods}&\multicolumn{2}{c|}{TimeMosaic}&\multicolumn{2}{c|}{$\mathrm{TimeMoE_{l}}$}&\multicolumn{2}{c|}{$\mathrm{TimeMoE_{b}}$}&\multicolumn{2}{c|}{$\mathrm{MOIRAI_{l}}$}&\multicolumn{2}{c|}{$\mathrm{MOIRAI_{b}}$}&\multicolumn{2}{c|}{$\mathrm{Chronos_{b}}$}&\multicolumn{2}{c|}{$\mathrm{Chronos_{s}}$}&\multicolumn{2}{c|}{TimesFM}&\multicolumn{2}{c}{Moment}\\
\midrule
\multicolumn{2}{c|}{Metric} & MSE & MAE & {MSE} & {MAE} & MSE & MAE & MSE & MAE & MSE & MAE & MSE & MAE & MSE & MAE & MSE & MAE & MSE & MAE \\
\midrule
\multirow{5}{*}{\rotatebox[origin=c]{90}{ETTh1}}  
& 96  & 0.367 & 0.395 
& \best{0.350} & 0.382 
& \second{0.357} & \second{0.381} 
& 0.381 & 0.388  
& 0.376 & 0.392 
& 0.384 & \best{0.379} 
& 0.394 & \second{0.381} 
& 0.414 & 0.404 
& 0.688 & 0.557 \\

& 192 
& 0.395 & \second{0.412} 
& \second{0.388} & \second{0.412} 
& \best{0.384} & \best{0.404} 
& 0.434 & 0.415 
& 0.412 & 0.413 
& 0.441 & \second{0.412} 
& 0.455 & 0.414 
& 0.465 & 0.434 
& 0.688 & 0.560 \\

& 336 
& \best{0.410} & \best{0.423} 
& \second{0.411} & 0.430 
& \second{0.411} & 0.434 
& 0.495 & 0.445 
& 0.433 & \second{0.428} 
& 0.475 & 0.430 
& 0.499 & 0.444 
& 0.503 & 0.456 
& 0.675 & 0.563 \\

& 720 
& \best{0.422} & \best{0.443} 
& \second{0.427} & 0.455 
& 0.449 & 0.477 
& 0.611 & 0.510 
& 0.447 & \second{0.444} 
& 0.472 & 0.446 
& 0.520 & 0.476 
& 0.511 & 0.481 
& 0.683 & 0.585 \\

\cmidrule(lr){2-20}

& Avg 
& \second{0.399} & \second{0.418} 
& \best{0.394} & 0.419 
& 0.400 & 0.424 
& 0.480 & 0.439 
& 0.417 & 0.419 
& 0.443 & \best{0.416} 
& 0.467 & 0.428 
& 0.473 & 0.443 
& 0.683 & 0.566 \\

\midrule
\multirow{5}{*}{\rotatebox[origin=c]{90}{ETTh2}}  
& 96  & \second{0.286} & 0.346 & 0.302 & 0.354 & 0.305 & 0.359 & 0.296 & 0.330 & 0.294 & \best{0.325} & 0.289 & 0.330 & \best{0.282} & \second{0.328} & 0.315 & 0.349 & 0.342 & 0.396  \\
& 192 & 0.353 & 0.389 & 0.364 & 0.385 & \best{0.351} & 0.386 & 0.361 & \second{0.371} & 0.365 & 0.375 & 0.359 & \best{0.369} & \second{0.354} & 0.373 & 0.388 & 0.395 & \second{0.354} & 0.402  \\
& 336 & \second{0.376} & 0.411 & 0.417 & 0.425 & 0.391 & 0.418 & 0.390 & \best{0.390} & \second{0.376} & \best{0.390} & 0.399 & 0.400 & 0.416 & 0.410 & 0.422 & 0.427 & \best{0.356} & 0.407 \\
& 720 & \second{0.402} & 0.435 & 0.537 & 0.496 & 0.419 & 0.454 & \best{0.423} & \best{0.418} & 0.416 & 0.433 & 0.420 & \second{0.425} & 0.428 & 0.431 & 0.443 & 0.454 & \best{0.395} & 0.434 \\

\cmidrule(lr){2-20}

& Avg & \best{0.354} & 0.395 & 0.405 & 0.415 & 0.366 & 0.404 & 0.367 & \best{0.377} & 0.362 & 0.382 & 0.366 & \second{0.381} & 0.370 & 0.385 & 0.392 & 0.406 & \second{0.361} & 0.409 \\

\midrule
\multirow{5}{*}{\rotatebox[origin=c]{90}{ETTm1}}  
& 96  & 0.392 & 0.399 
& \best{0.309} & 0.357 
& 0.338 & 0.368  
& 0.380 & \second{0.361} 
& 0.363 & \best{0.356} 
& 0.331 & \second{0.333} 
& \second{0.328} & \best{0.332} 
& 0.361 & 0.370 
& 0.654 & 0.527 \\

& 192 
& 0.422 & 0.416 
& \best{0.346} & \second{0.381} 
& \second{0.353} & 0.388 
& 0.412 & 0.383  
& 0.388 & 0.375 
& 0.386 & \best{0.365} 
& \best{0.365} & 0.384 
& 0.414 & 0.405 
& 0.662 & 0.532 \\

& 336 
& 0.450 & 0.432 
& \best{0.373} & 0.408 
& \second{0.381} & 0.413 
& 0.436 & \second{0.400}  
& 0.416 & 0.392 
& 0.408 & \best{0.382} 
& 0.391 & 0.425 
& 0.445 & 0.429 
& 0.672 & 0.537  \\

& 720 
& 0.500 & 0.459 
& 0.475 & 0.477 
& 0.504 & 0.493  
& \best{0.462} & \best{0.420}  
& \second{0.460} & \second{0.418} 
& 0.503 & \second{0.430} 
& 0.445 & 0.525 
& 0.512 & 0.471 
& 0.692 & 0.551 \\

\cmidrule(lr){2-20}

& Avg 
& 0.441 & 0.427 
& \best{0.375} & \second{0.405} 
& 0.394 & 0.415  
& 0.422 & 0.391  
& 0.406 & 0.385 
& 0.407 & \best{0.377} 
& \second{0.382} & 0.416 
& 0.433 & 0.418 
& 0.670 & 0.536 \\

\midrule
\multirow{5}{*}{\rotatebox[origin=c]{90}{ETTm2}}  
& 96  
& 0.190 & 0.276 
& 0.197 & 0.286 
& 0.201 & 0.291 
& 0.211 & \second{0.274} 
& 0.205 & \second{0.273} 
& \best{0.177} & \best{0.244} 
& \second{0.180} & 0.251 
& 0.202 & 0.270 
& 0.260 & 0.335 \\

& 192 
& \best{0.249} & 0.313 
& 0.250 & 0.322 
& 0.258 & 0.334 
& 0.281 & 0.318 
& 0.275 & 0.316 
& \second{0.251} & \best{0.293} 
& \second{0.251} & \second{0.298} 
& 0.289 & 0.321 
& 0.289 & 0.350 \\

& 336 
& \best{0.305} & 0.347 
& 0.337 & 0.375 
& 0.324 & 0.373 
& 0.341 & 0.355 
& 0.329 & 0.350 
& \best{0.305} & \best{0.327} 
& \second{0.315} & \second{0.338} 
& 0.360 & 0.366 
& 0.324 & 0.369 \\

& 720 
& \second{0.396} & 0.407 
& 0.480 & 0.461 
& 0.488 & 0.464 
& 0.428 & \second{0.428} 
& 0.437 & \second{0.411} 
& 0.419 & \best{0.394} 
& 0.421 & \best{0.403} 
& 0.462 & 0.430 
& \best{0.394} & 0.409 \\

\cmidrule(lr){2-20}

& Avg 
& \best{0.285} & 0.336 
& 0.316 & 0.361 
& 0.317 & 0.365 
& \second{0.315} & 0.343 
& 0.311 & \second{0.337} 
& \second{0.288} & \best{0.314} 
& 0.291 & 0.330 
& 0.328 & 0.346 
& 0.316 & 0.365 \\

\midrule
\multirow{5}{*}{\rotatebox[origin=c]{90}{Weather}}  
& 96  
& 0.178 & 0.236 
& \best{0.159} & 0.213 
& \second{0.160} & 0.214 
& 0.278 & 0.376 
& 0.220 & 0.217 
& 0.177 & \second{0.210} 
& 0.172 & \best{0.206} 
& - & - 
& 0.243 & 0.255 \\

& 192 
& 0.227 & 0.279 
& 0.215 & 0.266 
& \best{0.210} & \best{0.260} 
& 0.301 & 0.409 
& 0.271 & 0.259 
& 0.224 & \second{0.253} 
& 0.218 & \second{0.248} 
& - & - 
& 0.278 & 0.329 \\

& 336 
& 0.279 & 0.316 
& 0.291 & 0.322 
& 0.309 & \second{0.309} 
& 0.329 & 0.420 
& 0.286 & 0.297 
& \best{0.260} & \best{0.276} 
& \second{0.266} & 0.282 
& - & - 
& 0.306 & 0.346 \\

& 720 
& 0.351 & 0.365 
& 0.415 & 0.400 
& 0.418 & 0.405 
& 0.370 & 0.463 
& 0.373 & 0.354  
& \best{0.345} & 0.331 
& \second{0.358} & \best{0.339} 
& - & - 
& \second{0.350} & \second{0.374} \\

\cmidrule(lr){2-20}

& Avg 
& 0.259 & 0.299 
& 0.270 & 0.300 
& 0.274 & 0.297 
& 0.319 & 0.417  
& 0.287 & 0.281  
& \best{0.251} & \best{0.267} 
& \second{0.253} & \second{0.268} 
& - & - 
& 0.294 & 0.326 \\

\midrule
\multirow{5}{*}{\rotatebox[origin=c]{90}{Global Temp}}  
& 96  
& 0.237 & 0.366 
& \best{0.219} & \best{0.341} 
& 0.230 & 0.350 
& 0.278 & 0.376 
& 0.273 & 0.377  
& \second{0.236} & 0.352 
& 0.233 & \second{0.348} 
& 0.255 & 0.375 
& 0.363 & 0.472 \\

& 192 
& 0.288 & 0.409 
& \best{0.265} & \best{0.381} 
& 0.268 & 0.385 
& 0.301 & 0.409 
& 0.304 & 0.409  
& \second{0.287} & 0.398 
& \second{0.287} & \second{0.397} 
& 0.313 & 0.423 
& 0.387 & 0.489 \\

& 336 
& 0.341 & 0.449 
& \best{0.326} & \best{0.426} 
& \best{0.326} & \second{0.427} 
& 0.329 & \second{0.420} 
& 0.332 & 0.437 
& 0.332 & 0.433 
& \second{0.320} & 0.430 
& 0.362 & 0.460 
& 0.430 & 0.517 \\

& 720 
& 0.479 & 0.542 
& \best{0.344} & \best{0.453} 
& 0.377 & 0.467  
& 0.379 & 0.467 
& 0.379 & 0.469 
& \second{0.463} & 0.524 
& 0.452 & \second{0.521} 
& 0.486 & 0.545 
& 0.582 & 0.617 \\

\cmidrule(lr){2-20}

& Avg 
& 0.336 & 0.442 
& \best{0.288} & \best{0.400} 
& \second{0.300} & 0.407 
& 0.321 & \second{0.418}  
& 0.322 & 0.423 
& 0.329 & 0.426 
& 0.323 & 0.424 
& 0.354 & 0.451 
& 0.440 & 0.524 \\

\bottomrule
\end{tabular}
}
\caption{
Zero-shot forecasting results on six datasets. TimeMosaic is trained with an input length of 512 and an output horizon of 720. The symbols $s$, $b$, and $l$ represent the small, base, and large versions, respectively.
}

\label{tab:blast}
\end{center}
\end{table*}

\begin{table}[h]
\centering
\scriptsize
\begin{tabular}{c|c|cc|cc}
\toprule
\multicolumn{2}{c|}{\multirow{2}{*}{\scalebox{1.1}{Dataset}}}&\multicolumn{2}{c|}{z-score}&\multicolumn{2}{c}{RevIN}\\
\cmidrule(lr){3-4} \cmidrule(lr){5-6} 
 \multicolumn{2}{c|}{}  &  MSE & MAE &  MSE & MAE \\
\midrule
\multirow{4}{*}{\rotatebox[origin=c]{90}{ETTm1}}  & 96  & 0.304 & 0.344 & 0.299 & 0.340 \\
      & 192 & 0.339 & 0.364 & 0.341 & 0.365 \\
      & 336 & 0.375 & 0.385 & 0.359 & 0.380 \\
      & 720 & 0.422 & 0.413 & 0.424 & 0.414 \\
\midrule
\multirow{4}{*}{\rotatebox[origin=c]{90}{ETTm2}}  & 96  & 0.165 & 0.249 & 0.164 & 0.247 \\
      & 192 & 0.230 & 0.290 & 0.226 & 0.291 \\
      & 336 & 0.282 & 0.327 & 0.279 & 0.324 \\
      & 720 & 0.356 & 0.374 & 0.353 & 0.376 \\
\midrule
\multirow{4}{*}{\rotatebox[origin=c]{90}{ETTh1}}  & 96  & 0.362 & 0.387 & 0.358 & 0.386 \\
      & 192 & 0.400 & 0.411 & 0.398 & 0.410 \\
      & 336 & 0.436 & 0.435 & 0.435 & 0.435 \\
      & 720 & 0.451 & 0.460 & 0.481 & 0.480 \\
\midrule
\multirow{4}{*}{\rotatebox[origin=c]{90}{ETTh2}} & 96  & 0.300 & 0.342 & 0.302 & 0.342 \\
      & 192 & 0.354 & 0.387 & 0.356 & 0.386 \\
      & 336 & 0.378 & 0.403 & 0.378 & 0.402 \\
      & 720 & 0.421 & 0.439 & 0.438 & 0.451 \\
\midrule
\multirow{4}{*}{\rotatebox[origin=c]{90}{Weather}} & 96  & 0.144 & 0.241 & 0.152 & 0.189 \\
        & 192 & 0.197 & 0.232 & 0.192 & 0.229 \\
        & 336 & 0.245 & 0.276 & 0.245 & 0.272 \\
        & 720 & 0.327 & 0.327 & 0.328 & 0.328 \\
\midrule
\multirow{4}{*}{\rotatebox[origin=c]{90}{Electricity}} & 96  & 0.140 & 0.236 & 0.140 & 0.236 \\
            & 192 & 0.156 & 0.250 & 0.155 & 0.250 \\
            & 336 & 0.173 & 0.267 & 0.173 & 0.267 \\
            & 720 & 0.212 & 0.298 & 0.212 & 0.298 \\
\midrule
\multirow{4}{*}{\rotatebox[origin=c]{90}{Traffic}} & 96  & 0.410 & 0.276 & 0.409 & 0.274 \\
        & 192 & 0.424 & 0.281 & 0.424 & 0.280 \\
        & 336 & 0.435 & 0.287 & 0.437 & 0.287 \\
        & 720 & 0.464 & 0.302 & 0.464 & 0.301 \\
\midrule
\multirow{4}{*}{\rotatebox[origin=c]{90}{Solar-Energy}}  & 96  & 0.200 & 0.226 & 0.204 & 0.222 \\
             & 192 & 0.225 & 0.244 & 0.223 & 0.235 \\
             & 336 & 0.238 & 0.245 & 0.236 & 0.238 \\
             & 720 & 0.241 & 0.251 & 0.236 & 0.243 \\
\bottomrule
\end{tabular}
\caption{Forecasting performance (MSE and MAE) under z-score vs. RevIN normalization across eight datasets.}
\label{tab:normalization-comparison}
\end{table}

\begin{figure}[t!]
  % \centerin
  \begin{subfigure}[b]{0.23\textwidth}
    \includegraphics[width=\linewidth]{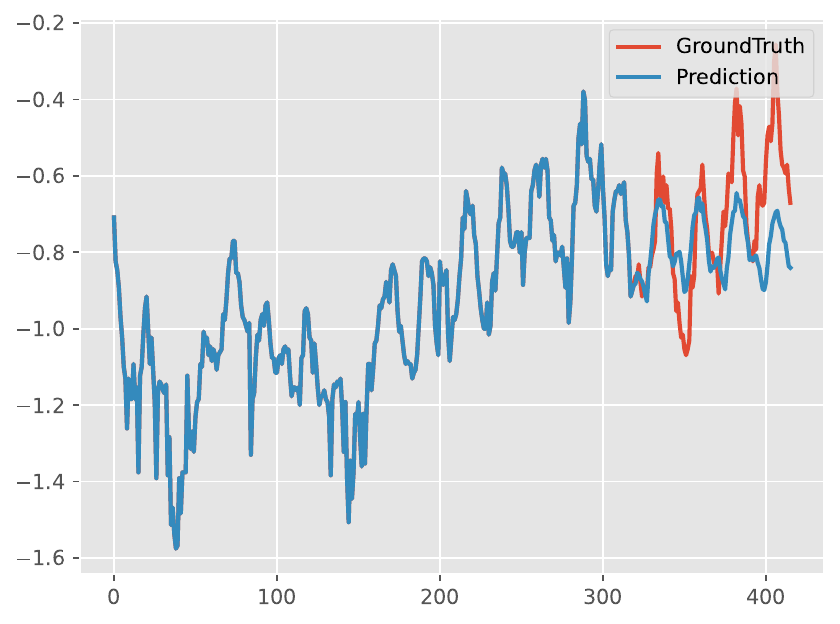}
  \end{subfigure}
  \begin{subfigure}[b]{0.23\textwidth}
    \includegraphics[width=\linewidth]{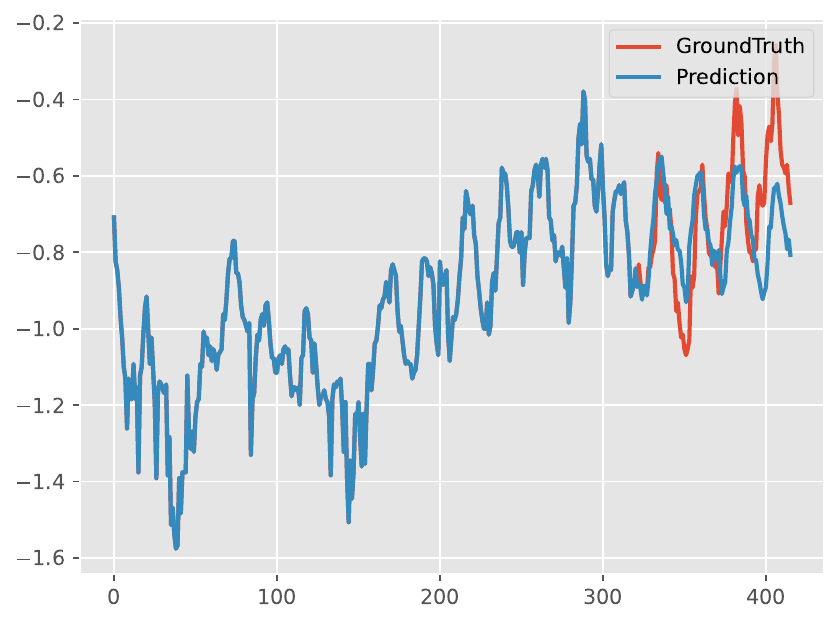}
  \end{subfigure}
  \begin{subfigure}[b]{0.23\textwidth}
    \includegraphics[width=\linewidth]{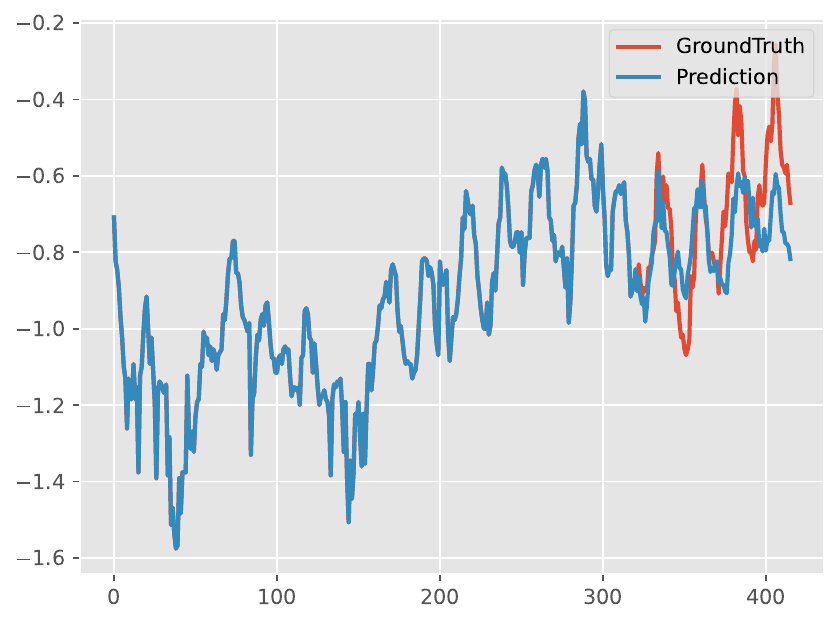}
  \end{subfigure}
  \begin{subfigure}[b]{0.23\textwidth}
    \includegraphics[width=\linewidth]{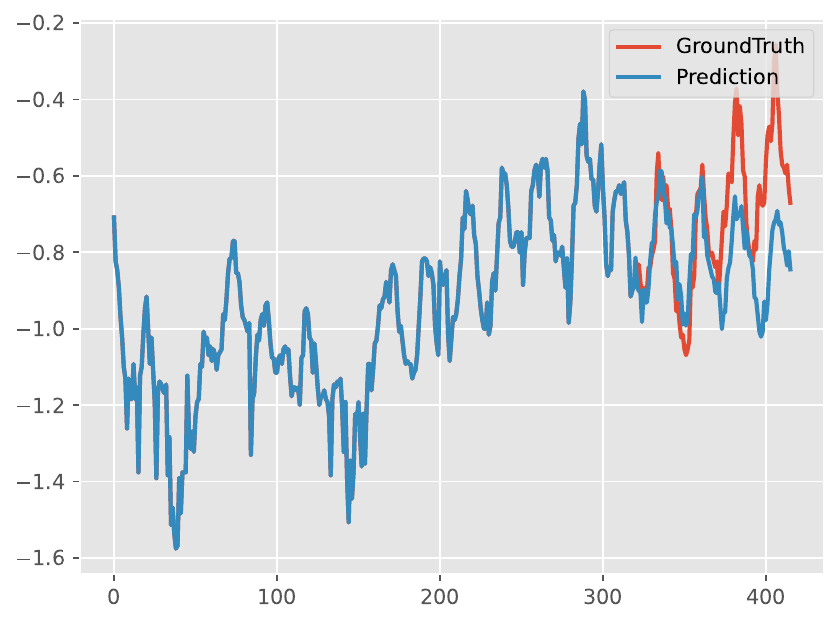}
  \end{subfigure}
  \caption{Prediction cases from ETTh1 by different models (TimeMosaic, SimpleTM, PatchTST, iTransformer) under input-320-predict-96 settings.}
  \label{fig:case1}
\end{figure}

\begin{figure}[t!]
  % \centerin
  \begin{subfigure}[b]{0.23\textwidth}
    \includegraphics[width=\linewidth]{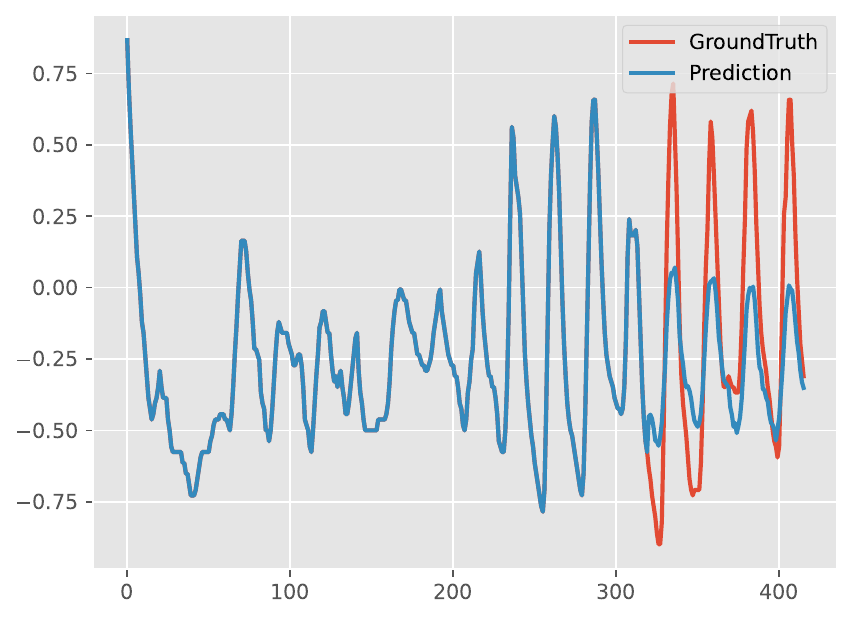}
  \end{subfigure}
  \begin{subfigure}[b]{0.23\textwidth}
    \includegraphics[width=\linewidth]{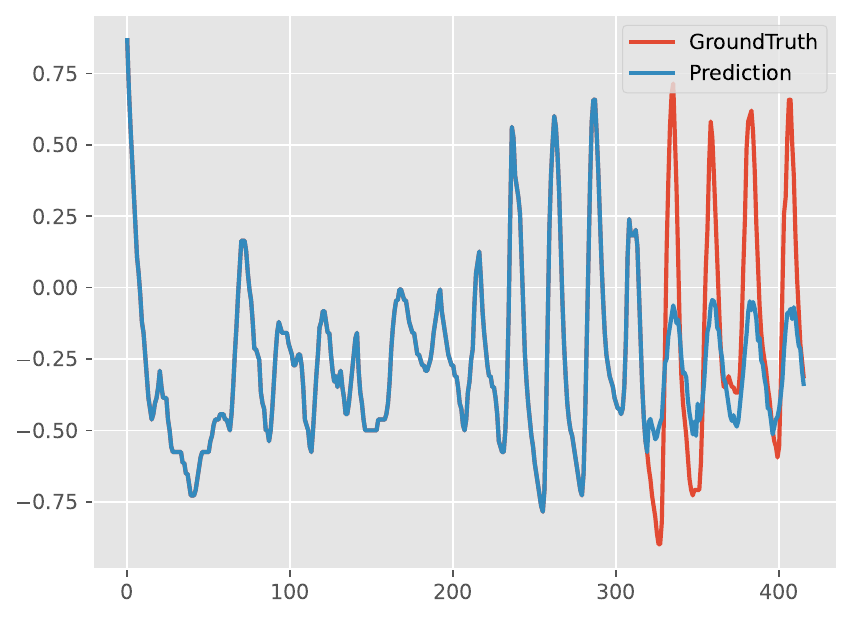}
  \end{subfigure}
  \begin{subfigure}[b]{0.23\textwidth}
    \includegraphics[width=\linewidth]{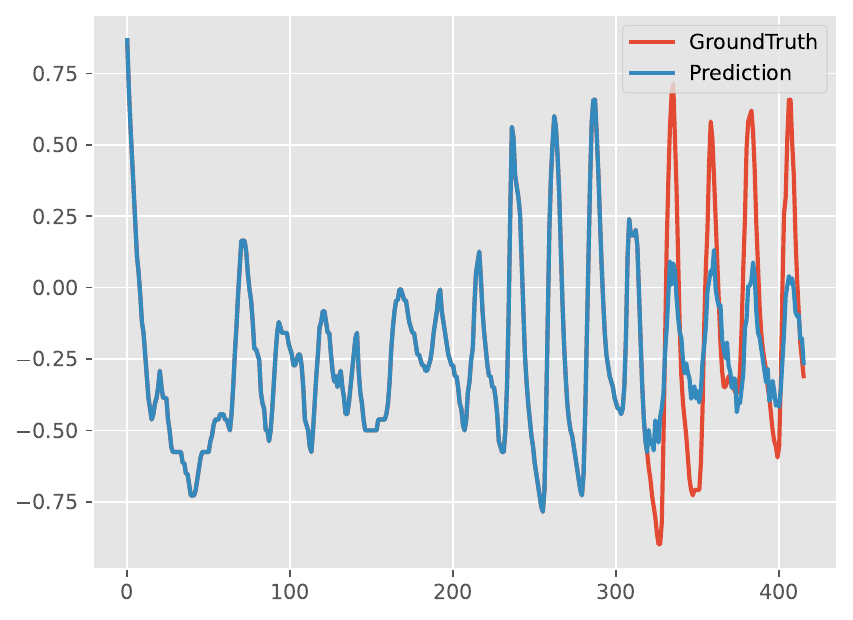}
  \end{subfigure}
  \begin{subfigure}[b]{0.23\textwidth}
    \includegraphics[width=\linewidth]{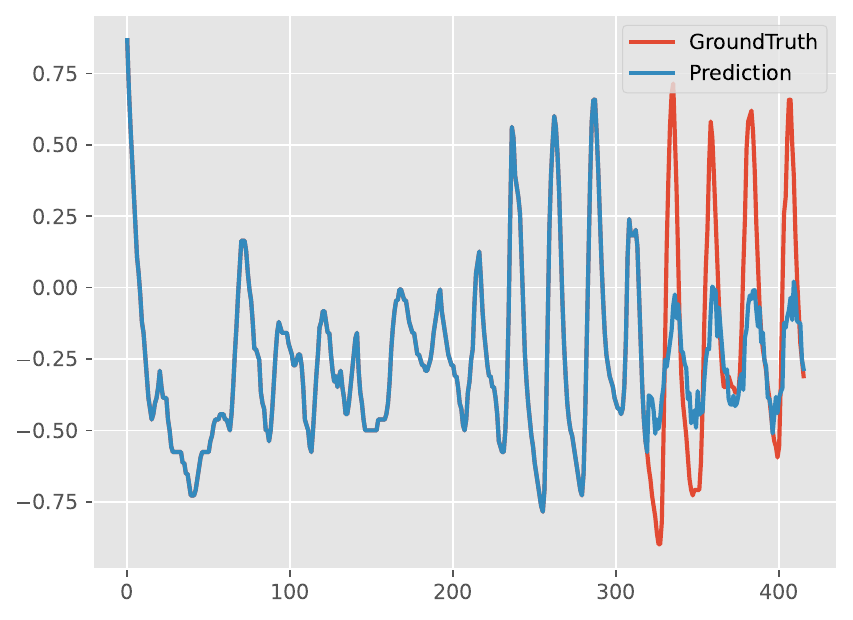}
  \end{subfigure}
  \caption{Prediction cases from ETTh2 by different models (TimeMosaic, SimpleTM, PatchTST, iTransformer) under input-320-predict-96 settings.}
  \label{fig:case2}
\end{figure}

\begin{figure}[t!]
  % \centerin
  \begin{subfigure}[b]{0.23\textwidth}
    \includegraphics[width=\linewidth]{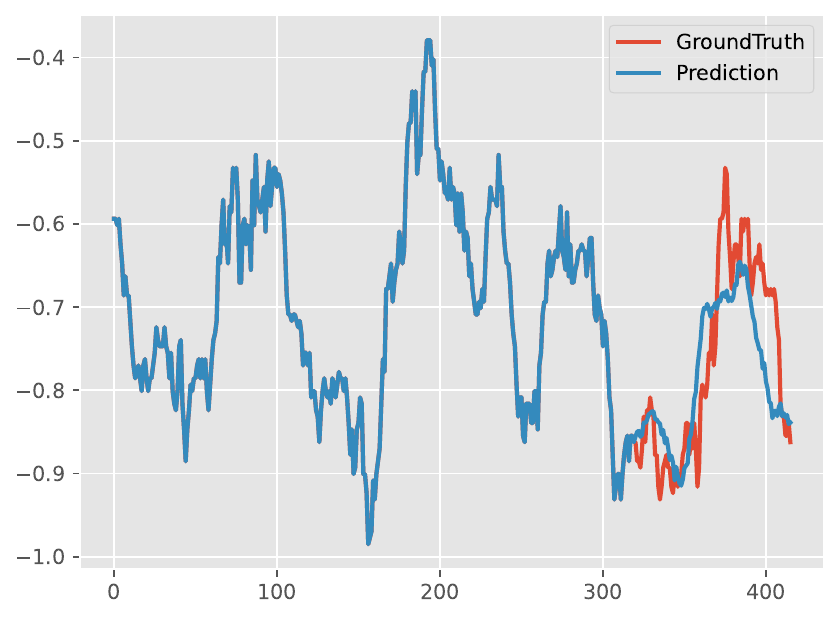}
  \end{subfigure}
  \begin{subfigure}[b]{0.23\textwidth}
    \includegraphics[width=\linewidth]{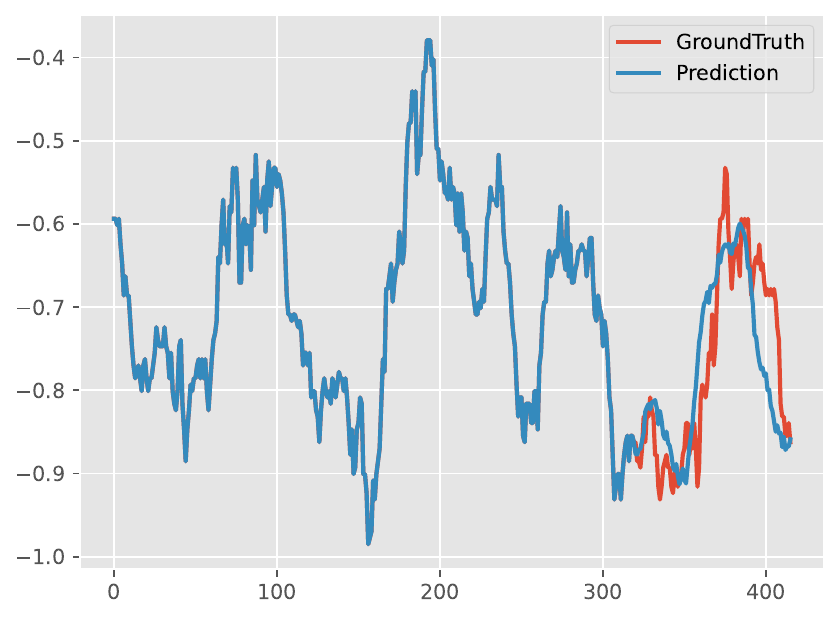}
  \end{subfigure}
  \begin{subfigure}[b]{0.23\textwidth}
    \includegraphics[width=\linewidth]{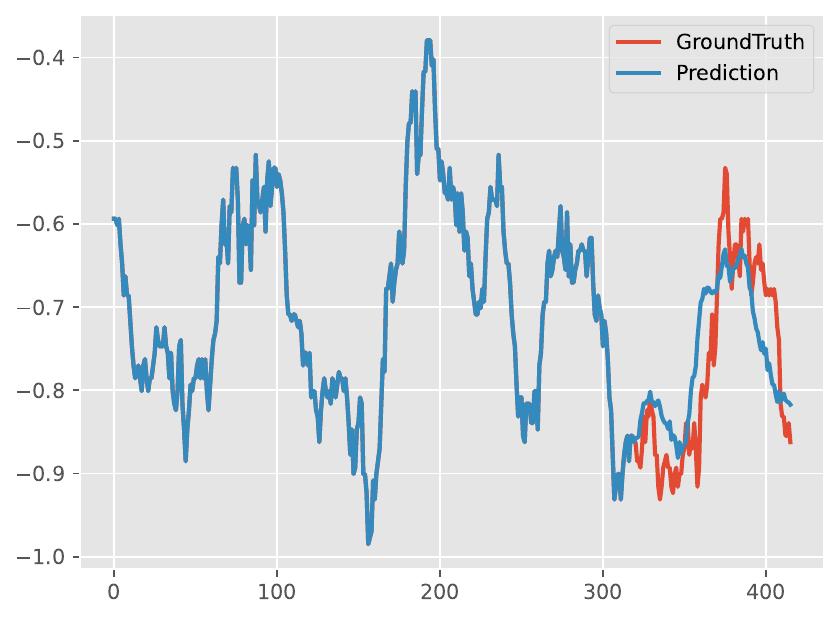}
  \end{subfigure}
  \begin{subfigure}[b]{0.23\textwidth}
    \includegraphics[width=\linewidth]{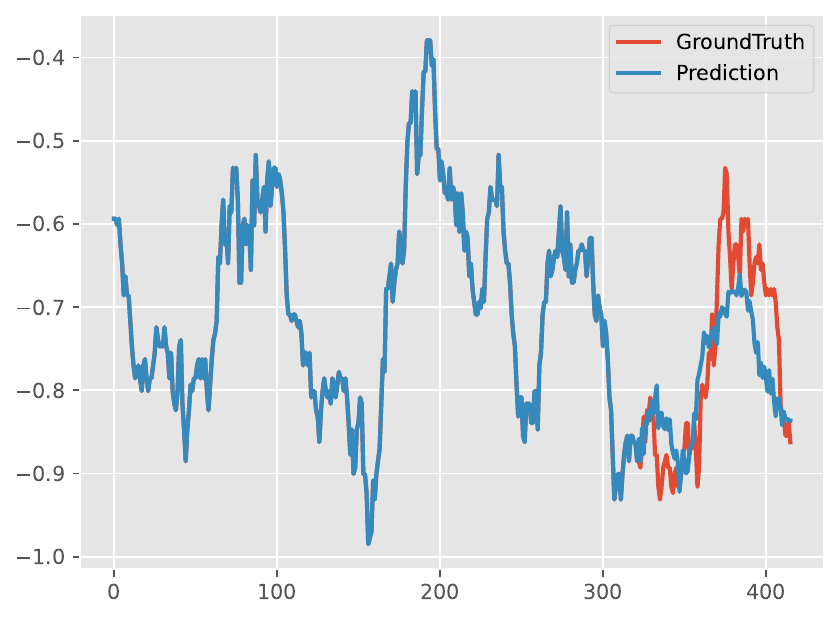}
  \end{subfigure}
  \caption{Prediction cases from ETTm1 by different models (TimeMosaic, SimpleTM, PatchTST, iTransformer) under input-320-predict-96 settings.}
  \label{fig:case3}
\end{figure}

\begin{figure}[t!]
  % \centerin
  \begin{subfigure}[b]{0.23\textwidth}
    \includegraphics[width=\linewidth]{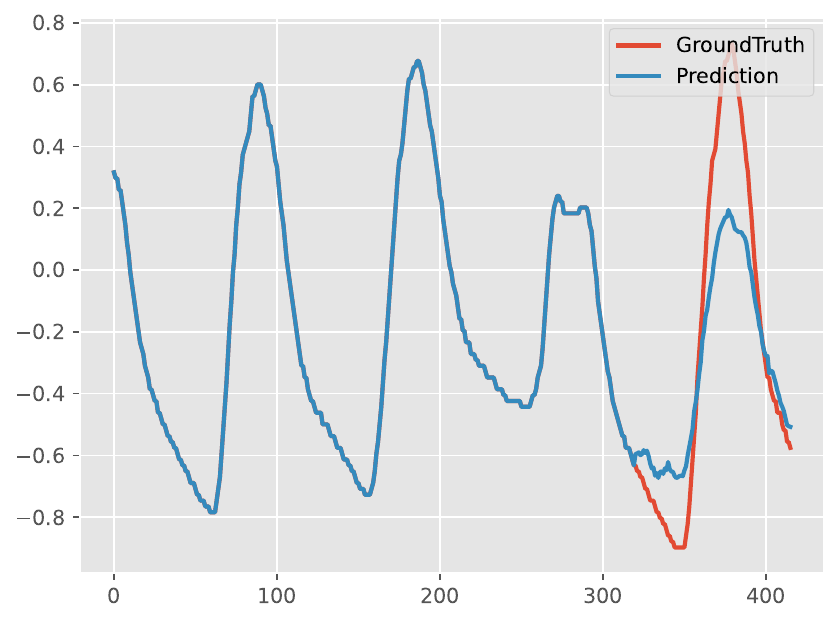}
  \end{subfigure}
  \begin{subfigure}[b]{0.23\textwidth}
    \includegraphics[width=\linewidth]{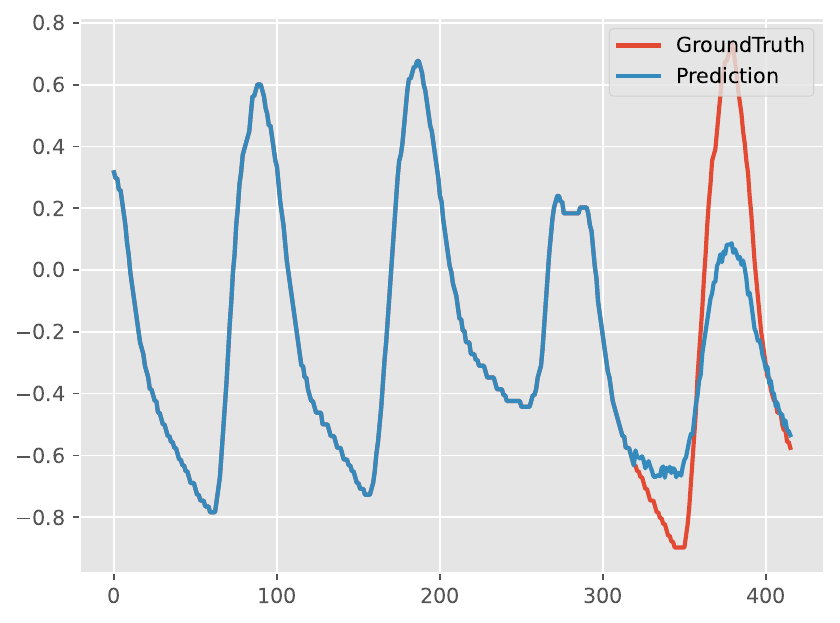}
  \end{subfigure}
  \begin{subfigure}[b]{0.23\textwidth}
    \includegraphics[width=\linewidth]{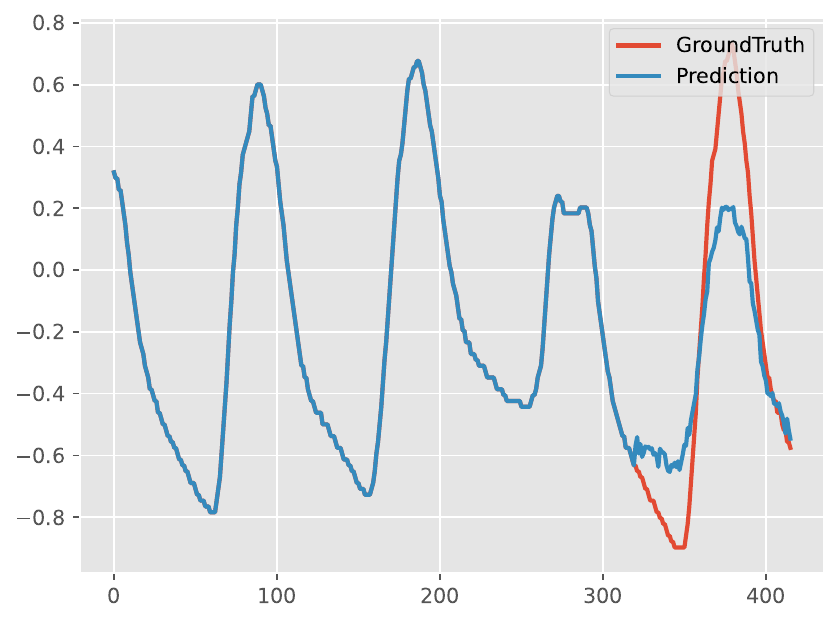}
  \end{subfigure}
  \begin{subfigure}[b]{0.23\textwidth}
    \includegraphics[width=\linewidth]{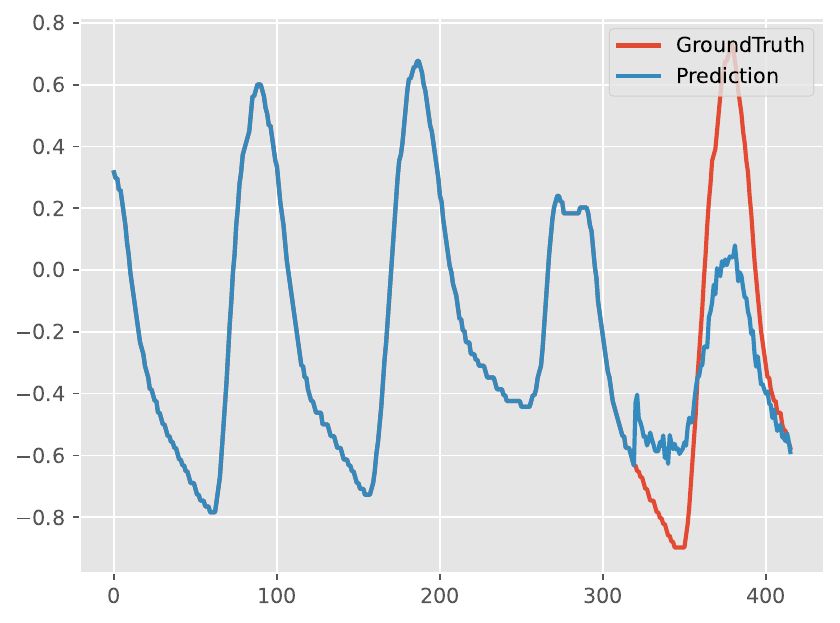}
  \end{subfigure}
  \caption{Prediction cases from ETTm2 by different models (TimeMosaic, SimpleTM, PatchTST, iTransformer) under input-320-predict-96 settings.}
  \label{fig:case4}
\end{figure}

% \begin{figure}[t!]
%   % \centerin
%   \begin{subfigure}[b]{0.23\textwidth}
%     \includegraphics[width=\linewidth]{figures/Weather_iTransformer.pdf}
%   \end{subfigure}
%   \begin{subfigure}[b]{0.23\textwidth}
%     \includegraphics[width=\linewidth]{figures/Weather_SimpleTM.pdf}
%   \end{subfigure}
%   \begin{subfigure}[b]{0.23\textwidth}
%     \includegraphics[width=\linewidth]{figures/Weather_PatchTST.pdf}
%   \end{subfigure}
%   \begin{subfigure}[b]{0.23\textwidth}
%     \includegraphics[width=\linewidth]{figures/Weather_iTransformer.pdf}
%   \end{subfigure}
%   \caption{Prediction cases from Weather by different models (TimeMosaic, SimpleTM, PatchTST, iTransformer) under input-320-predict-96 settings.}
%   \label{fig:case5}
% \end{figure}

\end{document}